\documentclass{article}

\usepackage{microtype}
\usepackage{graphicx}
\usepackage{subfigure}
\usepackage{booktabs} %
\usepackage{wrapfig}
\usepackage{multirow}
\usepackage{multicol}

\usepackage{eso-pic} %
\usepackage{url}
\usepackage{multirow}
\usepackage{subfigure} %
\usepackage{xspace}
\usepackage{enumitem}

\usepackage[accepted]{icml2024}

\usepackage{amsmath}
\usepackage{amssymb}
\usepackage{mathtools}
\usepackage{amsthm}

\usepackage{multicol}
\usepackage{fancyhdr}
\usepackage{svg}
\usepackage{xcolor} %
\usepackage{algorithm}
\usepackage{algorithmic}
\usepackage{amsfonts}       %
\usepackage{nicefrac}       %
\usepackage{microtype}      %
\usepackage{hyperref}
\usepackage[capitalize,noabbrev]{cleveref}
\theoremstyle{plain}
\newtheorem{theorem}{Theorem}[section]
\newtheorem{example}[theorem]{Example}
\newcommand{\cA}{{\mathcal A}}

\newcommand{\cG}{{\mathcal G}}

\newcommand{\cM}{{\mathcal M}}

\newcommand{\cR}{{\mathcal R}}

\newtheorem{proposition}[theorem]{Proposition}

\newtheorem{corollary}[theorem]{Corollary}
\theoremstyle{definition}
\newtheorem{definition}[theorem]{Definition}

\theoremstyle{remark}

\newcommand{\eps}{\varepsilon}
\usepackage[textsize=tiny]{todonotes}
\usepackage{stfloats}

\icmltitlerunning{A New Linear Scaling Rule for Private Adaptive Hyperparameter Optimization}

\begin{document}

\twocolumn[
\icmltitle{A New Linear Scaling Rule for Private Adaptive Hyperparameter Optimization}
\icmlsetsymbol{equal}{*}

\begin{icmlauthorlist}
\icmlauthor{Ashwinee Panda}{equal,princeton}
\icmlauthor{Xinyu Tang}{equal,princeton}
\icmlauthor{Saeed Mahloujifar}{princeton}
\icmlauthor{Vikash Sehwag}{princeton}
\icmlauthor{Prateek Mittal}{princeton}
\end{icmlauthorlist}

\icmlaffiliation{princeton}{Princeton University}

\icmlcorrespondingauthor{Ashwinee Panda}{ashwinee@princeton.edu}
\icmlkeywords{Machine Learning, ICML}
\vskip 0.3in
\vskip 0.3in
]

\printAffiliationsAndNotice{\icmlEqualContribution} %
\begin{figure}
    \centering
    \includegraphics[width=1\linewidth]{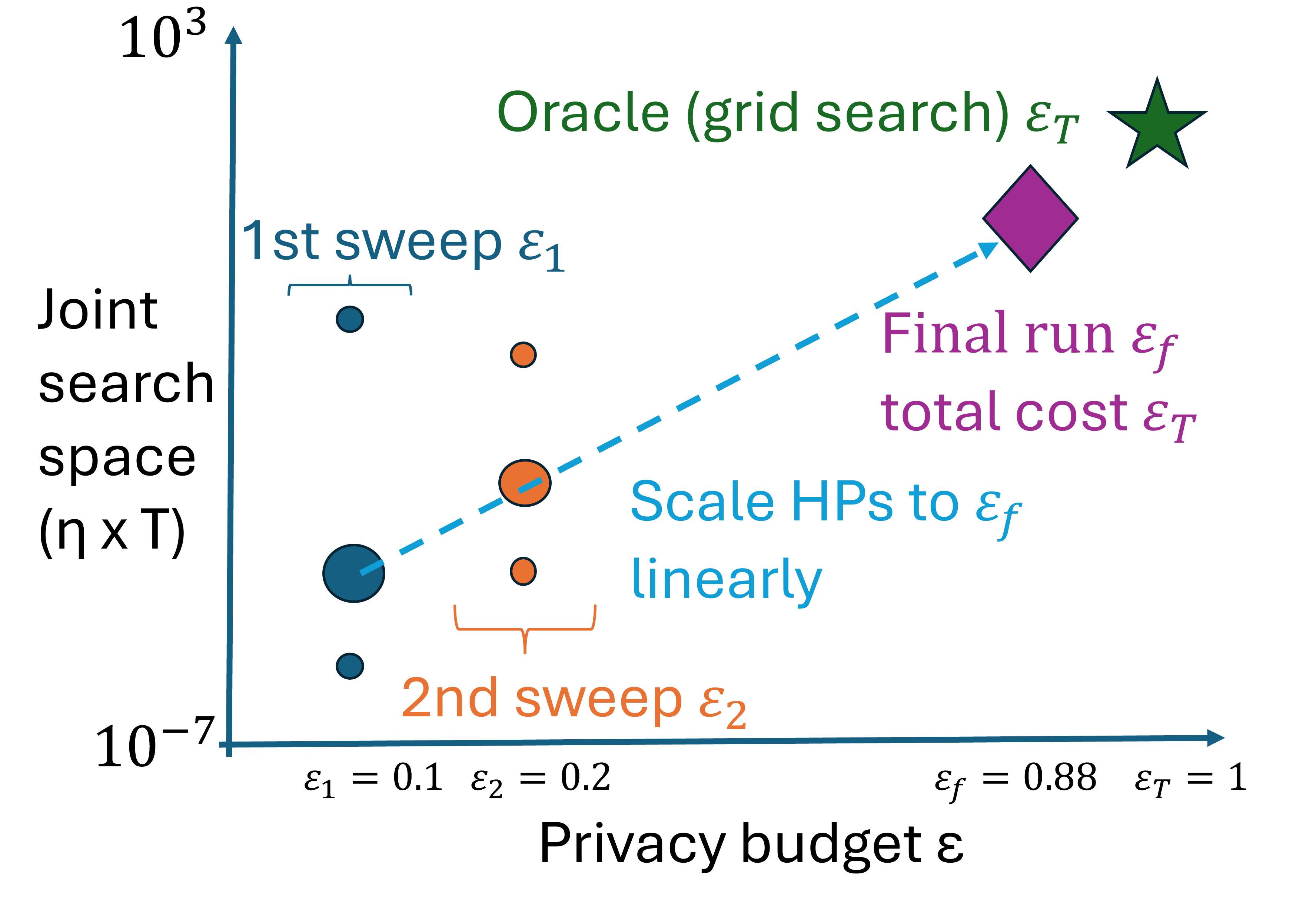}
    \caption{Visualization of our method. We use low-cost trials (small \(\eps\)) to estimate hyperparameters (HPs) and scale these up to the privacy budget for the final run. We combine multiple HPs together, and have a prior that the scaling is linear.}
    \label{fig:main-hpo}
\end{figure}
\begin{abstract}
An open problem in differentially private deep learning is hyperparameter optimization (HPO).
DP-SGD introduces new hyperparameters and complicates existing ones, forcing researchers to painstakingly tune hyperparameters with hundreds of trials, which in turn makes it impossible to account for the privacy cost of HPO without destroying the utility.
We propose an adaptive HPO method that uses cheap trials (in terms of privacy cost and runtime) to estimate optimal hyperparameters and scales them up.
We obtain state-of-the-art performance on 22 benchmark tasks, across computer vision and natural language processing, across pretraining and finetuning, across architectures and a wide range of $\varepsilon \in [0.01,8.0]$, all while accounting for the privacy cost of HPO.
\end{abstract}

\begin{figure}
    \centering
    \includegraphics[width=\linewidth]{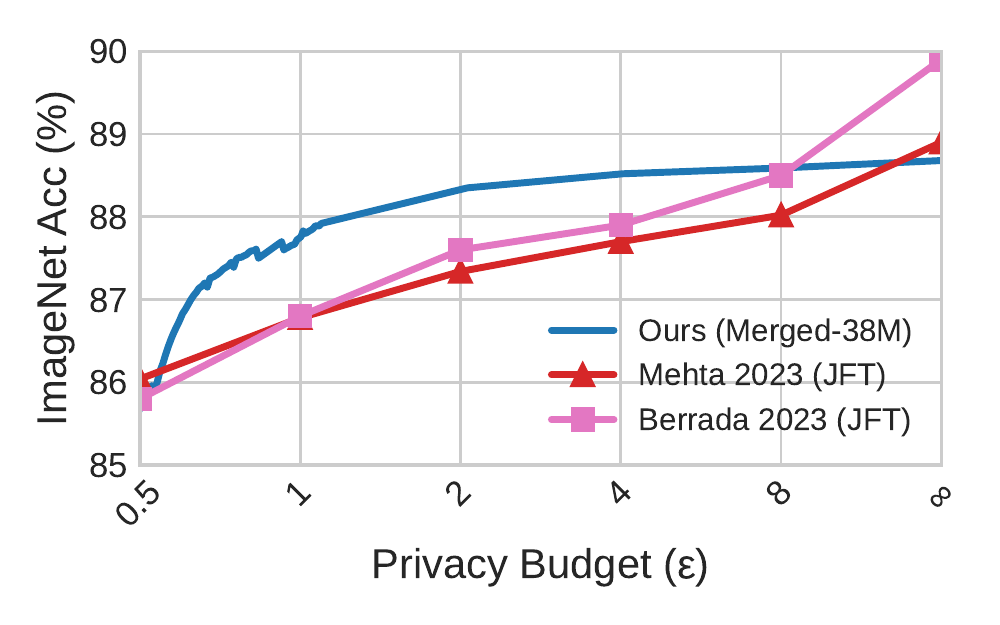}
    \caption{Evaluation on ImageNet-1k finetuning. Our HPO only requires paying the privacy cost once, and can then be used to find good HPs for all values of \(\eps>0.5\).  We outperform prior work~\citep{mehtadptransfer, berrada2023unlocking} because our HPO finds better HPs, even though prior work has better non-private performance and \textbf{does not report the privacy cost of their HPO}.}
    \label{fig:main-imagenet}
\end{figure}
\section{Introduction}
A crucial component of interfacing machine learning models closely with user data is ensuring that the process remains \emph{private}~\citep{2017LearningWP}, and 
Differential Privacy (DP) is the gold standard for quantifying privacy risks and providing provable guarantees against attacks~\citep{dworkdp}.
DP implies that the output of an algorithm e.g., the final weights trained by stochastic gradient descent (SGD) do not change much if a single datapoint in the dataset changes.

\begin{definition}[{{Differential Privacy}}]
    A randomized mechanism  $\mathcal{M}$ with domain $\mathcal{D}$ and range $\mathcal{R}$ preserves $(\varepsilon,\delta)$-differential privacy iff for any two neighboring   datasets $D,D' \in \mathcal{D}$ and for any subset $S \subseteq \mathcal{R}$ we have \(\Pr[\mathcal{M}(D) \in S] \leq e^{\varepsilon} \Pr[\mathcal{M}(D') \in S] + \delta\)
\label{def:dp}
    
   where $D$ and $D'$ are neighboring datasets if they differ in a single entry, $\varepsilon$ is the privacy budget and $\delta$ is the failure probability. 
\end{definition}

Differentially Private Stochastic Gradient Descent (DP-SGD)~\citep{songdpsgd, abadidpsgd} is the standard privacy-preserving training algorithm for training neural networks on private data. For a batch size \(B\) and learning rate \(\eta\), DP-SGD has an update rule given by
\(w^{(t+1)} = w^{(t)} - \frac{\eta_t}{|B_t|} \left(\sum_{i \in B_t} \frac{1}{\textsc{C}} \textbf{clip}_{\textsc{C}}(\nabla \ell(x_i, w^{(t)})) + \sigma \xi\right)\) 
\label{eq:dpsgd}
where the changes to SGD are the per-sample gradient clipping $\textbf{clip}_{\textsc{C}}(\nabla \ell(x_i, w^{(t)}))=\frac{C\times \nabla \ell(x_i, w^{(t)})}{\max(C, ||\nabla \ell(x_i, w^{(t)})||_2)}$, and addition of noise sampled from a $d$-dimensional Gaussian distribution $\xi \sim \mathcal{N}(0,1)$ with standard deviation $\sigma$.
These steps alter the bias-variance tradeoff of SGD and degrade utility, creating a challenging privacy-utility tradeoff.

Because private training introduces additional hyperparameters, biases optimization by clipping the gradient, and imposes privacy-utility tradeoffs for existing hyperparameters, hyperparameter optimization (HPO) in DP is challenging.
Many prior works report doing hundreds of hyperparameter trials and do not report the privacy cost of HPO in their final privacy guarantee~\citep{deepmind, bu2022scalable, bu2022differentially, googlefinetune, mehtadptransfer, berrada2023unlocking}. These works either assume that HPO does not leak privacy, that the best HPs are known beforehand, or that they can be transferred from a public dataset that is similar to the private dataset.

More recently, researchers have proposed methods that do private HPO~\citep{papernothparamtuning, koskela2023practical, wang2023dphypo} with Rényi DP.
These private HPO methods have been evaluated on MNIST and CIFAR10, but have not been validated on more challenging tasks in CV, or on LLMs.

We propose a new private adaptive HPO method (~\cref{fig:main-hpo}), which we call the new linear scaling rule. We first estimate the optimal HPs for small privacy budgets. We then scale the searched HPs linearly up to larger privacy budgets. Our full method is described in~\cref{alg:dp-raft}.
We summarize our contributions:
\begin{itemize}[leftmargin=15pt]
    \item We demonstrate that our new linear scaling rule reduces the computation and privacy cost of HPO by an order of magnitude without sacrificing performance
    \item We compare our private HPO method to random search, grid search, and 3 prior methods for private HPO
    \item We evaluate our private HPO on 22 tasks spanning computer vision and natural language processing, fine-tuning and training from scratch, training models spanning from ResNets to multi-billion-parameter Transformers
    \item We find that models trained with our method can provide good performance even when there is a large shift between public and private data
\end{itemize}

\section{Design}

We provide a set of design goals for our adaptive private HPO method and explain their importance. We use simple axioms for optimization and privacy as building blocks to motivate the high-level design of our method. We conduct preliminary experiments to quantitatively determine the relationships between key hyperparameters. Ultimately we compose the many hyperparameters of interest in DP into a single scalar variable \(r\), and present a simple adaptive approach for privately optimizing this parameter.

\subsection{Design Goals}

We draw our goals from the two simple baselines for HPO, random search and grid search. We define random search as drawing hyperparameters from the search space randomly and doing a single run with the entire privacy budget. We discuss variations on random search, such as doing multiple runs with smaller privacy budgets, in~\cref{sec:evaluation}. Random search has low runtime, is parallelizable, and has low privacy cost, but typically does not provide good performance when the hyperparameter search space is large and the set of viable solutions is sparse. Grid search typically has high runtime and privacy cost, and is also parallelizable. Given sufficient trials, grid search should approach the performance of the oracle, the run with perfectly chosen hyperparameters. 

Our method should provide:

\begin{itemize}[leftmargin=15pt]
\item Better performance than random search and almost as good as the oracle; if we define the error rate of any HPO method as the difference in performance between that method and the oracle, our method should reduce the error rate relative to random search significantly.
\item Better privacy cost than grid search and almost the same privacy cost as random search; the difference in privacy-utility tradeoff between our method and the oracle should not be the difference between a run with \(\eps=0.1\) and \(\eps=1.0\), it should be the relatively smaller gap between ex. \(\eps=0.9\) and \(\eps=1.0\).
\end{itemize}

\subsection{Building Blocks of Linear Scaling}

The design of our adaptive private HPO method is based on simple building blocks derived from known theorems in optimization and privacy. First we inspect the definition of DP-SGD and the nature of adaptive composition. Suppose we are taking T steps with noise \(\sigma\) to produce some \(\eps\) guarantee. If we relax our privacy guarantee, so now we want to achieve some \(\eps^{*} > \eps\), we can either a) fix T and reduce \(\sigma\), b) increase T and fix \(\sigma\), or c) some combination of (a) and (b). Second we turn to a rule of thumb that is popularly known as the original linear scaling rule; the optimal learning rate is inversely proportional to the noise scale in GD~\citep{malladi2022on}. In the case of DP-GD, that is in the full batch setting where there is no noise due to SGD, the learning rate should be inversely proportional to \(\sigma\). If we combine these two axioms, we get the following heuristic:

\begin{proposition}
    If we are taking T steps with noise \(\sigma\) and learning rate \(\eta\) to achieve a target \(\eps^{*}\), we can achieve a target \(\hat{\eps}>\eps^{*}\) by either: a) Fix T, reduce \(\sigma\), increase \(\eta\)) b) Increase T, fix \(\sigma\), fix \(\eta\) c) Increase T slightly, reduce \(\sigma\) slightly, increase \(\eta\) slightly.
\end{proposition}

We now formalize this intuition.

\section{Analysis of Private Gradient Descent}

We analyze the excess empirical risk of private GD as the sum of two terms. The first term is the risk of non-private GD with the same hyperparameters. The second term is the divergence of private GD from non-private GD due to the added noise term. 

We consider optimizing a function using Differentially Private Gradient Descent (DP-GD). The presence of noise in GD introduces a deviation between the iterates of GD with noise, denoted as $w_T$, and without noise, denoted as $w_{T_b}$, at iteration $T$. We first upper bound this deviation in expectation, which we refer to as the radius r. We then use this to upper bound the excess empirical risk of noisy GD. We finally use this bound to motivate the design of our private adaptive HPO method.

\subsection{Assumptions}

We present four assumptions that simplify the convergence analysis. 
We acknowledge that these assumptions do not hold true in all settings, but nevertheless provide an important foundation for illustrating the intuition of our method. 
We empirically validate the success of our algorithm in complex neural network settings, such as training a 13B-parameter OPT Transformer model on the benchmark SQuAD task, in Section 3.

\begin{itemize}
    \item A function is $\alpha$-strongly convex if for any two points $x, y$ and any subgradient $g$ at $x$, it holds that $f(y) \geq f(x) + g^\top(y-x) + \frac{\alpha}{2}\|y-x\|^2$.
    \item A function is $\beta$-smooth if its gradient is $\beta$-Lipschitz continuous, meaning for any two points $x, y$, $\| \nabla f(x) - \nabla f(y) \| \leq \beta \|x - y\|$.
    \item A function is \(L\)-Lipschitz if there exists a positive \(L\) such that \(|f(x) - f(y)| \leq L \|x - y\|\)
    \item A function satisfies the \emph{bounded gradient assumption} if there exists a constant \(C\) such that \(E[\|\nabla f(w)\| \leq C \: \forall w \in R^{d}\)
\end{itemize}

The bounded gradient assumption is implied by convexity and Lipschitzness. This allows us to ignore the impact of clipping in DP-SGD, which reduces the analysis to that of noisy GD. 

The noise added at each iteration for privacy has an expected norm $\rho = \sqrt{d} \cdot \sigma$, where $d$ is the dimension of the model, and $\sigma$ is the scale of the noise. The learning rate $\eta$ satisfies $0 < \eta < \frac{2}{\beta}$, ensuring convergence.

Let $c = \max(|1 - \eta\alpha|, |1 - \eta\beta|)$, which characterizes the contraction factor in the optimization process. Given that $\eta$ is chosen appropriately, we have $0 < c < 1$.  

\subsection{Definitions}

The empirical loss $\mathcal{L}(w_T)$ for a model parameterized by $w_T$ (ex. iterate \(T\) of GD) over a dataset $\mathcal{D} = \{ (x_i, y_i) \}_{i=1}^{N}$ is defined as the average loss over all training samples:

\[
\mathcal{L}(w_T) = \frac{1}{N} \sum_{i=1}^{N} \ell(f(x_i; w_T), y_i)
\]

The goal of our private Hyperparameter Optimization (HPO) is to find the hyperparameter set $\Lambda^*$ that minimizes the loss:

\[
\Lambda^* = \underset{\Lambda}{\mathrm{argmin}} \; \mathcal{L}_{\text{val}}(\Lambda)
\]

where $\mathcal{L}_{\text{val}}(\Lambda)$ denotes the loss on a validation dataset for a given hyperparameter configuration $\Lambda$.
Because Noisy GD typically does not overfit due to the heavy regularization effect of the noise, and to make the convergence analysis straightforward, we use the empirical loss as a proxy for the validation loss throughout. We will analyze the excess empirical risk to motivate the design of our private HPO method.

Let $w_T$ be the \(T^{th}\) iterate of noisy GD that optimizes a function satisfying the assumptions, and \(w_{T_b}\) be the \(T^{th}\) iterate of non-noisy GD that optimizes that same function.
We define the excess empirical risk of noisy GD as:

\begin{align*}
R_{\text{noisy}} &= E[\mathcal{L}(w_T)] - \mathcal{L}(w^*), \\
&\leq E[\mathcal{L}(w_T) - \mathcal{L}(w_{T_b})] + \mathcal{L}(w_{T_b}) - \mathcal{L}(w^*) \\
&\leq E[L \cdot \|w_T - w_{T_b}\|] + R_{\text{non-noisy}}
\end{align*}

Where $\mathcal{L}(w^*)$ denotes the empirical loss at the optimal parameter set (without noise). \(R_{\text{non-noisy}} = \mathcal{L}(w_{T_b}) - \mathcal{L}(w^*\) is the excess empirical risk of non-noisy GD. In the last line, we upper bounded the excess risk induced by noise \(\mathcal{L}(w_{T_b}) - \mathcal{L}(w^*)\) by applying Lipschitzness of the loss.

We now bound \(\|w_T - w_{T_b}\|\).

\begin{theorem}\label{thm:general}

Let $w_T$ be the \(T^{th}\) iterate of noisy GD that optimizes an \(\alpha\)-strongly convex and \(\beta\)-smooth function, and let \(w_{T_b}\) be the \(T^{th}\) iterate of non-noisy GD that optimizes that same function. The "noisy radius" distance, the $\ell_2$-norm between $w_T$ and $w_{T_b}$ at iteration $T$, can be bounded in expectation as follows:

\[ E[\|w_T - w_{T_b}\|] \leq \rho\eta \times \left(\sum_{i=0}^{T-1} c^i\right) = r\]

\end{theorem}

\emph{Proof sketch.}
The full proof is in~\cref{appen:theory}. At each iteration the distance between the noisy iterate and the non-noisy iterate contracts by a factor of \(c = \max(|1 - \eta\alpha|, |1 - \eta\beta|)\) and then increases additively by \(\rho \eta\). The overall distance then can be represented by scaling the additive noise term \(\rho \eta\) by a geometric series that converges. Future work might incorporate additional factors such as momentum acceleration, bias introduced by clipping, or extend our analysis to the setting of more general neural networks. However, our objective here is to provide some theoretical intuition for our algorithm.

Substituting~\cref{thm:general} into the excess empirical risk we get \[R_{\text{noisy}} \leq Lr + R_{\text{non-noisy}}\] where L is the Lipschitz constant, we can see that our private HPO needs to find HPs that are good for non-noisy optimization but do not create a large divergence between the noisy and non-noisy iterates.

\section{Our Private HPO}

We have established a relationship between the excess empirical risk and the noisy radius. We can now connect this back to our goal of doing private HPO, which is to find the HPs that minimize the excess empirical risk. We want to find \(r^* = r(\varepsilon)\), the optimal value of \(r\) for a given value of \(\varepsilon\). We will first \emph{reduce the dimensionality} of the search problem and then \emph{introduce a principled approximation}.

\subsection{Reducing the Dimensionality of HPO}
We want to reduce the dimensionality of HPO so that we can reduce the cost of HPO. 

For fixed \(\varepsilon\), if we increase or decrease \(T\) then we will correspondingly increase or decrease \(\sigma\) by the Composition Theorem of DP. 
The actual statements of DP composition are somewhat complicated, but we can simplify them as saying \(\sigma\) grows slower than \(\alpha T\) for some constant \(\alpha\). Because \(E[\rho] = \sqrt{d} \sigma\), we have that \(\rho\) grows slower than \(T\).

The geometric series converges to \(\dfrac{1}{1-c}\) as \(T\) increases, giving us \(E[\|w_T - w_{T_b}\|] \leq (T \eta) \cdot (\sqrt{d} \dfrac{1}{1-c})\).
Because we are interested in writing the radius in terms of hyperparameters that we can optimize, we drop the second term for simplicity.
Now we can write our hyperparameter of interest as \(r = \eta \times T\), reducing the 2D HPO to 1D. If we wanted to search for additional terms such as the batch size or clipping threshold, we could incorporate them into our theory, but we empirically find that it's best to fix all other HPs to the values we provide and just search for \(\eta, T\).

\subsection{Our Private HPO}
In order to find the optimal \(r^* = r(\varepsilon)\) without exhaustively searching, we need to approximate \(r(\varepsilon)\). A natural choice is Taylor approximation. 
We can sample points from \(r(\varepsilon)\) at different values of \(\varepsilon\) via random search, use this to approximate a Taylor polynomial, and then use that Taylor polynomial to estimate \(r\) for any desired target \(\varepsilon\). After we have our estimated \(r\), we can decompose it into \(\eta, T\) by randomly sampling \(\eta, T\) until their product is close to \(r\). This is the procedure we use in Figure 2, paying for the privacy cost of building the approximation and then using it to estimate the optimal HPs for many values of \(\varepsilon \in [0.5, 8]\). 

We now elaborate on the implementation of the method.

The first-order Taylor approximation of a function \(r(\varepsilon)\) around a point \(\varepsilon_0\) is given by \(r(\varepsilon) \approx r(\varepsilon_0) + \left. \frac{dr}{d\varepsilon} \right|_{\varepsilon=\varepsilon_0} \cdot (\varepsilon - \varepsilon_0)\), which linearly approximates \(r\) near \(\varepsilon_0\). Because we cannot analytically determine \(\frac{dr}{d\varepsilon}\), we will have to approximate this.

To approximate the first-order Taylor polynomial we fit a line. We first use random search to find two empirical points \((\varepsilon_1, r(\varepsilon_1))\) and \((\varepsilon_2, r(\varepsilon_2))\). We then fit a line to these points to obtain the parameters of the line \(m, b\) (slope and intercept). We finally estimate the optimal \(r(\varepsilon_f) = m \varepsilon_f + b\) such that the composition of privacy guarantees for the entire private HPO satisfies a target privacy budget according to Theorem 2.3. In practice we choose smaller values of \(\varepsilon\) for these points such as \(\varepsilon_1=0.1, \varepsilon_2=0.2\), that we find provide a good privacy-utility tradeoff.

More generally, we can approximate the Taylor polynomial by fitting a degree \(N\) polynomial with \(N+1\) points  \((\varepsilon_1, r(\varepsilon_1)) \cdots (\varepsilon_{N+1}, r(\varepsilon_{N+1}))\). We provide results comparing the linear approximation to quadratic approximation in the 2nd common response PDF, but use the linear approximation throughout our work because we find that it provides a good privacy-utility tradeoff.

The full method is detailed in~\cref{alg:dp-raft}. The final privacy guarantee including the cost of HPO is given by~\cref{thm:composition}.
\begin{theorem}\label{thm:composition}
The privacy guarantee of~\cref{alg:dp-raft} in terms of \(\mu\) in \(f\)-DP is \(\mu_t = \sqrt{n\mu_1^{2} + n\mu_2^{2} + \mu_f^{2}}\).
\end{theorem}
The proof and conversion to \((\eps, \delta)\)-DP follow directly from~\citet{dong2022gaussian}, so we defer it to~\cref{appen:theory}. Implementing our method requires decomposing a target \(\eps, \delta\)-DP guarantee into a set of \(\mu\)s; we provide code for this.

\subsection{Limitations}
Although this theory does not hold in general for training neural networks, we quantitatively evaluate the heuristic we develop in Section 3.4 and find that our method holds even for the complex setting of training Transformers on NLP benchmarks. Our HPO also requires more runtime than random search because it is adaptive.

\begin{table*}[htbp]
    \centering
    \caption{Our method fixes six design choices: the architecture and initialization (for CV tasks only), the batch size (full batch), the optimizer (SGD with momentum=0.9), the accounting method (PLV where all prior HPO methods use RDP), and the clipping norm (unit clipping). We report the improvement derived from following each of these techniques with respect to a competitive baseline from prior work on CIFAR100 at $\eps=0.1$.}
    \begin{tabular}{cccc}
    \toprule
    Method & Baseline & Baseline Accuracy & Improvement\\
    \midrule
    Classifier (no bias) & ~\citep{mehtadptransfer} & $71.3$ & $0.36$ \\
    Zero Initialization & Random Initialization~\citep{deepmind} & $64.85$ & $6.81$ \\
    Gradient Descent & SGD(Batch=4096)~\citep{deepmind} & $70.2$ & $1.46$ \\
    Momentum ($\rho=0.9$) & $\rho=0$~\citep{bu2022scalable} & $69.02$ & $2.09$ \\
    PLV Accounting & RDP~\citep{deepmind} & $68.43$ & $3.23$ \\
    Unit Clipping ($C=1$) & $C \ll 1$~\citep{googlefinetune} & $71.2$ & $0.46$ \\
    \bottomrule
    \end{tabular}
    \label{tab:recipe-main}
\end{table*}

\begin{algorithm}
\caption{Model Training Subroutine}
\label{alg:model-train}
\begin{algorithmic}
\STATE Initialize model weights \( w \) at \(0\)
\STATE Decompose \( r \) into \( \eta, T \) without exceeding \(T_{max}\) or \(\eta_{max}\)
\STATE Use the PLD accountant to calibrate \(\sigma\) given \(\mu, T\)
\FOR{\(i = 1, 2, \dots, T\)}
    \STATE Compute gradient with unit clipping and add noise \(\nabla^{(i)} = \frac{1}{|D|} \left(\sum_{i \in D} \textbf{clip}_{1}(\nabla \ell(x_i, w^{(i)})) + \sigma \xi\right)\)
    \STATE Take a step with momentum: \( v^{(i)} \leftarrow \rho \cdot v^{(i-1)} + \nabla^{(i)} \), \( w^{(i)} \leftarrow w^{(i-1)} - \eta v^{(i)} \) 
\ENDFOR
\STATE \textbf{return} trained model \( w \)
\end{algorithmic}
\end{algorithm}

\begin{algorithm}
\caption{Adaptive HPO Routine}
\label{alg:dp-raft}
\begin{algorithmic}
\STATE \textbf{Inputs:} Privacy parameters for hyperparameter sweeps and final run \(\mu_1, \mu_2, \mu_f\), number of runs per sweep \(n\), maximum learning rate \(\eta_{max}\), maximum number of iterations \(T_{max}\), dataset \(D\), model \(M\)
\STATE Perform \(n\) runs with \(\mu_1\) using Hyperparameter Sweep Subroutine (Algorithm~\ref{alg:hparam-sweep}); obtain the best-performing \(r_1\)
\STATE Perform \(n\) runs with \(\mu_2\) using Hyperparameter Sweep Subroutine (Algorithm~\ref{alg:hparam-sweep}), obtain the best-performing \(r_2\)
\STATE Perform linear interpolation to estimate the slope \(\alpha\) and bias \(b\) of the line \( r = \alpha \varepsilon + b \) given \((\mu_1,r_1),(\mu_2, r_2)\)
\STATE Set \( r^* = \alpha \mu_f + b \) given the estimated linear interpolation
\STATE Launch the Model Training Subroutine (Algorithm~\ref{alg:model-train}) with \(r^*, \mu_f\), obtaining the final performance \(A_f\)
\STATE \textbf{Output:} Final performance \(A_f\), trained model \(M\)
\end{algorithmic}
\end{algorithm}

\begin{algorithm}
\caption{Hyperparameter Sweep Subroutine}
\label{alg:hparam-sweep}
\begin{algorithmic}
\STATE \textbf{Inputs:} Privacy parameter \(\mu\), number of runs per sweep \(n\), search space for \(r\)
\FOR{\(i = 1, 2, \dots, n\)}
    \STATE Uniformly sample \( r \) from the search space
    \STATE Launch Model Training Subroutine (Algorithm~\ref{alg:model-train}) with configuration \(r, \mu\), returning performance \(P_i\)
    \STATE \textbf{if} \(P_i\) is the best performance so far on the training set \textbf{then} set best-performing \(r_i = r\)
\ENDFOR
\STATE \textbf{return} best-performing \(r_i\)
\end{algorithmic}
\end{algorithm}

\section{Evaluation}\label{sec:evaluation}
We provide results on a range of image classification, distribution shift, and natural language processing tasks, for both finetuning of models pretrained on public data and training from scratch without any additional data.
Due to the large scope of our evaluation, we defer all experimental details and full results for all datasets and models to~\cref{appen:cv}. We provide ablations on all steps of our method~(\ref{appen:ablations}). We provide hyperparameter grid search results~(\ref{appen:heatmaps}). We also provide the code to reproduce our results at \href{https://anonymous.4open.science/r/dp-custom-32B9/README.md}{this link}.

\textbf{Datasets.}          
Image classification: ImageNet~\citep{imagenet}, CIFAR10 (training from scratch and finetuning), CIFAR100~\citep{krizhevsky2009learning}, FashionMNIST~\citep{fashionmnist}, STL10~\citep{stl10}, EMNIST~\citep{emnist}.
Because these image classification datasets are generally considered in-distribution of the pretraining data, we also provide results on a number of distribution shift datasets~\citep{stanfordwilds} 
. CIFAR10 $\rightarrow$ STL, CIFAR10p1, CIFAR10C, CIFAR100 $\rightarrow$ CIFAR100C~\citep{hendrycks}, Waterbirds~\citep{waterbirds}, FMoW~\citep{fmow}, and Camelyon17~\citep{camelyon}.
For NLP tasks we consider SQuAD~\citep{squad} for Questions Answering, text classification tasks from the GLUE benchmark~\citep{wang2018glue}: SST-2, QNLI, QQP, MNLI(m/mm) and for next word generation we use PersonaChat~\citep{persona} and WikiText-2~\citep{merity2017pointer}, and Enron Emails~\citep{klimt2004enron}.

\subsection{Effectiveness of the Linear Scaling Rule}
\textbf{ImageNet (with Public Data)}
In~\cref{fig:main-imagenet} we compare the performance of our method on ImageNet against the competitive prior works of~\citet{mehtadptransfer, berrada2023unlocking}. Note that these works do not report the privacy cost of HPO and pretrain their models with JFT, Google's proprietary internal dataset; as a result the non-private performance of their models exceeds ours (rightmost points).
Despite this, given sufficient budget \((\eps>0.5)\) we match or exceed their performance \textbf{while accounting for the privacy cost of HPO}. The downside of our method is that for sufficiently small \(\eps\) on sufficiently difficult datasets such as ImageNet, there is no way to keep the privacy cost of HPO small enough to retain enough budget to do a final run, because HP trials with too small a budget do not provide any information.
We provide a deep dive into these points of comparison in~\cref{appen:cv}.

\textbf{CIFAR-10 (without Public Data)}
In~\cref{tab:cifar-comparison-rdp} we compare our method to random search and the grid search baseline, which does not consider the privacy cost of HPO.
We significantly outperform random search, and approach the performance of grid search.
To the best of our knowledge, we are the first to provide competitive performance when training on CIFAR10 without public data under a strict privacy budget while accounting for the privacy cost of HPO.

\begin{table}[h]
    \centering
    \caption{Performance comparison of different methods on CIFAR10. Our method outperforms prior work in linear probing settings when using a feature extractor pretrained on CIFAR100. In the setting where we do not have public data, we compare to random search and grid search and our method greatly outperforms random search.}
    \label{tab:cifar-comparison-rdp}    
    \begin{tabular}{lccc}
        \toprule
        CIFAR10 Acc \textbf{with} Public Data~(\(\eps = 1\)) \\
        \midrule
        \citet{koskela2023practical} & 67\% \\
        \citet{papernothparamtuning} & 66\% \\
        \textbf{Ours} & \textbf{70.5\%} \\
        \midrule
        CIFAR10 Acc \textbf{without} Public Data~(\(\eps = 1\)) \\
        \midrule
        Random Search & 44\% \\
        Grid Search (cost of HPO not incl.) & 68\% \\
        \textbf{Ours} & \textbf{62.63}\% \\
        \bottomrule
    \end{tabular}

\end{table}

\subsection{Comparison to other Private HPO}
We provide a detailed comparison to 5 prior works in private HPO as well as the baselines of random search and grid search, and explain the design choices that enable our method to dominate all prior work.

\subsubsection{Comparison to Renyi HPO}
We compare to three prior works that use Renyi DP to analyze HPO~\citep{papernothparamtuning, koskela2023practical, wang2023dphypo}.

In~\cref{tab:cifar-comparison-rdp} we report that our linear scaling is 3.5\% better on CIFAR10 at \(\eps=1\) in the experimental setting of~\citet{koskela2023practical}: linear probing on a ResNet20 checkpoint pretrained on CIFAR100. ~\citet{koskela2023practical} achieve 67\% on CIFAR10 at \(\eps=1\). In the same setting, the method of~\citet{papernothparamtuning} obtains 66\%. We apply the linear scaling rule in the same setting, so that only the hyperparameters our method selects are different, and obtain \(70.5\%\) at \(\eps=1\). All methods use the same hyperparameter search space.
The reason our method outperforms~\citet{koskela2023practical, papernothparamtuning} is because our prior is better than their random search, which is required by their method, enabling us to simultaneously allocate a smaller portion of the privacy budget to HPO while still finding better hyperparameters. We also use PLD accounting which is tighter than the RDP accounting their method requires. Neither of these can be fixed; that is, we cannot modify their method to integrate the linear scaling prior or to use PLD accounting. Even with PLD accounting, we would not be able to make up for the gap in accuracy that comes from our adaptive method. An interesting question for future work is whether we can do RDP analysis of our adaptive method. More details in~\cref{appen:subsample}.

\subsubsection{Comparison to parameter-free methods.}
A related area is parameter-free HPO, that builds optimizers that do not require specifying the learning rate as a hyperparameter. In general it can be challenging to apply these parameter-free methods to DP, because the update rule for the scale of the gradient may not maintain its guarantees in the presence of noise~\citep{li2023differentially}. 

\textbf{Comparison to DPAdamWosm.}
One parameter-free optimizer specifically designed for DP is DPAdamWOSM~\citep{mohapatra2021role}.
On ImageNet DPAdamWOSM achieves 79\% at \(\eps=1)\), which is 8\% lower than our method (87\% at \(\eps=1\)). We do not find that the data-independent learning rate selection works well for ImageNet, and still requires tuning the number of iterations (see~\cref{appen:wosm}).

\textbf{Comparison to~\citet{mehtadptransfer}}
~\citet{mehtadptransfer} propose an approach where they fix the batch size to full batch, the number of steps to 1, and take a single step of DP-Adam with a very small learning rate. Their approach obtains just \(81\%\) at \(\eps \geq 1\) on ImageNet for a model whose non-private accuracy is \(88.7\%\), because they take only a single step. Our method smoothly interpolates between the low-r setting for small \(\eps\) and the large-r setting for large \(\eps\), and outperforms their method across all privacy budgets.

\subsubsection{Comparison to baselines: random search and grid search.}

\textbf{Linear scaling significantly outperforms random search.}
In~\cref{tab:randomsearch-comparison} we report the performance for random search, our method, the oracle, and the relative error rate reduction (RERR). Across all datasets, our method significantly outperforms random search. 
We use the same logarithmic grid for both our method and random search that can be found in~\cref{appen:cv}. We vary this grid and find that the larger the search space, the more our method outperforms random search. 

\begin{table}
    \centering
        \caption{Comparing various HPO methods on CIFAR10 (without public data), CIFAR100, ImageNet, and SQuAD (with public data). Reported numbers are mean over 5 trials.}
    \label{tab:randomsearch-comparison}
    \begin{tabular}{c|cccc}
    \toprule
         Dataset&  Random Search&  Oracle&  Ours& RERR\\
         \midrule
         CIFAR10&  44&  68&  62.63& 77.63\\
         CIFAR100&  84.44&  89.62&  89.10& 84.85\\
         ImageNet&  81.2&  88.6&  86.7 & 73.97\\
 SQuAD& 49.33& 82.43& 78.08&86.85\\
 \bottomrule
    \end{tabular}
\end{table}

\begin{figure}[htbp]
    \centering
    \subfigure[
    ]{
    \begin{minipage}[t]{0.46\linewidth}
    \centering
    \includegraphics[width=\linewidth]{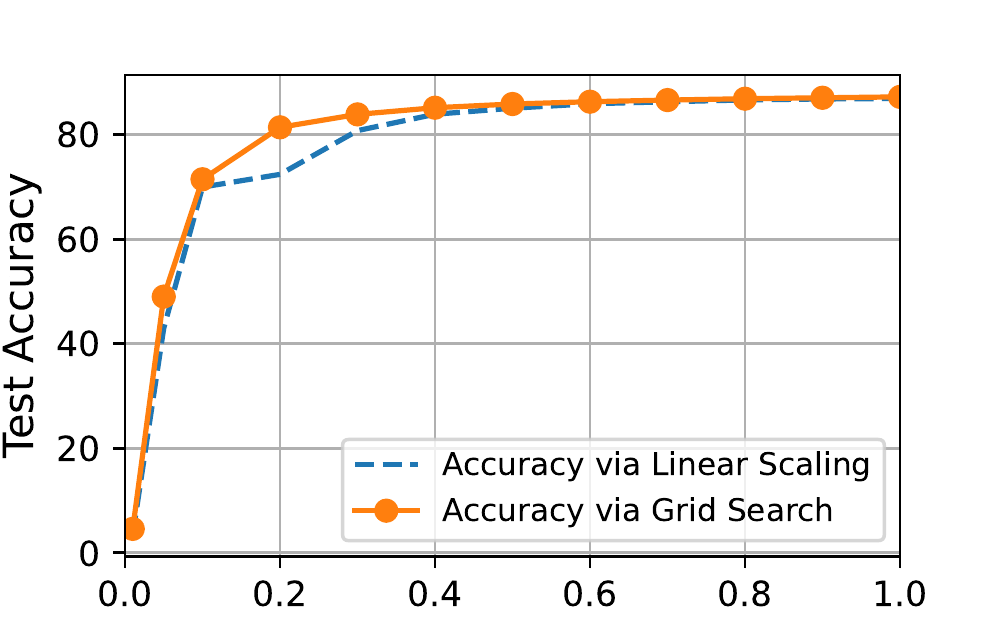}
    \end{minipage}
    }
        \subfigure[
        ]{
    \begin{minipage}[t]{0.46\linewidth}
    \centering
    \includegraphics[width=\linewidth]{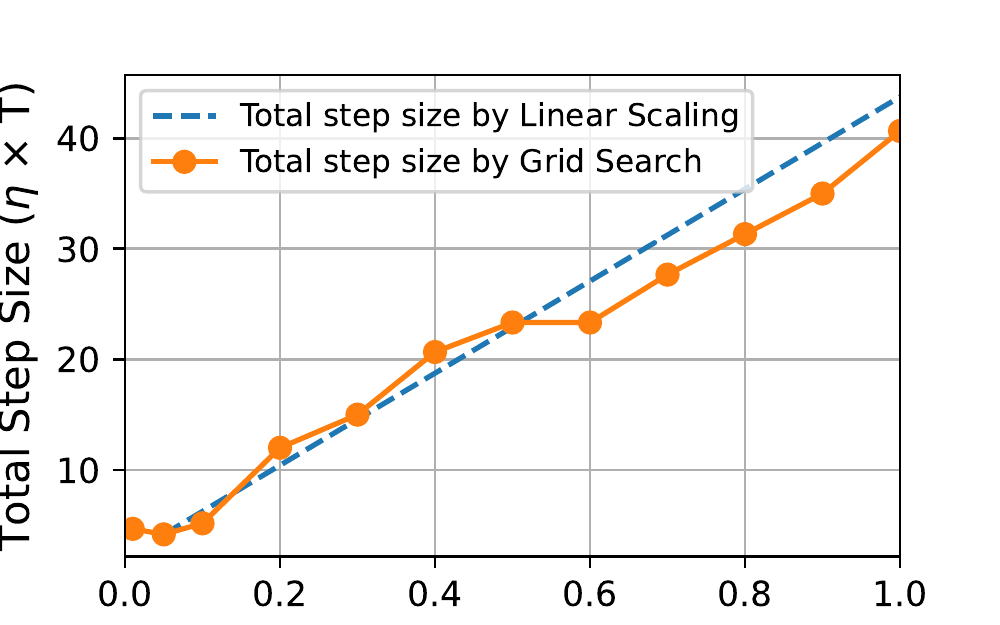}
    \end{minipage}
    }
    \caption{Training the beit architecture on CIFAR100, the linear scaling rule produces values for $r = \eta \times T$ close to that of grid search, and the performance drop is only apparent at $\varepsilon>0.2$ because of the cost of HPO, and vanishingly small for larger $\varepsilon$.
    } 
\label{fig:pareto}
\end{figure}
\textbf{Linear Scaling approaches grid search.}
We validate the effectiveness of linear scaling against the grid search baseline.
In Fig. \ref{fig:pareto} (right) we compare Alg.~\ref{alg:dp-raft} to the best run across 100 trials from the search space.
The privacy cost of grid search is many times higher than that of our method at each value of \(\eps\), because we do not account for the privacy cost of grid search to illustrate that even when our method has to account for the privacy cost of HPO and the oracle (grid search) does not, our method is competitive.
Our method finds near-optimal hyperparameters with just a fraction of the runtime and privacy cost of grid search.

\subsection{Empirical Analysis of Linear Scaling}
We now consider different architectures and validate our HPO method in the presence of distribution shifts. Full results can be found in~\cref{appen:analysis}.

\begin{table}
    \centering
    \caption{We compare the best private and best non-private performances of all models on all datasets. We use the linear scaling rule to scale hyperparameters from $\varepsilon=0.1$ to $\varepsilon=1$, so our privacy analysis includes the cost of hyperparameter tuning.}
    \label{tab:main-cv-results}
\centering
\small
\begin{tabular}{@{}lccccc@{}}
\toprule
Model & Dataset & $\varepsilon=1$ & $\varepsilon=\infty$ & Gap \\
\midrule
beitv2 & CIFAR10 & $98.90$ & $99.00$ & $ 0.10$ \\
 & CIFAR100 & $89.10$ & $91.57$ & $2.47$ \\
 & FMNIST & $91.02$ & $91.53$ & $0.51$ \\
 & STL10 & $99.69$ & $99.81$ & $0.12$ \\
  & EMNIST & $81.77$ & $82.00$ & $0.23$ \\
\hline 
convnext & CIFAR10 & $96.75$ & $97.22$ & $0.47$ \\
 & CIFAR100 & $83.47$ & $86.59$ & $3.12$ \\
 & FMNIST & $90.23$ & $91.13$ & $0.9$ \\
 & STL10 & $99.61$ & $99.71$ & $0.10$ \\
   & EMNIST & $78.38$ & $79.05$ & $0.67$ \\
\hline
beit & CIFAR10 & $98.19$ & $98.51$ & $0.32$ \\
 & CIFAR100 & $87.1$ & $90.08$ & $2.98$ \\
 & FMNIST & $90.55$ & $91.6$ & $1.05$ \\
 & STL10 & $99.62$ & $99.78$ & $0.16$ \\
 & EMNIST & $81.48$ & $83.25$ & $1.77$ \\ 
\hline 
vit-L & CIFAR10 & $98.29$ & $98.44$ & $0.40$ \\
 & CIFAR100 & $86.18$ & $89.72$ & $3.54$ \\
 & FMNIST & $90.58$ & $91.37$ & $0.79$ \\
 & STL10 & $99.62$ & $99.76$ & $0.14$ \\
\bottomrule
\end{tabular}
\end{table}

\textbf{Architecture Search.}
In~\cref{tab:main-cv-results} we apply our method to different architectures that can serve as good backbones for high-accuracy DP classification across CIFAR10, CIFAR100, FMNIST, STL10, and EMNIST.
The private-non private utility gap diminishes with model accuracy.
One architecture, beitv2, performs the best on all benchmarks and also has the highest non-private zero-shot ImageNet accuracy~\citep{rw2019timm}.
We conclude that architecture search can be done without any privacy cost by selecting the model with the best zero-shot performance on a representative benchmark such as ImageNet.

\begin{table*}
    \caption{Evaluating our DP-HPO method on datasets with distribution shifts at \(\eps=1\).}
    \label{tab:main-dro-results}
    \begin{tabular}{@{}lccccccc@{}}
\toprule
 & Waterbirds & fMoW & Camelyon & C10 $\rightarrow$ STL & C10 $\rightarrow$ C10p1 & C10 $\rightarrow$ C10C & C100 $\rightarrow$ C100C \\
\midrule
ID & 92.31 & 45.44 & 93.91 & 98.90 & 98.90 & 98.90 & 89.65 \\
OOD & 91.59 & 35.31 & 93.55 & 98.82 & 97.85 & 89.98 & 68.69 \\
\bottomrule
\end{tabular}
\end{table*}

\textbf{Distribution Shift.}
A concern in DP fine-tuning is that the pretraining datasets are too similar to the downstream tasks, which can violate privacy~\citep{tramer2022considerations}. 
In~\cref{tab:main-dro-results} we evaluate the robustness to distribution shift of models trained with our private HPO to non-private models, in the absence of any explicit regularization methods or any information about the distribution shift. 
These datasets are considered benchmark tasks for distribution shifts~\citep{kumarfinetunedistort,kumarfreezefinetune, raghavwaterbirds} and include data that is not in-distribution of the training data, making for a more realistic evaluation of the capabilities of our method to solve challenging tasks.
We show that DP-SGD provides robustness to covariate, subpopulation and label distribution shifts for synthetic and natural datasets. Full details in~\cref{appen:ood}.

On Waterbirds, DP degrades the ID performance but actually improves the OOD performance. On fMoW and Camelyon17 that are datasets with a significant distribution shift from ImageNet and very different subgroups, DP does not significantly degrade performance and does not exacerbate disparities among subgroups. We also show that we can train models on CIFAR10 with DP and do zero-shot transfer to STL and CIFAR10p1. Finally, we evaluate the robustness of CIFAR-DP-trained models to the common corruptions of CIFAR10C/CIFAR100C, and note that DP training achieves some measure of intrinsic robustness to image corruptions.

\subsection{Linear Scaling for language modeling}\label{sec:nlp}
\begin{table}
\caption{Linear scaling holds for GLUE tasks when training the full RoBERTa-base model}
\centering
\begin{small}
\begin{tabular}{lccc}
\toprule
Task & $\varepsilon$ & Acc & $r=\eta \times T$ \\
\midrule
\multirow{3}{*}{SST-2} & 0.1 & 90.60 & 0.975 \\
                        & 0.2 & 90.83 & 1.95 \\
                        & 0.7 & 91.06 & 5.07 \\%9.75??? \\
\midrule
\multirow{3}{*}{QNLI} & 0.1 & 82.54 &3.9 \\
                        & 0.2 & 84.00 &4.68 \\
                        & 1.3 &86.25 &26.52 \\
\midrule
\multirow{3}{*}{QQP} & 0.1 & 81.07 &11.7\\
                        & 0.2 & 82.21 &17.55 \\
                        & 1.2 &84.69 &64.35 \\
\midrule
\multirow{3}{*}{MNLI(m/mm)} & 0.1 & 77.52(78.24) &11.7 \\
                        & 0.2 & 79.40(79.98) &17.55 \\
                        & 1.2 &81.86(82.76) &64.35 \\                        
\bottomrule
\end{tabular}
\end{small}
\label{tab:glue}
\end{table}

Prior work has generally focused on either CV or NLP because the methods used in DP fine-tuning differ greatly across data modalities~\citep{xuechendpnlp, googlefinetune}; here we show that our method extends to NLP by validating on text classification and language modeling tasks with LoRA~\citep{hu2021lora} and full fine-tuning.
We fine-tune GPT-2~\citep{radford2019language} with our method for three language modeling tasks that have been benchmarked in prior works~\citep{xuechendpnlp,shi2022just,gupta2022recovering} on private fine-tuning: Persona-Chat~\citep{zhangchitchat}, WikiText-2~\citep{merity2017pointer} and Enron Emails~\citep{klimt2004enron}. 
We also fine-tune RoBERTa-base on four tasks in the GLUE benchmark: SST-2, QNLI, QQP and MNLI(m/mm) in Table~\ref{tab:glue}.
While prior works mainly focus on $\varepsilon$ in $\{3,8\}$, in this work we are also interested in smaller $\varepsilon$s like $0.1$. 
\cref{appen:hyperparam} includes the details for the experimental set-up.

\textbf{Linear scaling succeeds when random search fails.}
We consider the challenging setting from~\citet{malladi2024finetuning} of fine-tuning an OPT-13B model on just 1000 samples from SQuADv2 with DP-SGD-LoRA. Random search runs sometimes do not improve much over zero shot performance, because the search space is so large and the viable set so small. In the initial phases of our method, the trials do not always succeed. Regardless, our method achieves \(78\% \pm 3\%\) close to the oracle \(82\%\), a RERR of \(87.5\%\). Random search performs poorly for NLP tasks because HPO is generally more challenging~\citep{xuechendpnlp}. In the rest of the NLP datasets we consider, we compare our performance to prior work that doesn't consider the privacy cost of HPO. 

\begin{figure}[htbp]
     \centering
     \subfigure[Pareto Frontier for $\varepsilon$ vs Test Perplexity.]{
         \begin{minipage}[t]{0.46\linewidth}
         \centering
         \includegraphics[width=\linewidth]{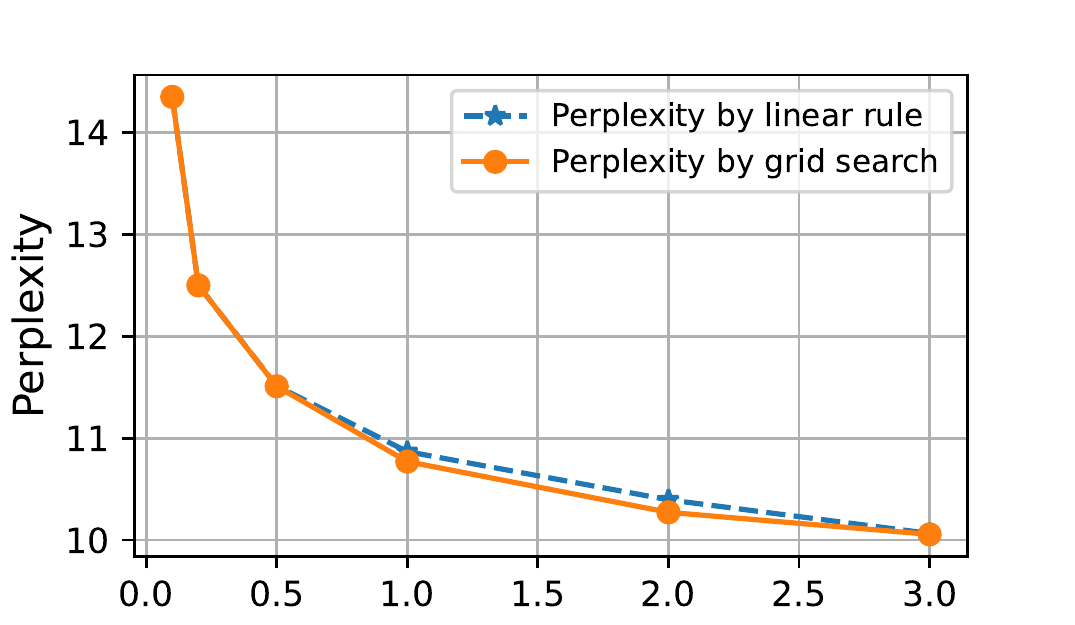}
         \end{minipage}
     }
     \hfill
     \subfigure[Pareto Frontier for $\varepsilon$ vs Total Step Size.]{
         \begin{minipage}[t]{0.46\linewidth}
         \centering
         \includegraphics[width=\linewidth]{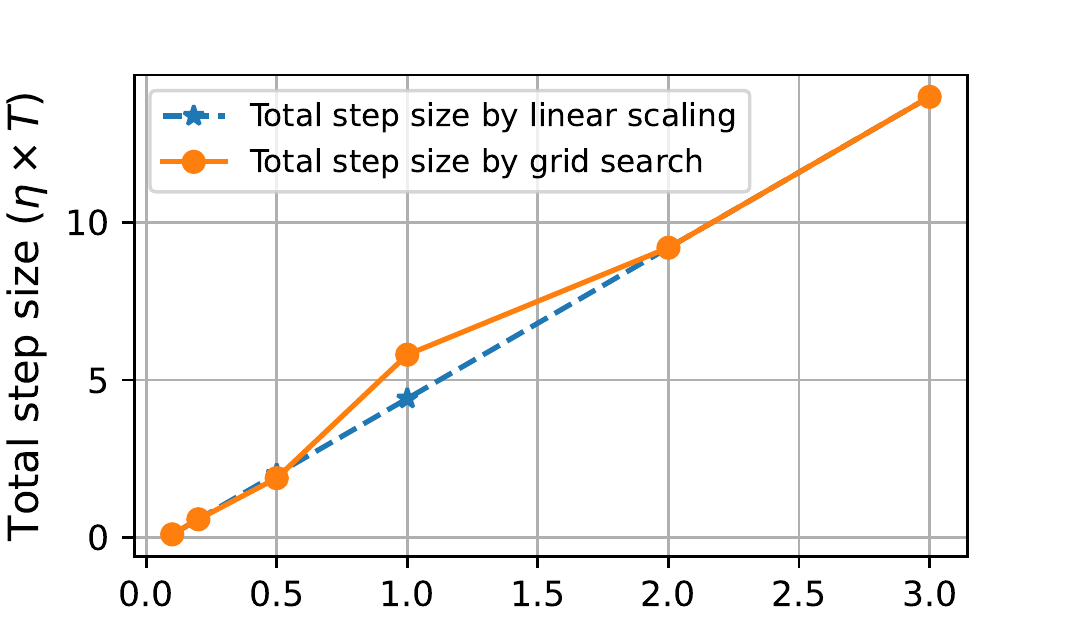}
         \end{minipage}
     }
     \caption{The linear scaling rule (accounting for the privacy cost of hyperparameter tuning) is competitive with grid search (non-private, doing N trials each with the given $\varepsilon$) on the Enron Emails dataset. Left: y-axis is Perplexity (lower is better).}
\label{fig:enron}
\end{figure}

\textbf{Linear scaling holds for NLP tasks}
We analyze the performance gap between estimated total step size and optimal total step size by grid search to understand how well linear scaling performs on language modeling tasks. 
Fig.~\ref{fig:enron} plots the optimal perplexity and perplexity by estimated total step size at different values of $\varepsilon$ on Enron emails.
We can see that the linear scaling rule generalizes well for reported values of $\varepsilon$ and the perplexity by the estimated total step size is close to the optimal perplexity. 
From Table~\ref{tab:glue} we can see that our method also works for the GLUE benchmark.

\textbf{The linear scaling rule outperforms prior results on differentially private language modeling tasks.} 
We first run a qualitative evaluation on the previous benchmark SOTA~\citep{xuechendpnlp} on PersonaChat trained with DP-SGD by following the linear scaling rule to increase the number of epochs. 
    \begin{table}
        \centering
    \caption{Linear scaling holds when fine-tuning all layers of GPT2 on PersonaChat and outperforms ~\citet{xuechendpnlp}}%
    \begin{small}
    \begin{tabular}{cccccc}
    \toprule
    $\varepsilon$ ($\delta=\frac{1}{2|D_{\text{train}}|}$) &$0.7$ & $3$ &$\infty$\\
    \citet{xuechendpnlp} &- &$24.59$  &$18.52$\\
    Our Work &$21.25$ & - &$17.69$\\
    \bottomrule
    \end{tabular}\label{tab:chitchat}
    \end{small}
\end{table}
We can see in Table~\ref{tab:chitchat} that we can push the perplexity under $18$ for $\varepsilon=3$ and $\varepsilon=8$; this performance is competitive with the non-private baseline. 
Furthermore, even when pushing for a stricter privacy guarantee $\varepsilon=0.5$, we can still get perplexity of $21.25$, that is better than the result of $\varepsilon=8$ in~\citep{xuechendpnlp}. 
We also report the results of ablating these hyper-parameters and varying the number of layers trained in \cref{appen:chat}.
    \begin{table}
        \centering
    \setlength\tabcolsep{3pt}
    \caption{Finetuning GPT-2 on WikiText-2 ($\delta=10^{-6}$) 
    and Enron ($\delta=\frac{1}{2|D_{\text{train}}|}$) with DP-SGD. 
    Ppl is perplexity and TSS is Total Step Size. ($^*$ means estimated). Previously reported best perplexity of GPT-2 on WikiText-2 at $\varepsilon=3$ is 28.84 in \citep{shi2022just}.}
    \begin{small}
    
    \begin{tabular}{ccccccccc}
        \toprule
        &$\varepsilon$&$0.1$ &$0.2$ &$0.5$&$1.4$&$2.2$&$3.0$  \\
        \cmidrule{2-8}
        WikiText-2 &Ppl &-&$28.81$&$28.37$&$28.15$&$27.98$ &$27.69$\\
        &TSS &- &$0.008$ &$0.02$ &$0.04^*$&$0.08^*$&$0.12^*$ \\
        \midrule
        &$\varepsilon$ &$0.1$&$0.2$ &$0.7$&$1.1$&$1.9$&$2.7$  \\
        \cmidrule{2-8}
        Enron&Ppl &$14.35$&$12.50$ &$11.56$ &$10.91$&$10.45$ &$10.14$\\
        &TSS &$0.10$ &$0.58$ &$2.02^*$ &$4.41^*$&$9.19^*$&$13.98^*$ \\        
        \bottomrule
    \end{tabular}\label{tab:enron}
    \end{small}
    \end{table}
We quantitatively validate the linear scaling rule on WikiText-2 
and report the result in Table~\ref{tab:enron}. 
For WikiText-2, a key observation is that when we compare our results to the best prior reported results in~\citep{shi2022just}, for the same number of passes over the training data (20), we obtain lower perplexity for $\varepsilon=0.2$ than they report for $\varepsilon=3$.
That is, by just increasing the effective step size from $\sim 8\times{10}^{-6}$ to  $\sim 8\times{10}^{-3}$ we can strengthen the privacy guarantee without degrading performance.

\subsection{Additional Ablations}

We can apply our method to methods other than DP-SGD.~\citet{tang2024private} recently proposed DP-ZO, a method for DP zeroth-order optimization, that privatizes the zeroth-order update to the model weights in the form of a scalar rather than the first-order gradient update. In~\cref{tab:dpzo} we find that our method can also optimize HPs for DP-ZO.

\begin{table}
    \centering
    \begin{tabular}{lc}
    \toprule
         Method & Mean Accuracy (Std) \\
         \midrule
         Random Search & 82.53 (1.01) \\
         Our Private HPO & 83.02 (0.86) \\
         Grid Search & 83.87 (0.50) \\
         \bottomrule
    \end{tabular}
    \caption{Our method works beyond DP-SGD.}
    \label{tab:dpzo}
\end{table}

Although our method fits a \(1\)-d polynomial with \(2\) points, we can in principle fit any degree-d polynomial with \(d+1\) points. to approximate the relationship between \(r\) and \(\eps\). However, because using more points to fit the polynomial imposes more privacy cost for HPO, we use the same number of points and degree throughout all experiments. It is likely that for some datasets, it's important to tune the privacy budget allocated to the smaller trials, the number of points, etc. However, we are not interested in tuning the hyperparameters of our hyperparameter optimization method.

In~\cref{tab:nonlinear} we evaluate the linear fit with 2 and 3 points for evaluation, and the quadratic fit with 3 points for evaluation, on ImageNet.

\begin{table}
    \centering
    \begin{tabular}{lcc}
    \toprule
         Method & Mean Accuracy (Std) \\
         \midrule
         Linear (2) & 86.69 (0.86) \\
         Linear (3) & 87.81 (0.86) \\
         Quadratic (3) & 86.64 (1.08) \\
         \bottomrule
    \end{tabular}
    \caption{Using more points or a higher order approximation can improve performance.}
    \label{tab:nonlinear}
\end{table}

Throughout the paper we search for the two main HPs of interest \(\eta, T\) and fix other HPs such as batch size \(B\) and clipping threshold \(C\). However, we can search for these as well. We can incorporate new HPs by updating the decomposition of \(r\) so that we optimize for the joint product of the HPs being optimized. We change it from \(r = \eta \times T\) to \(r = \eta \times T \times B \times C\). We evaluate on CIFAR100. The performance of our method when optimizing \(\eta, T, B, C\) is \(87.9 \pm 1.9\), which is worse than the \(89.10\) where we optimized \(\eta, T\) and fixed \(B,C\).

One limitation of our method is the runtime, shown in~\cref{tab:runtime}.. We have worse runtime than random search and worse parallelization than grid search, which is embarrassingly parallel while our method requires serial runs.

\begin{table}
    \centering
    \begin{tabular}{ccc}
    \toprule
         Method&  GPU Hours& Wall-clock time (hours)\\
         \midrule
         Random&  1& 1\\
 Grid& 100&1\\
         Ours&  7& 3\\
         \bottomrule
    \end{tabular}
    \caption{Our method trades off runtime for performance with random search and grid search.}
    \label{tab:runtime}
\end{table}

Here the base time for a single HP trial is just 1 hour; this can change based on the task, but these proportions should remain similar. Random search just does 1 run so it has both the lowest GPU hours and wall-clock time. The oracle does a number of runs equal to the granularity of the search space, which here we approximate as 100. In the setting where 100 GPUs are available for the oracle, which may be realistic for large companies but is not realistic for our academic compute setting, these can all be done in parallel, so it uses 100 GPU hours but just 1 hour in wall-clock time. Our method typically does 7 runs: 3 for \(\eps_1\), 3 for \(\eps_2\), and 1 for 
\(\eps_f\), so the total number of GPU hours is 7. The serial dependency of our method requires that \(\eps_f\) runs after \(\eps_1\) and \(\eps_2\), but the 3 runs for \(\eps_1, \eps_2\) can be parallelized so the wall-clock time is just 3 hours.

\section{Related Work and Discussion}
\textbf{Related Work.}
We have performed detailed quantitative and qualitative comparisons to prior private HPO methods~\citep{papernothparamtuning, koskela2023practical, wang2023dphypo}. These build on earlier work by~\citet{liu2018private} that can do HPO by increasing the privacy cost roughly threefold. We improve over these works by significantly reducing the privacy budget required by HPO and adopting a robust prior on our hyperparameter search. Many non-private HPO methods have been used by prior DP papers that do not report the privacy cost of HPO, and a valuable future task would be to consider privatizing these.
Multiple prior works~\citep{deepmind, papernotfinetuning, bu2022scalable, bu2022differentially, googlefinetune, mehtadptransfer, berrada2023unlocking, xuechendpnlp, xuechendimensionalityreduction, hu2021lora} consider the task of maximizing the privacy utility tradeoff of finetuning pretrained models. Although the main focus of our paper is private HPO, we also critically evaluate the efficacy of a range of techniques that have been proposed by these works such as data augmentation, fine-tuning the embedding layer, and weight averaging. A detailed discussion of these techniques is deferred to~\cref{appen:ablations} and~\cref{appen:lm}.
~\citet{golatkar2022mixed, nasr2023effectively, amid2021public} treat $<10\%$ of the private training dataset and public and use it to improve DP-SGD.
Although we do not use any private data during pretraining, future work can tackle applying our method to this alternate threat model.
~\citet{sanderdptanhparam} suggest doing HPO with smaller batch sizes and then scaling up the HPs to the full batch update. This idea is similar in spirit to the adaptive scaling that we propose, because the HP trials are cheaper from a runtime perspective than the final run. However, our approach is not only compute-efficient but also accounts for the privacy cost of HPO. ~\citet{kuo2023noisy} find that noisy HPO in the federated setting suffers, and suggest doing HPO on public proxy data (whose existence we don't assume) and transferring it to the private dataset. 

~\citet{wang2023dphypo} propose a method for private adaptive HPO that provides an RDP guarantee for Bayesian HPO. They compare their method to random search under a total privacy budget of \(\eps=15\), where at each iteration they sample a new set of HPs from their prior, and update their prior, and at each iteration random search samples a new set of HPs uniformly from the search space; each run has a base privacy cost, and it takes many runs for the distribution to converge. Their method can be seen as a version of ours that lacks the linear scaling prior and does not use cheap trials to find the parameters for the prior. As a result, they use much more budget in order to find a good distribution for the HPs. This more flexible approach can be superior to ours in settings where the HPs are not linear. 

\textbf{Discussion.} 
DP researchers commonly confront the compute-intensive, privacy-expensive task of doing grid search with hundreds of trials to optimize the privacy-utility tradeoff of DP methods. Our work provides an alternative HPO method that reduces the compute and privacy costs of grid search by an order of magnitude without compromising accuracy across 20 tasks. Researchers using our method can accelerate the pace of research by reducing the compute needed to produce good results, and address the open question of accounting for the privacy cost of hyperparameter tuning, whether they are doing transfer learning in the presence of domain shifts, training from scratch, or applying PEFT methods to LLMs.

\section{Impact Statement}
This paper presents work whose goal is to improve the quality of models trained with differential privacy. Privacy is at the heart of many ongoing debates about AI. We submit that any work that makes it easier for organizations to train and release models with DP guarantees will positively benefit society.
Additionally, this paper presents work whose goal is to advance the field of Machine Learning. There are many potential societal consequences of our work, none which we feel must be specifically highlighted here. 

\bibliography{bib}
\bibliographystyle{icml2024}

\appendix
\pagebreak
\onecolumn
\section{Further Results for Computer Vision Tasks}\label{appen:cv}
Our code is available at the following URL: https://anonymous.4open.science/r/dp-custom-32B9/README.md
\subsection{Experimental Set-up}
\paragraph{Hyperparameter Search Space.}
We use a logarithmic grid for the learning rate \(\eta\ \in [10^{-7}, 10^{-4}]\). We use the same grid for the CIFAR training from scratch experiments, and the NLP experiments. We scale the learning rate by the batch size (the original linear scaling rule). The number of epochs depends on the maximum number of iterations that we can do in the provided time.

\paragraph{ImageNet details.}
The architecture is a modernized ViT~\citep{fang2023eva02} pretrained on IN-21k~\citet{imagenet} with CLIP. We use a resource-efficient finetuning approach where we create a linear layer aggregating the intermediate representations from each Transformer block, following~\citet{tang2023differentially}. 
We apply the method from~\citet{sun2023importance} to preprocess the features, allocating a budget of \(\eps=0.05\) for the private mean estimation.
For the HPO, we do 3 runs at \(\eps=0.1\), followed by 3 runs at \(\eps=0.2\) and a final run at \(\eps=0.88\), which produces a cumulative privacy cost including HPO of \(\eps=1.0\). Here we express the privacy values in terms of \(\eps\) for brevity, but the actual expressions are in terms of \(\mu\), the parameter for f-DP. Because of the nature of composition, the hyperparameter search only costs us the difference in performance between \(\eps=0.88\) and \(\eps=1.0\), which is minimal.
We search across values of T ranging from 1 (a single epoch) to 20 (the most we can do in the maximum amount of time a job will run on our cluster).
~\citep{berrada2023unlocking} report that fine-tuning the full architecture produces better results than linear probing. However, we lack the computational resources to do full fine-tuning of large transformers, but we can do linear probing of the extracted features in under an hour on a single A100. 

\paragraph{CIFAR training from scratch.}
We use the model from~\citet{tang2023differentially}, a WideResNet-16-4, and train only the last layer on the extracted features from previous layers. The model is pretrained on synthetic data that does not resemble real-world data. We choose this model because it is the SOTA model for CIFAR training from scratch, and we want to validate that our private HPO can produce competitive results in a setting where the zero-shot performance is poor (indeed, the zero-shot performance of this model is just random chance, because it has never seen any real images before) but the ceiling for performance is quite high. 

\paragraph{Comparison to DPAdamWOSM details.}~\label{appen:wosm}
We implement DPAdamWOSM~\citep{mohapatra2021role} to the best of our ability in wosm\_impl.py since there is no code available, and tune the necessary hyperparameter T (\# of epochs) between 1 and 200 and report the performance for the best value of T without accounting for the privacy cost of this tuning. The rest of the hyperparameter choices and model architecture mirror our own. 
At a high level, our linear scaling rule attempts to do a data-dependent learning rate selection, while DPAdamWOSM does a data-independent learning rate selection. It is natural that for hard tasks (ImageNet) the data-independent choice may not work well. We note that while DPAdamWOSM does not require tuning the learning rate, we still need to tune the number of epochs. Therefore, even if further tuning for DPAdamWOSM could match the utility of the linear scaling rule, it would not match the privacy guarantee. Ultimately we think these works are compatible, because we can use our HPO to tune the number of epochs in DPAdamWOSM.

\paragraph{Comparison to ~\citet{koskela2023practical} details.}~\label{appen:subsample}.
We implement the ability to train on a subset of ImageNet in our codebase by passing the flags start\_idx, end\_idx. As an initial test, we tried doing HPO on half the dataset by passing start\_idx=0, end\_idx=625. Our code will produce a random permutation of the chunks of ImageNet (1251 total) and then load the first half. We compare \(\eta=0.01, \eta=1.0\) on this half-dataset. On the full dataset, these produce very different performance; \(\eta=0.01\) achieves \(81\%\) at \(\eps=1.0\), while \(\eta=1.0\) achieves \(87\%\). However, on the half dataset, the first learning rate achieves \(43.2\%\) performance, while the second achieves \(41.4\%\). Inspecting the loss curves, we find that both learning rates overfit the training dataset and do not generalize to the validation set, but the second learning rate overfits more. We then try training two models on disjoint sets of the dataset and combining them via parallel composition. This achieves \(83\%\), which is worse than training on the entire dataset. This may be an interesting direction for future work. We tried a number of other strategies to try and scale the idea of~\citet{koskela2023practical} to ImageNet scale, such as weight decay, smaller learning rate, single-epoch training, etc. but were unable to produce a recipe where performance on the half-dataset was consistently positively correlated with performance on the full dataset. We suspect that there is a factor of the number of classes that is needed to properly calibrate the subsampling.

\paragraph{Models.}\label{appen:methods}
We evaluate five models: two masked-image modeling transformers, beit~\citep{beit} and beitv2~\citep{beitv2}, their backbone architecture ViT~\citep{vit} at both the base and large scales, and the pure convolutional architecture convnext~\citep{convnext}.
All models are pretrained on ImageNet-21k~\citep{imagenet}.
These models span a range of input resolutions: beitv2 (224x224), convnext, vit-base, vit-large (384x384), and beit (512x512) and we upsample images to the necessary input size.
For text generation we use GPT-2~\citep{radford2019language} at the smallest scale, and RoBERTa-base.

\paragraph{Availability.}
Our results tune open source models from the PyTorch timm package~\citep{rw2019timm} using existing privacy accounting from~\citep{glwmsaccountant} and per-sample clipping code in~\citep{opacus}, and can be reproduced in minutes.

\section{Further ImageNet Results.}
We perform additional experiments on ImageNet with the same architecture as prior work to better understand the tradeoffs of our method.
We use a ViT-g that was pretrained on laion-2b, to compare to the ViT-g models in prior work that were pretrained on JFT-4b.
It is trivial that linear scaling outperforms a naive grid search, but we also compare the effectiveness of linear scaling against the hyperparameter selection strategies used in prior work~\citep{googlefinetune}.
We find that use of our new rule can unlock significant improvements for a range of $\varepsilon$ when we hold both approaches accountable for the privacy cost of hyperparameter tuning.
We apply linear scaling to the ViT model used in ~\citep{googlefinetune} on CIFAR100.
Although~\citep{googlefinetune} do not directly state the hyperparameters for their best results, they specify that they use 200 hyperparameter trials with Bayesian optimization.
While they obtain RDP guarantees, these guarantees do not include the privacy cost of non-privately tuning hyperparameters.
We apply the linear scaling rule to extrapolate a value of $r$ from $\varepsilon=0.1$ to $\varepsilon=1$, obtaining $r=20=\eta (0.2) \times T (100)$.
\textit{We recover performance of $82.7 \%$ for $\varepsilon=1$, a $2 \%$ improvement over the best result for DP-Adam in~\citep{googlefinetune} while accounting for the privacy cost of hyperparameter tuning.}
They obtain their best result for DP-Adam at $T=10$, but we cannot compute the corresponding value of $r$ because they do not provide $\eta$.
However, because they use a clipping norm of $0.005$ we can reasonably infer that their value of $r$ is $\approx 1000 \times$ smaller than ours.
This is farther from the optimal non-private training, as evidenced by the performance gap.

\paragraph{Linear Scaling scales to ImageNet}
\begin{table}
    \centering
        \caption{Linear Scaling on ImageNet is competitive with ~\citep{mehtadptransfer} and~\citep{googlefinetune}}
\centering
\begin{tabular}{lcccc}
\toprule
$\varepsilon$ & ~\citep{mehtadptransfer} & ~\citep{googlefinetune} & Ours & $r=\eta \times T$ \\
\midrule
0.25 & 75.6 & - & 79.0 & 250 \\
0.50 & 79.4 & 86.1 & 81.6 & 750\\
1.00 & 81.1 & 86.8 & 83.2 & 1100\\
2.00 & 81.5 & 87.4 & 84.2 & 2000\\
10.0 & 81.7 & - & 85.4 & 2000\\
$\infty$ & 86.9 & 88.9 & 85.7 & 2000\\
\bottomrule
\end{tabular}
\label{tab:imagenet}
\end{table}
In Table~\ref{tab:imagenet} we do a granular comparison between our method and~\citep{mehtadptransfer, googlefinetune}.
We observe that our method is competitive with~\citep{googlefinetune} even when accounting for the privacy cost of hyperparameter search, and that the linear scaling rule holds up at the scale of ImageNet for very large values of $r = \eta \times T$.
The non-private accuracy of their closed-source model is $3.2\%$ higher than our open-source model, and so the private accuracy at $\varepsilon=2$ is also $3.2\%$ higher.

However, ultimately our method and the method of~\citet{googlefinetune} are complementary, because their method introduces new hyperparameters that we intuit our linear scaling rule can optimize.
We attempted to validate this intuition empirically but were unable to reproduce the results of~\citet{googlefinetune} because they and~\citet{mehtadptransfer} pretrain on the closed-source JFT dataset with billions of images.
We note that all numbers we report for models pretrained on ImageNet-21k using first-order methods surpass those in~\citep{googlefinetune}, but for sufficiently small values of $\varepsilon$ on harder datasets the second-order methods they propose provide better performance.
We note that the method in~\citet{googlefinetune} only works for vision tasks, whereas our approach works for both vision and language tasks.

\paragraph{The marginal cost of linear scaling is low.}
~\cref{tab:marginalutility} shows that the marginal cost of our HPO method is low. In the case of CIFAR10, this is because the oracle at \(\eps=0.1\) achieves \(>98\%\) accuracy.  

 \begin{table}
    \centering
    \begin{tabular}{ccccl}
    \toprule
         \(\eps_1\)& \(\eps_2\) & \(\eps_f\) & Acc  &Std\\
         \midrule
 -& -& 1.0& 99.00&0.01\\
         0.01&  0.05&  0.99&  98.88&0.01
\\
         0.05&  0.1&  0.96&  98.85&0.03
\\
 0.05& 0.2& 0.9& 98.81&0.01\\
         0.1&  0.2&  0.88&  98.81&0.01\\
         0.2& 0.3&   0.7&   98.79&0.03\\
         \bottomrule
    \end{tabular}
    \caption{The marginal cost of our HPO method is low. The first row represents the oracle. The dataset is CIFAR10.}
    \label{tab:marginalutility}
\end{table}

\paragraph{Linear Scaling produces robust results.}
In Fig. \ref{fig:pareto} we report that following Algorithm~\ref{alg:dp-raft} produces new state-of-the-art results for all values of $\varepsilon$, shown in Table \ref{tab:nonprivate-acc}. In \cref{appen:methods} we provide detailed computations of the linear interpolation for multiple datasets and in \cref{appen:heatmaps} we provide full results across the entire hyperparameter search space.
\begin{figure*}[htbp]
    \centering
{
   \begin{minipage}[t]{1\linewidth}
    \centering
    \includegraphics[width=5in]{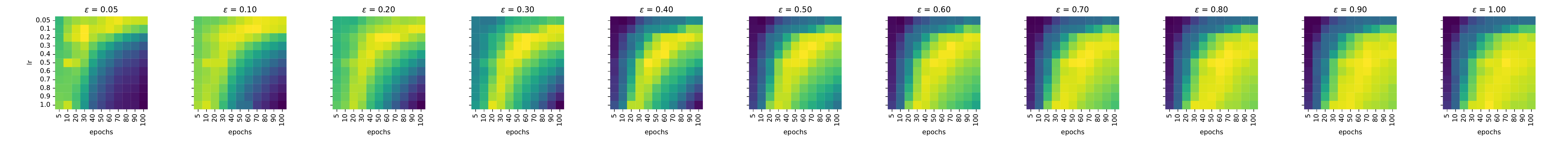}
    \end{minipage}
    }
\caption{Heatmaps for beit on CIFAR100. $\varepsilon$ increases from $0.05 \rightarrow 1.0$ left to right on the grid-axis, iterations $T$ increases from $5 \rightarrow 100$ left to right on the individual plot axis, and the learning rate $\eta$ increases from $0.05 \downarrow 1$ top to bottom on the individual plot axis.  As $\varepsilon$ increases, left to right, the optimal value of $\eta \times T$ increases in accordance with the new linear scaling rule. Prior work has generally operated in the top-left regime, that is often suboptimal.}
\label{fig:heatmaps_abbr}
\end{figure*}
\paragraph{Decomposing \(r\) into \(\eta, T\)}\label{appen:decompose}

\begin{figure}[htbp]
    \centering
    \centering
    \includegraphics[width=2.8in]{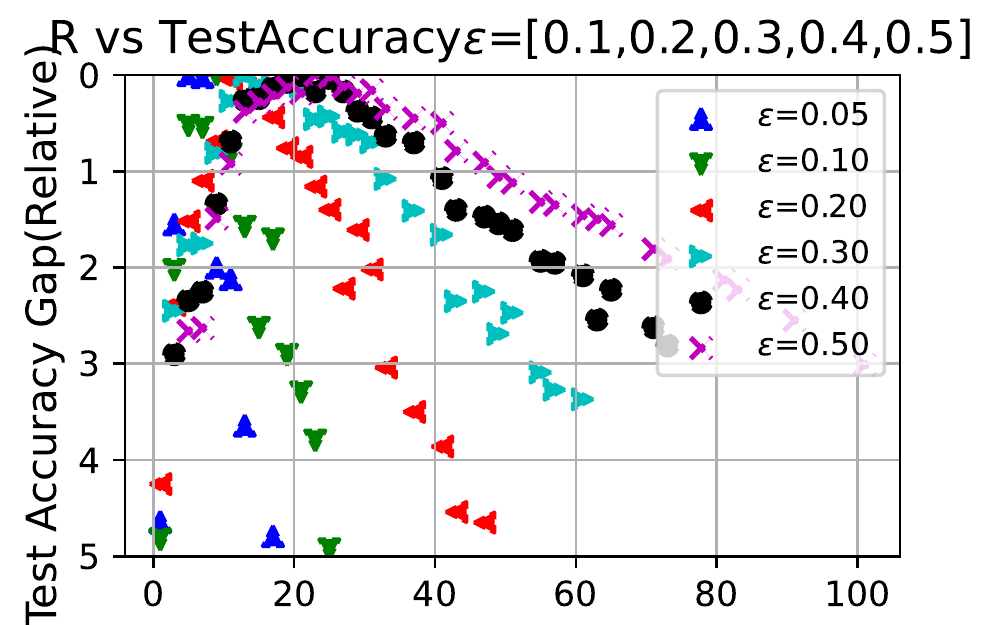}
\caption{A scatter plot of $r = \eta \times T$ the total step size vs the relative gap in test accuracy on CIFAR100 Beitv2; this gap is measured as the difference between the test accuracy at the plotted value of $r$ and the optimal value of $r$. Optimizing $r$ for any value of $\varepsilon$ and transferring this, e.g. via the linear scaling rule, will not reduce accuracy by much compared to the optimal hyperparameters.}
\label{fig:scatter}
\end{figure}

One of the advantages of our search method is that we combine the parameters that we need to search into one meta-parameter, the radius \(r\), which allows us to perform linear interpolation and also allows us to improve the runtime of the intermediate HP trials. We uniformly sample \(r\) from the search space defined by \(r_{min} = \eta_{min} \times T_{min}, (r_{max} = \eta_{max} \times T_{max}\). We evaluate 3 methods for decomposing \(r\). 1) We decompose \(r\) by sampling \(\eta, T\) from their search spaces until their product is close to the target \(r\). 2) We sample T uniformly, then get \(\eta = r / T\). 3) We sample \(\eta, T\) uniformly from their search spaces. We don't observe any significant difference between these methods. Note that the product of uniform distributions is not uniform. The robustness of the rule that ``combinations of \(\eta, T\) that evaluate to the same product perform similarly'' is crucial to the success of our method, because it enables us to fit a line rather than a more complex function that might require more evaluations.
In~\cref{fig:scatter} and ~\cref{fig:heatmaps_abbr} our results validate that this rule is robust: we can move from one set of hyperparameters to another similarly performing set of hyperparameters by increasing the number of iterations $T$ by a constant factor and decreasing the learning rate $\eta$ by the same factor (or vice versa).
We find that any inaccuracy incurred by estimating the best value of $r$ with the linear scaling rule will not reduce accuracy by much compared to doing grid search for the optimal value of $r$, but does reduce the privacy cost of hyperparameter tuning immensely.

\begin{table}[h]
    \centering
    \begin{tabular}{lcc}
        \toprule
        Method to Reduce \(\eps\) & \(\eps = 0.7\) & Degradation from \(\eps = 1.0\) \\
        \midrule
        Subsampling & 41.34\% & 45.66\% \\
        Increasing Noise & 84.07\% & 2.93\% \\
        \bottomrule
    \end{tabular}
    \caption{Comparing methods to reduce privacy cost (\(\eps\)) on ImageNet. Increasing noise is more effective than subsampling for reducing \(\eps\) with minimal performance degradation.}
    \label{tab:imagenet-loweps-comparison}
\end{table}

\subsection{Linear Scaling enables empirical analysis}\label{appen:analysis}
Many interesting questions in DP fine-tuning remain unanswered because of the immense computational overhead of evaluating hundreds of hyperparameter trials for each privacy budget, model architecture and dataset~\citep{googlefinetune}.
We now employ the linear scaling rule to efficiently answer key questions in DP fine-tuning for vision tasks.

\paragraph{Impact of model architectures on differential privacy}
\begin{figure}
    \centering
    \caption{We compare the best private and best non-private performances of all models on all datasets. We use the linear scaling rule to scale hyperparameters from $\varepsilon=0.1$ to $\varepsilon=1$, so our privacy analysis includes the cost of hyperparameter tuning.}
\centering
\small
\begin{tabular}{@{}lccccc@{}}
\toprule
Model & Dataset & $\varepsilon=1$ & $\varepsilon=\infty$ & Gap \\
\midrule
beitv2 & CIFAR10 & $98.90$ & $99.00$ & $ 0.10$ \\
 & CIFAR100 & $89.10$ & $91.57$ & $2.47$ \\
 & FMNIST & $91.02$ & $91.53$ & $0.51$ \\
 & STL10 & $99.69$ & $99.81$ & $0.12$ \\
  & EMNIST & $81.77$ & $82.00$ & $0.23$ \\
\hline 
convnext & CIFAR10 & $96.75$ & $97.22$ & $0.47$ \\
 & CIFAR100 & $83.47$ & $86.59$ & $3.12$ \\
 & FMNIST & $90.23$ & $91.13$ & $0.9$ \\
 & STL10 & $99.61$ & $99.71$ & $0.10$ \\
   & EMNIST & $78.38$ & $79.05$ & $0.67$ \\
\hline
beit & CIFAR10 & $98.19$ & $98.51$ & $0.32$ \\
 & CIFAR100 & $87.1$ & $90.08$ & $2.98$ \\
 & FMNIST & $90.55$ & $91.6$ & $1.05$ \\
 & STL10 & $99.62$ & $99.78$ & $0.16$ \\
 & EMNIST & $81.48$ & $83.25$ & $1.77$ \\ 
\hline 
vit-L & CIFAR10 & $98.29$ & $98.44$ & $0.40$ \\
 & CIFAR100 & $86.18$ & $89.72$ & $3.54$ \\
 & FMNIST & $90.58$ & $91.37$ & $0.79$ \\
 & STL10 & $99.62$ & $99.76$ & $0.14$ \\
\bottomrule
\end{tabular}

\label{tab:nonprivate-acc}
        \label{tab:cv-results}
  \end{figure}
Many pretrained model architectures are available~\citep{huggingface} but prior work has generally engaged with a single architecture, e.g. beit~\citep{bu2022scalable} or ViT~\citep{mehtadptransfer}.
We leverage our method to answer three questions:
\begin{itemize}[leftmargin=15pt]
    \item What model architectures can provide good DP classifiers?
    \item Is the best model task-specific, e.g., is an architecture search required?
    \item Does the private-non private utility gap depend on the model architecture?
\end{itemize}
We report our findings in Tab.~\ref{tab:cv-results}.
We evaluate multiple transformer architectures in ViT~\citep{vit}, beitv1~\citep{beit} and beitv2~\citep{beitv2}, as well as the purely convolutional architecture Convnext~\citep{convnext}.
We find that all architectures can serve as good backbones for high-accuracy DP classification.
This is somewhat surprising because the different inductive biases of transformers and purely convolutional architectures tend to produce differently structured features, but we reason that the noise added by DP will `smooth out' these decision boundaries regardless of architecture.
We note that one architecture, beitv2, performs the best on all benchmarks and also has the highest non-private ImageNet accuracy~\citep{rw2019timm}.
We therefore recommend that practitioners do not worry about architecture search when fine-tuning as this can incur further privacy costs, and instead pick the best model available.
We are encouraged to report that the private-non private utility gap diminishes with model accuracy, enabling us to report for the first time \emph{lossless privacy} of $99.0 \%$ on CIFAR10 at $\eps=1$ (without considering the cost of HPO) and the gap is only \(<0.10\%\) if we consider the cost of HPO. 
We expect that as pretrained models become even better, future works may even be able to attain lossless privacy on CIFAR100, that we note remains somewhat challenging for private fine-tuning.
We harness these insights for our next analyses.

\paragraph{DP models are robust to distribution shifts.}\label{appen:ood}
If we assume the existence of some publicly available data for pretraining and then do DP fine-tuning on the private data, it's crucial that there is no privacy leakage between the public data and private data. There is only 0 distribution shift when public = private, and this violates the key assumption (no privacy leakage because public and private data are sufficiently different) in DP fine-tuning. If the public data is so different from the private data that it can be used for pretraining without privacy leakage, there must be some distribution shift.
Benchmarking performance on datasets with distribution shifts is increasingly important because real-world problems almost always contain distribution shift between model training and inference~\citep{Rahimian_2022}.

\begin{figure}
\centering
\caption{In-distribution (ID) and out-of-distribution (OOD) performance on benchmark distribution shift datasets. 
We use the linear scaling rule to scale hyperparameters from $\varepsilon=0.1$ to $\varepsilon=1$, so our privacy analysis includes the cost of hyperparameter tuning.}    
\begin{tabular}{c|cccc|cc}
\toprule 
Dataset & $\varepsilon=1.0$ ID(OOD) & Prior ($\varepsilon=\infty$) \\
\midrule
    Waterbirds & $92.31$ ($91.59$) & $98.3$($80.4$) \\
    fMoW & $45.44$ ($35.31$)  & $49.1$ ($36.6$) \\
    Camelyon & $93.91$ ($93.55$) & $99.5$ ($96.5$) \\
    C10 $\rightarrow$ STL & $99.0$ ($98.82$) & $97.5$ ($90.7$) \\
    C10 $\rightarrow$ C10p1 & $99.0$ ($97.85$) & $97.5$ ($93.5$) \\
    C10 $\rightarrow$ C10C & $99.0$ ($89.98$) & $96.56$ ($92.78$) \\
    C100 $\rightarrow$ C100C & $89.65$ ($68.69$) & $81.16$ ($72.06$) \\
\bottomrule
\end{tabular}
\label{tab:ood}
\end{figure}

We show that DP-SGD provides robustness to covariate, subpopulation and label distribution shifts for synthetic and natural datasets. We compare to other methods that consider this question.

\paragraph{Details on OOD Experiments}
We specify exact details for all OOD experiments.
Our training details are drawn from prior work~\citep{kumarfreezefinetune,kumarfinetunedistort,diffenderfercifar10cbenchmark}.
\textbf{Waterbirds}: the ID$\rightarrow$OOD contains a well-studied spurious correlation in the binary classification problem.
~\citep{raghavwaterbirds} evaluate vision transformers without using group knowledge and obtain $\approx$ 80 $\%$ ID accuracy, but much worse ($\approx 60 \%$) OOD accuracy, and~\citep{kumarfreezefinetune} tailor their method to this task and get the reported results.
Surprisingly, just fine-tuning a linear model on the extracted features outperforms both works for OOD accuracy for $\varepsilon=0.1$.
This trend (sacrificing ID accuracy for increased OOD robustness) is seen in other OOD results, and we hypothesize that this is due to the inherent regularization present in DP-SGD.

\textbf{Fmow:} we train on region 3 (ID) and evaluate on regions 1,2 (OOD), following~\citep{kumarfinetunedistort}.

\textbf{Camelyon17:} we again follow~\citep{kumarfinetunedistort}.

\textbf{CIFAR10 $\rightarrow$ STL10, CIFAR10p1:} We train privately on CIFAR10 using our best hyperparameters returned from the linear scaling rule and then transfer this to STL10/CIFAR10p1, with the label reassignment following~\citep{kumarfreezefinetune}.

\textbf{Common Corruptions:} We evaluate on the average severity of the 'gaussian blur' corruption.

We leverage our method to answer three questions:
\begin{itemize}[leftmargin=15pt]
    \item Can DP help when there is a domain shift from private fine-tuning to test?
    \item Can DP help when there is a domain shift from public data to private fine-tuning?
    \item Can DP fine-tuned models perform well in the zero-shot setting?
\end{itemize}
In Table \ref{tab:ood} we compare the performance of our method across 8 benchmarks and find that the answer to all three of these questions is \emph{yes}.

The Waterbirds dataset is a well-known benchmark for evaluating the robustness of models to spurious correlations.
There is a domain shift between the private training data and the private test data created by class imbalance.
We are surprised to find that in the absence of any other regularization methods, DP fine-tuning actually \emph{improves} performance on the OOD split.
We hypothesize that the lackluster OOD non-private performance is caused by the model overfitting to the spurious correlation in the training data, and that the inherent regularization of DP prevents the model from memorizing this spurious correlation.
By comparing our results to~\citet{raghavwaterbirds} we determine that this robustness is unique to DP rather than an artifact of the pretrained model.
Although DP does significantly degrade the ID performance, in situations where minimizing OOD error is more important, we believe that DP by itself can mitigate the domain shift from private fine-tuning to test.

Because our central assumption in DP fine-tuning is that there is no privacy leakage from the pretraining data to the private training data, it is important to understand how DP fine-tuning performs when there is a distribution shift between public data and private data.
fMoW~\citep{fmow} and Camelyon17~\citep{camelyon} are two datasets that represent a signficant distribution from the pretraining data (ImageNet).
We observe a similar relationship between ID and OOD degradation as above, where the OOD degradation is somewhat mitigated by DP.
If we compare our results on Camelyon to the best results in~\citet{ghalebikesabi2023differentially} we find that we can improve their best performance from $91.1\%$ at $\varepsilon=10$ to $93.91\%$ at $\eps=1$.
Although performance on fMoW remains quite poor, we note that it is not significantly worse than in the non-private setting.
We believe that DP fine-tuning from pretrained models remains a viable strategy even when the publicly available pretraining data has a very large distribution shift from the private target data.

We finally consider the zero-shot setting, where we fine-tune a model on CIFAR and then transfer it without updating any parameters to private test datasets that once again represent a distribution shift from CIFAR. 
We report the performance in the OOD column. 
For the more minute distribution shifts of STL and CIFAR10p1 we find that the fine-tuned classifier can achieve remarkable performance without ever updating parameters on these datasets; that is, we just remap the labels as per~\citep{kumarfinetunedistort}.
CIFAR10C and CIFAR100C represent larger distribution shifts and are used to benchmark the robustness of models to commonly reported image corruptions~\citep{hendrycks}.
Our OOD performance on these larger distribution shifts is much worse, particularly for CIFAR100 where there is a $>20\%$ degradation.
Although this is lower than the top result on the RobustBench leaderboard~\citep{croce2021robustbench} obtains $85\%$ accuracy, we note that once again \emph{we used no additional methods beyond DP to ensure robustness but managed to achieve reasonable performance to distribution shifts in zero-shot classification}.

\paragraph{Comparison to other works on distribution shift under DP.}
Prior work in distributionally robust optimization (DRO) has addressed this problem by using knowledge of the relative imbalances between groups, but recent work with vision transformers has shown that linear probing can perform well on datasets with distribution shifts~\citep{raghavwaterbirds, ananyafinetuning, kumarfreezefinetune}.
~\citet{kulynych2022get} proposes DP-IS-SGD that improves the robustness of DP-SGD by removing per-sample gradient clipping (therefore removing the introduced bias but also losing the privacy guarantee; see 4.2) and uses knowledge of the groups to sample subpopulations at different rates to improve robustness. Because our method uses DP-GD to maximize the signal-to-noise ratio of updates and requires clipping (because our primary goal is the privacy guarantee, unlike~\citet{kulynych2022get} which focuses on DRO) and we do not assume knowledge of groups, we cannot make use of DP-IS-SGD.~\citet{hulkund2023limits} concludes that "[DP-SGD] is not a good candidate for improving robustness under covariate or subpopulation shift, as it comes at a major cost to accuracy." This conclusion runs counter to our findings, and we believe the reason is because their numerical findings are not conclusive. Our interpretation of their results is that because their DP-SGD degrades accuracy, it should also increase robustness; however we find that even when DP-SGD does not degrade accuracy it still improves robustness.

\label{tab:intro_appen}

\subsection{Detailed Ablations}\label{appen:ablations}
\begin{table*}[htbp]
    \centering
    \caption{Our method fixes six design choices: the architecture and initialization (for CV tasks only), the batch size (full batch), the optimizer (SGD with momentum=0.9), the accounting method (PLV where all prior HPO methods use RDP), and the clipping norm (unit clipping). We report the improvement derived from following each of these techniques with respect to a competitive baseline from prior work on CIFAR100 at $\eps=0.1$.}
    \begin{tabular}{cccc}
    \toprule
    Method & Baseline & Baseline Accuracy & Improvement\\
    \midrule
    Classifier (no bias) & ~\citep{mehtadptransfer} & $71.3$ & $0.36$ \\
    Zero Initialization & Random Initialization~\citep{deepmind} & $64.85$ & $6.81$ \\
    Gradient Descent & SGD(Batch=4096)~\citep{deepmind} & $70.2$ & $1.46$ \\
    Momentum ($\rho=0.9$) & $\rho=0$~\citep{bu2022scalable} & $69.02$ & $2.09$ \\
    PLV Accounting & RDP~\citep{deepmind} & $68.43$ & $3.23$ \\
    Unit Clipping ($C=1$) & $C \ll 1$~\citep{googlefinetune} & $71.2$ & $0.46$ \\
    \bottomrule
    \end{tabular}
    \label{tab:recipe-appen}
\end{table*}
In this subsection we deal with detailed ablations of each step in the method that we use.
We ablate each step and show their individual benefits in Table~\ref{tab:recipe-appen}.
At a high level, we want to maximize the signal-to-noise ratio of updates, accelerate training to minimize the impact of noise on the optimization trajectory, and apply the linear scaling rule to select the best hyperparameters while maintaining a given overall privacy budget.

\textbf{1) Extract features from a private dataset using an open source feature extractor pretrained on a public dataset.}
A valid criticism of this approach in private fine-tuning is that the fine-tuning dataset can be in-distribution with the training dataset, and this may violate privacy.
To address this we evaluate our method on eight datasets that have been used as distribution shift benchmarks in Sec.~\ref{sec:evaluation}.

\textbf{2) Zero-initialize a linear classifier that maps features to classes.}
Prior work has studied full network fine-tuning~\citep{papernotfinetuning,bu2022scalable,deepmind} but we find that by doing logistic regression on a linear classifier we minimize the number of parameters, and mitigate the curse of dimensionality. 
We further simplify the choice of initialization by initializing all parameters to zero.

\textbf{3) Apply linear scaling to privately select the step size and number of steps.} 
We propose a new linear scaling rule: increase either the step size $\eta$ or number of steps $T$ so that the total step size $r = \eta \times T$ is linear in $\varepsilon$. 
This reduces the hyperparameter search to a binary search in $r$. 
Furthermore we can do a hyperparameter search for $r$ using a small privacy budget, and then linearly scale up this value to minimize the cost of hyperparameter search(Alg.~\ref{alg:dp-raft}).
Using privacy loss accounting enables us to get competitive accuracy for privacy budgets as small as $\varepsilon=0.01$, so these low-cost trials can inform better hyperparameters.
our method already minimizes the private-nonprivate performance gap at $\varepsilon=1.0$ as we show in Table~\ref{tab:cv-results}, so spending $\varepsilon=0.1$ for hyperparameter tuning does not significantly degrade accuracy.
Unless stated explicitly otherwise, all privacy-utility tradeoffs reported for our method in the main body include the privacy cost of hyperparameter tuning via the linear scaling rule.

\textbf{4) Compute the full batch gradient.} 
This optimizes the signal-to-noise ratio of the update and enables use of large step sizes~\citep{goyallinearscalingrule}. We achieve 91.52$\%$ accuracy on CIFAR10 ($|D|=5e4$) for $\varepsilon=0.01$ when training for 100 epochs with noise multiplier $\sigma=2561$.
When the noise is divided by the batch size, the effective noise multiplier is $\frac{\sigma}{|B|=5e4} \approx 0.05$ and the SNR is $\frac{1}{0.05}=20$.
When we use subsampling with sampling probability $p=0.2$ and train for the same number of epochs under the same privacy budget, our effective noise multiplier is $\frac{\sigma}{|B|} = \frac{1145}{1e4} = 0.114$, and the corresponding SNR of ${\frac{1}{0.114}=8.7}$ is much worse than in the full batch setting.

\textbf{5) Clip per-sample gradients to unit norm. }
As per Eq.~\ref{eq:dpsgd} reducing the per-sample gradient below 1 is equivalent to reducing $\eta$ (and thus reducing the step size) while simultaneously biasing optimization. 
By setting $c=1$ we can simplify $r = \eta \times T \times c$ to $r = \eta \times T$.

\textbf{6) Use privacy loss variable accounting.}
~\citet{glwmsaccountant} provides a tool to calibrate Gaussian noise for the given privacy budget and add noise to the gradient: this enables budgeting for small values of $\varepsilon$ without underestimating privacy expenditure.

\textbf{7) Use momentum.}
Acceleration has a host of well-known benefits for optimization and is ubiquitous in non-private optimization~\citep{qian1999momentum,kingmaadam}, but prior work has not always used momentum because it can lead DP-SGD astray when the SNR of updates is low~\citep{deepmind}. 
Because we optimize the SNR of individual updates in (4), we can make use of momentum.

\begin{figure*}[htbp]
    \centering
    \subfigure[Momentum Acceleration]{
    \begin{minipage}[t]{0.48\linewidth}
    \centering
    \includegraphics[width=2.8in]{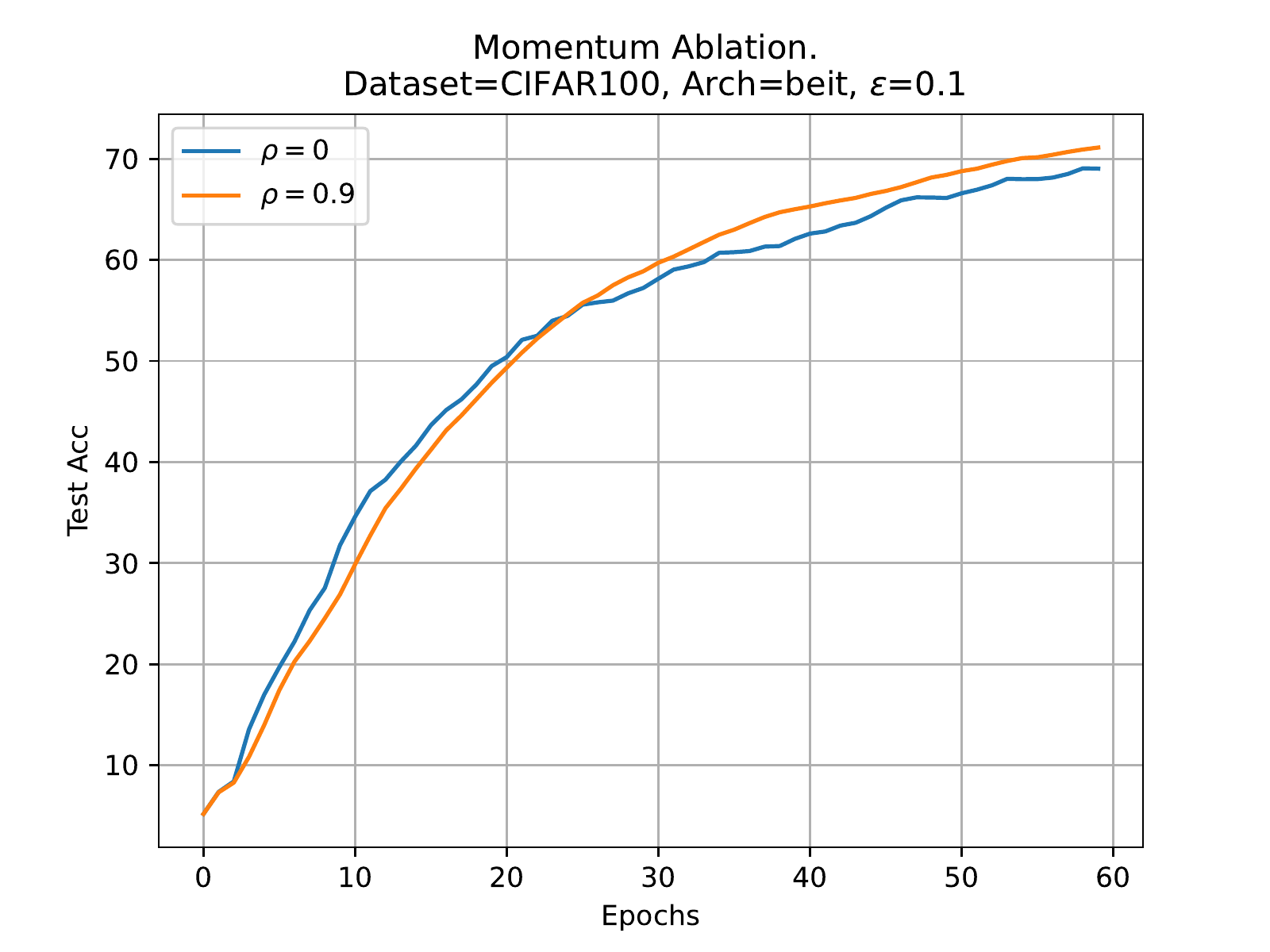}
    \end{minipage}
    }
    \subfigure[Momentum Post-Processing]{
    \begin{minipage}[t]{0.48\linewidth}
    \centering
    \includegraphics[width=2.8in]{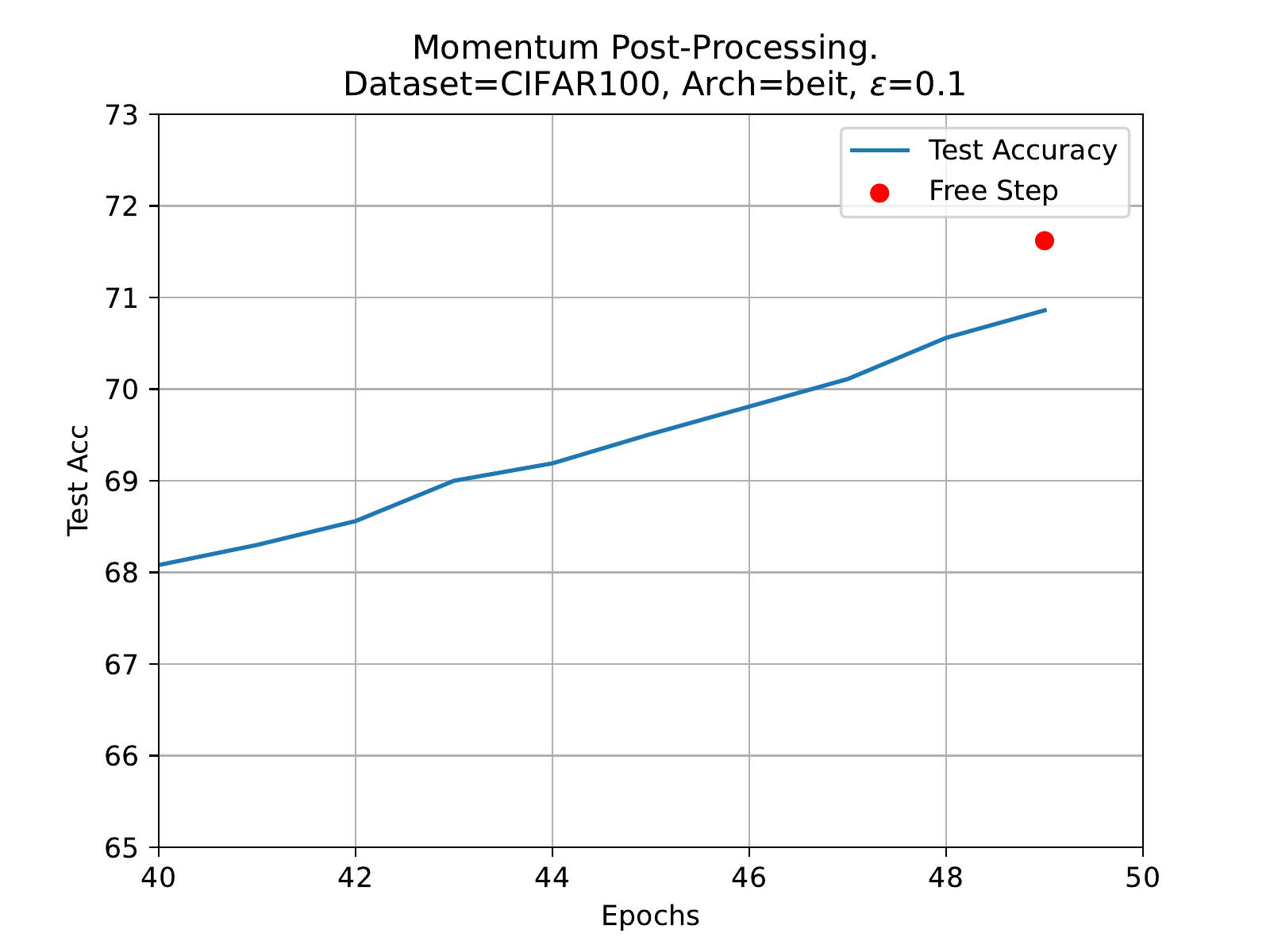}
    \end{minipage}
    }
    \centering
\caption{Ablation of momentum parameter during training (left) and post processing of the parameter exponential moving average stored in the momentum buffer to take an extra step 'for free' (right). Use of both methods increases performance slightly.}
\label{fig:ablations:momentum}
\end{figure*}

\paragraph{Momentum Accelerates Convergence.}
Despite the exhaustive study of the acceleration of gradient descent with momentum done by prior work~\citep{sutskevermomentum, qian1999momentum} work on DP-SGD generally eschews the use of a momentum term.
A notable exception~\citep{googlefinetune} use AdamW rather than SGD with momentum; in a later section we discuss the reason to prefer SGD with momentum.
The reason to use momentum to accelerated the convergence of DP-SGD is straightforward: the exponentially moving average of noisy gradients will have higher SNR than individual gradients. 
Furthermore, momentum is shown to provably benefit normalized SGD~\citep{pmlr-v119-cutkosky20b}.
In Fig. \ref{fig:ablations:momentum} we observe that momentum complements our new linear scaling rule and accelerates convergence.
Separately, we report the improvement of taking a step 'for free' in the direction of the exponential moving average stored during training in the momentum buffer.
Note that this exponential moving average is in no way tied to momentum, and it is equivalent to perform DP-SGD without acceleration, store an exponential moving average of gradients with decay parameter $\gamma=0.9$, and then take an additional step in the direction of the stored gradient average after training has finished; we only use the momentum buffer for ease of implementation.
As we discuss above when introducing the new linear scaling rule, we maximize performance by maximizing SNR and terminating training while the model is still improving.
Intuitively we therefore expect that the momentum buffer will contain a good estimate of the direction of the next step that we would have taken had we continued training, and taking a step in this direction with our usual learning rate should only improve performance without any privacy loss.
We use momentum with $\rho=0.9$ in all other experiments and also take a 'free step' at the end of private training. 

\begin{figure*}[htbp]
    \centering
    \subfigure[Batch Size]{
    \begin{minipage}[t]{0.48\linewidth}
    \centering
    \includegraphics[width=2.8in]{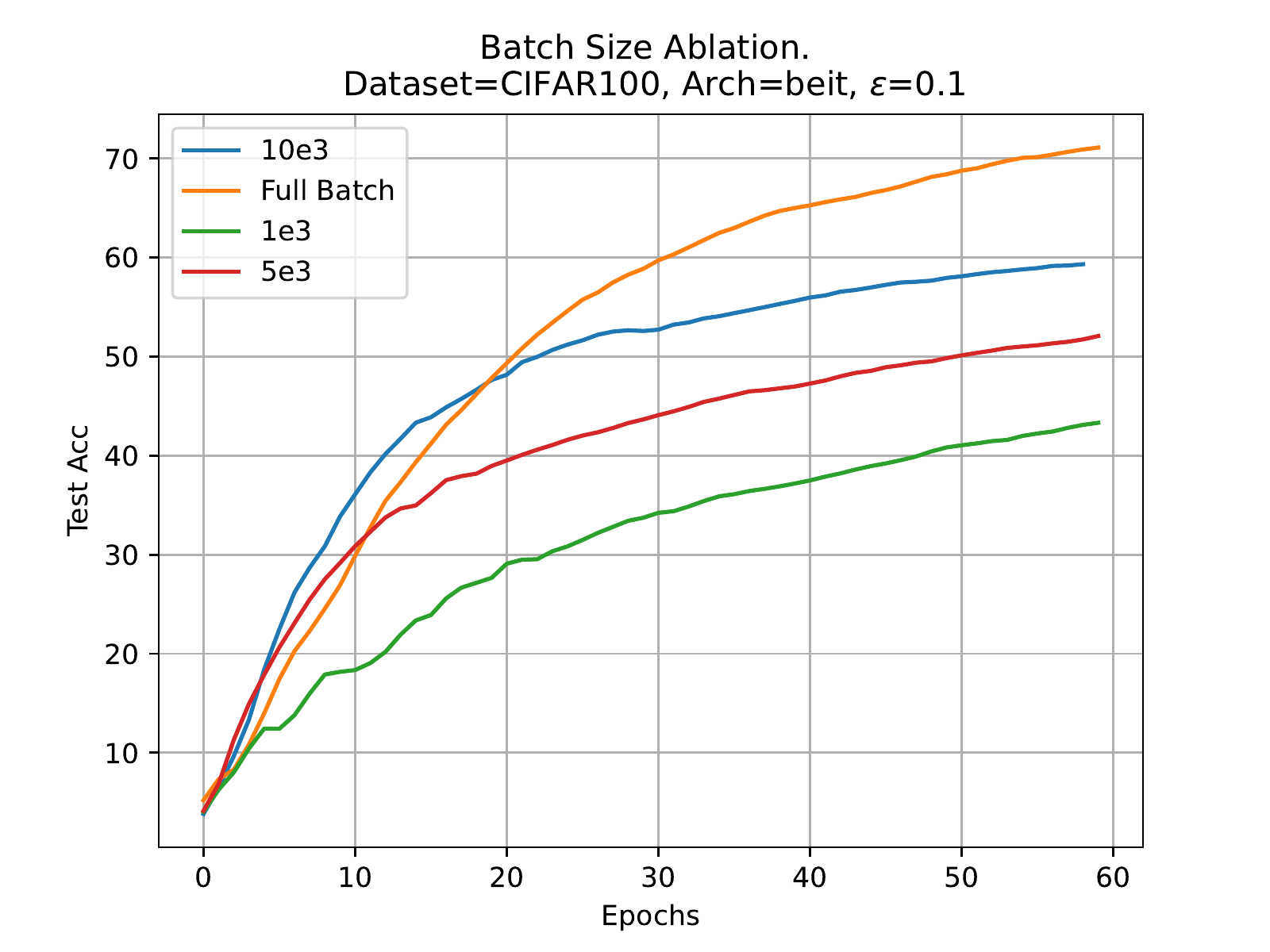}
    \end{minipage}
    }
    \subfigure[Gradient Descent vs SGD]{
    \begin{minipage}[t]{0.48\linewidth}
    \centering
    \includegraphics[width=2.8in]{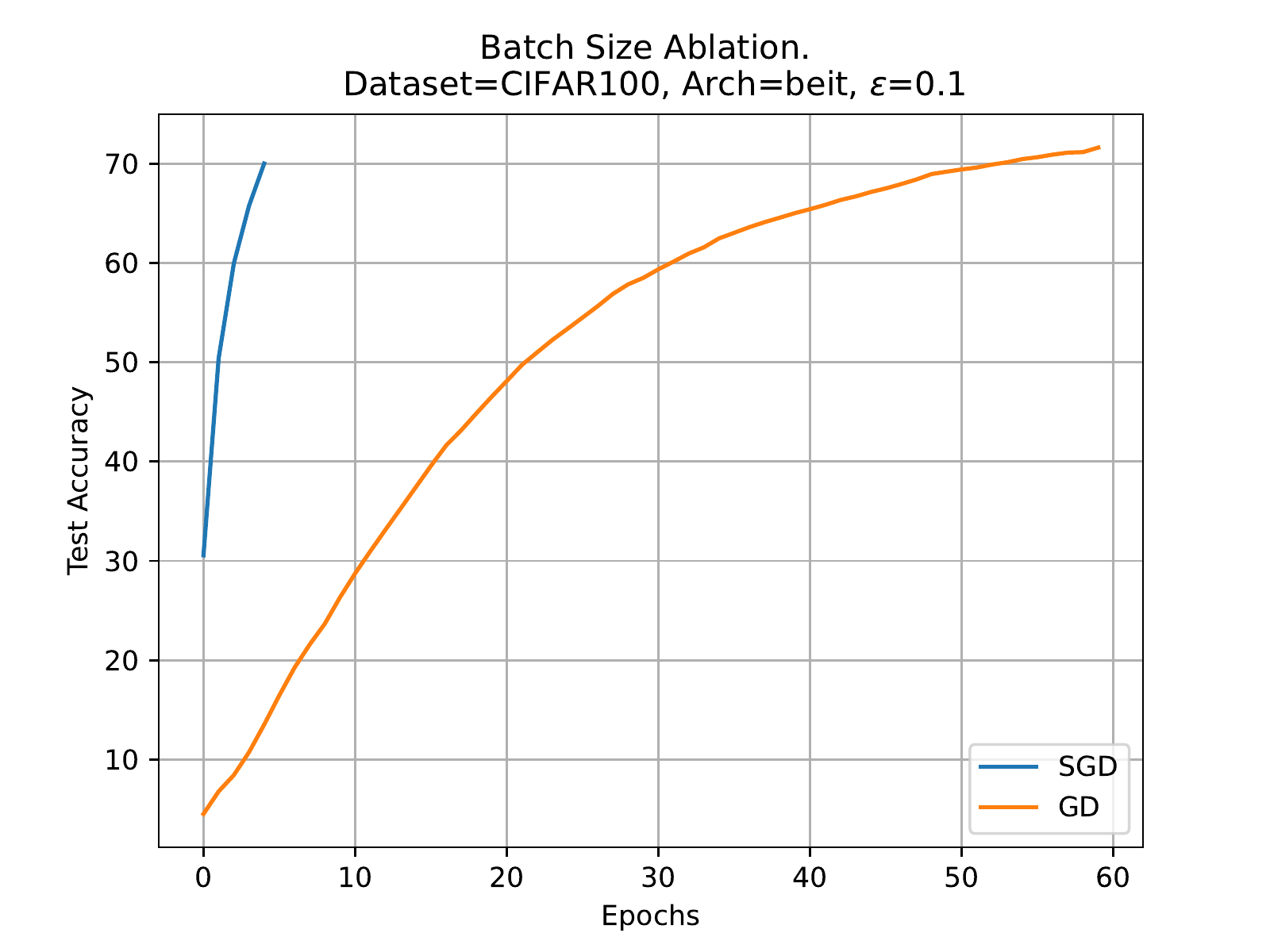}
    \end{minipage}
    }
    \centering
\caption{Ablation of batch size. Left: We vary the batch size using the learning rate and number of iterations tuned for full batch; all other batch sizes perform much worse. Right: We compare SGD and GD. For SGD we tune the batch size jointly with learning rate and number of iterations, arriving at a batch size of 4096 and plot the best performing run against full batch. 
}
\label{fig:ablations:bsz}
\end{figure*}

\paragraph{Full Batches Optimize Signal-to-Noise Ratio.}
Since its inception, the use of privacy amplification via Poisson subsampling and RDP has been a mainstay in the DP community~\citep{zhupoissonrdp, wangsubsampledrdp, erlingssondpshuffling}.
Prior work almost universally uses privacy amplification via subsampling, but as early as~\citep{mcmahandp}, and more recently in~\citep{deepmind} it has become apparent that DP-SGD can actually benefit from large batch sizes because the signal-to-noise ratio (SNR) improves.
Note that the noise term in \ref{eq:dpsgd} is divided by the batch size, so if we are willing to give up amplification via subsampling entirely, we can reduce the noise by a factor of $5e4$ for the benchmark computer vision tasks.
In Fig. \ref{fig:ablations:bsz} we report the improvement of full-batch DP-GD over Poisson subsampled DP-SGD. 
We attribute the success of DP-GD to the improvement in SNR.
For example, we achieve 91.52$\%$ accuracy on CIFAR10 for $\varepsilon=0.01$ when training for 100 epochs with learning rate $\eta=0.01$ and noise multiplier $\sigma=2561$.
When the noise is divided by the batch size, the effective noise multiplier is $\frac{\sigma}{|B|=5e4}=0.05$ and the SNR is $\frac{1}{0.051}=20$.
When we use subsampling with sampling probability $p=0.2$ and train for the same number of epochs under the same privacy budget, our effective noise multiplier is $\frac{\sigma}{|B|} = \frac{1145}{1e4} = 0.114$, and the corresponding SNR of ${\frac{1}{0.114}=8.7}$ is much worse than in the full batch setting.
Although at first glance our analysis merely supports the typical conclusion that large batches are better in DP-SGD,~\citep{deepmind} observe that DP-SGD is still preferrable to DP-GD because minibatching produces the optimal choice of noise multiplier. 
Our findings run counter to this: as discussed above, we contend that performance depends not only on the optimal noise multiplier but on our new linear scaling rule, and DP-GD unlocks the use of larger step sizes~\citep{goyallinearscalingrule}.
We use DP-GD instead of DP-SGD in all other experiments, removing the batch size from the hyperparameter tuning process and improving the overall privacy cost of deploying our baselines~\citep{papernothparamtuning}. 

\subsection{A Critical Evaluation of Proposed Techniques for Fine-Tuning}
Prior work has proposed a number of ad-hoc techniques that improve performance in DP fine-tuning.
Here we critically evaluate these techniques in the our method regime, and analyze why they reduce performance in our setting.

\paragraph{Small Clipping Norms Bias Optimization.}
The standard deviation of the noise added in DP-SGD scales with the sensitivity of the update, defined by the clipping norm parameter.
To decrease the amount of noise added, prior work has used very strict clipping~\citep{googlefinetune,bu2022scalable}.
Intuitively, if the clipping norm parameter is already chosen to be some value smaller than the norm of the unclipped gradient, the gradient estimator is no longer unbiased and this may have a negative impact on optimization.
In Fig. \ref{fig:ablations:clipnorm} we observe that decreasing the clipping norm below 1 only degrades performance.
As we can see in equation \ref{eq:dpsgd}, further decreasing the clipping norm is equivalent to training with a smaller learning rate, and this is suboptimal because Fig. \ref{fig:heatmaps-full-a} indicates that we can prefer to use larger learning rates.
We use a clipping norm of 1 in all other experiments. 

\begin{figure*}[htbp]
    \centering
    \subfigure[Weight Initialization]{
    \begin{minipage}[t]{0.48\linewidth}
    \centering
    \includegraphics[width=2.8in]{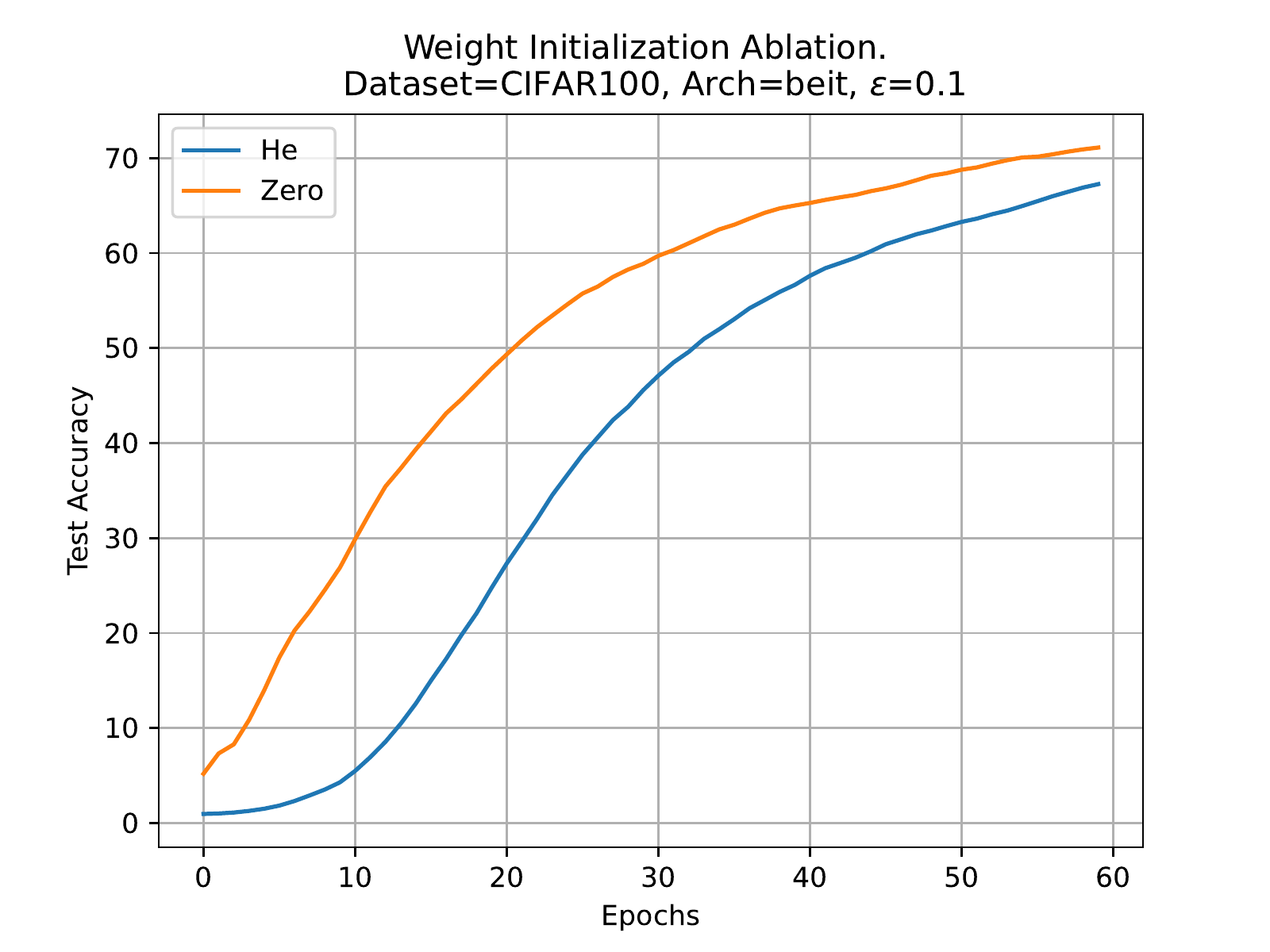}
    \end{minipage}
    }
    \subfigure[Weight Decay]{
    \begin{minipage}[t]{0.48\linewidth}
    \centering
    \includegraphics[width=2.8in]{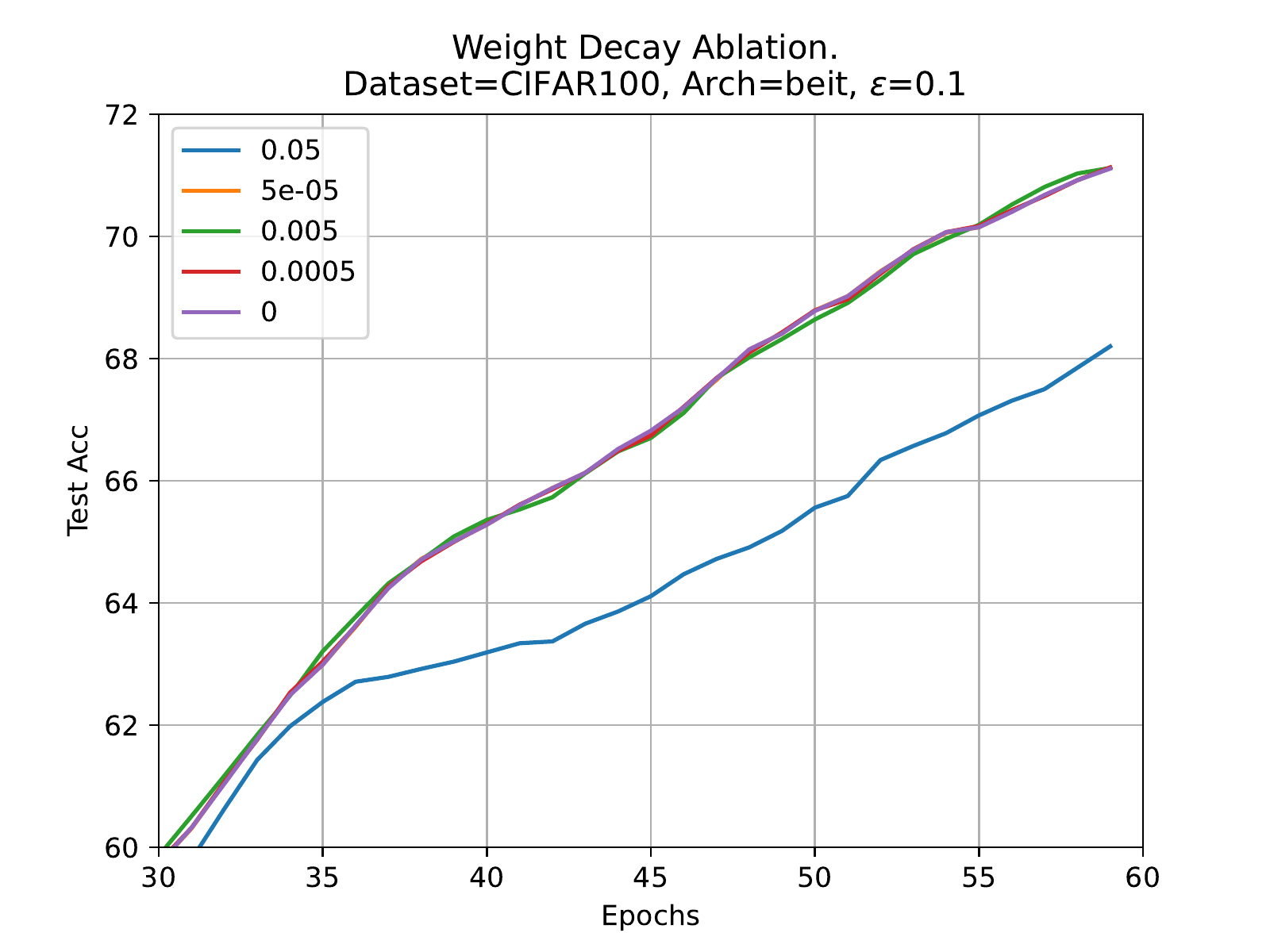}
    \end{minipage}
    }
    \centering
\caption{Ablation of two previously proposed methods: zero initialization of parameters and weight decay. Zero initialization increases accuracy in all experiments, but weight decay only degrades performance.}
\label{fig:ablations:standardization-decay}
\end{figure*}

\begin{figure*}[htbp]
    \centering
    \includegraphics[width=2.8in]{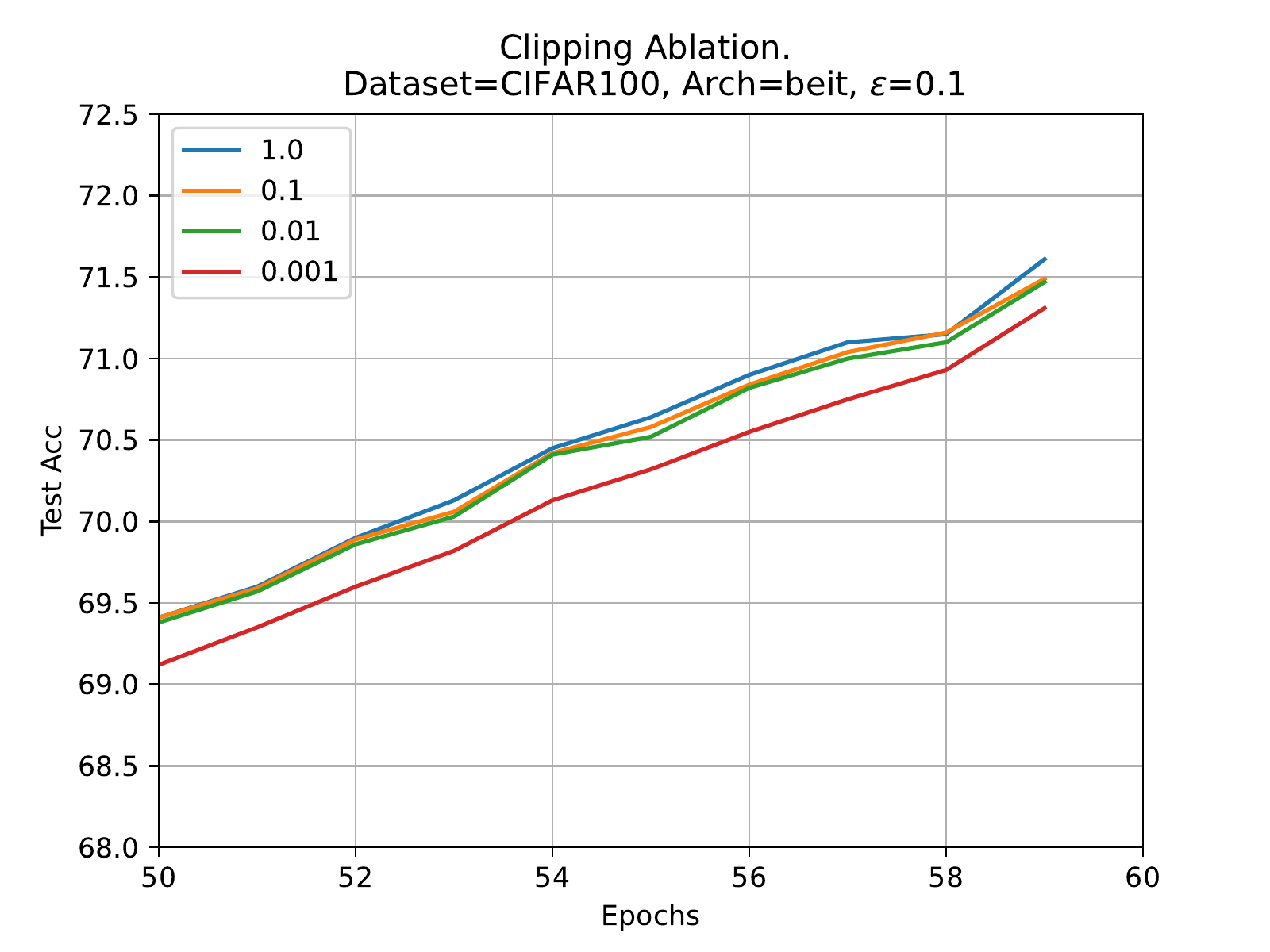}
        \centering
\caption{Because reducing the clipping norm is equivalent to reducing the learning rate, reducing the clipping norm below $1$ only degrades performance on CIFAR100 for the beit architecture at $\varepsilon=0.1$.}
\label{fig:ablations:clipnorm}
\end{figure*}

\paragraph{Initializing Weights to Zero Mitigates Variance in DP-GD.}
~\citep{qiaoweightstandardization} propose initializing the model parameters to very small values to improve the stability of micro-batch training, and~\citep{deepmind} find that applying this technique to DP-SGD improves performance.
In Fig. \ref{fig:ablations:standardization-decay} we ablate the effectiveness of \emph{zero initialization} with standard He initialization and find that the best performance comes from initializing the weights uniformly to zero.
We initialize the classifier weights to zero in all other experiments. 

\begin{figure*}[htbp]
    \centering
    \subfigure[Weight Averaging]{
    \begin{minipage}[t]{0.48\linewidth}
    \centering
    \includegraphics[width=2.8in]{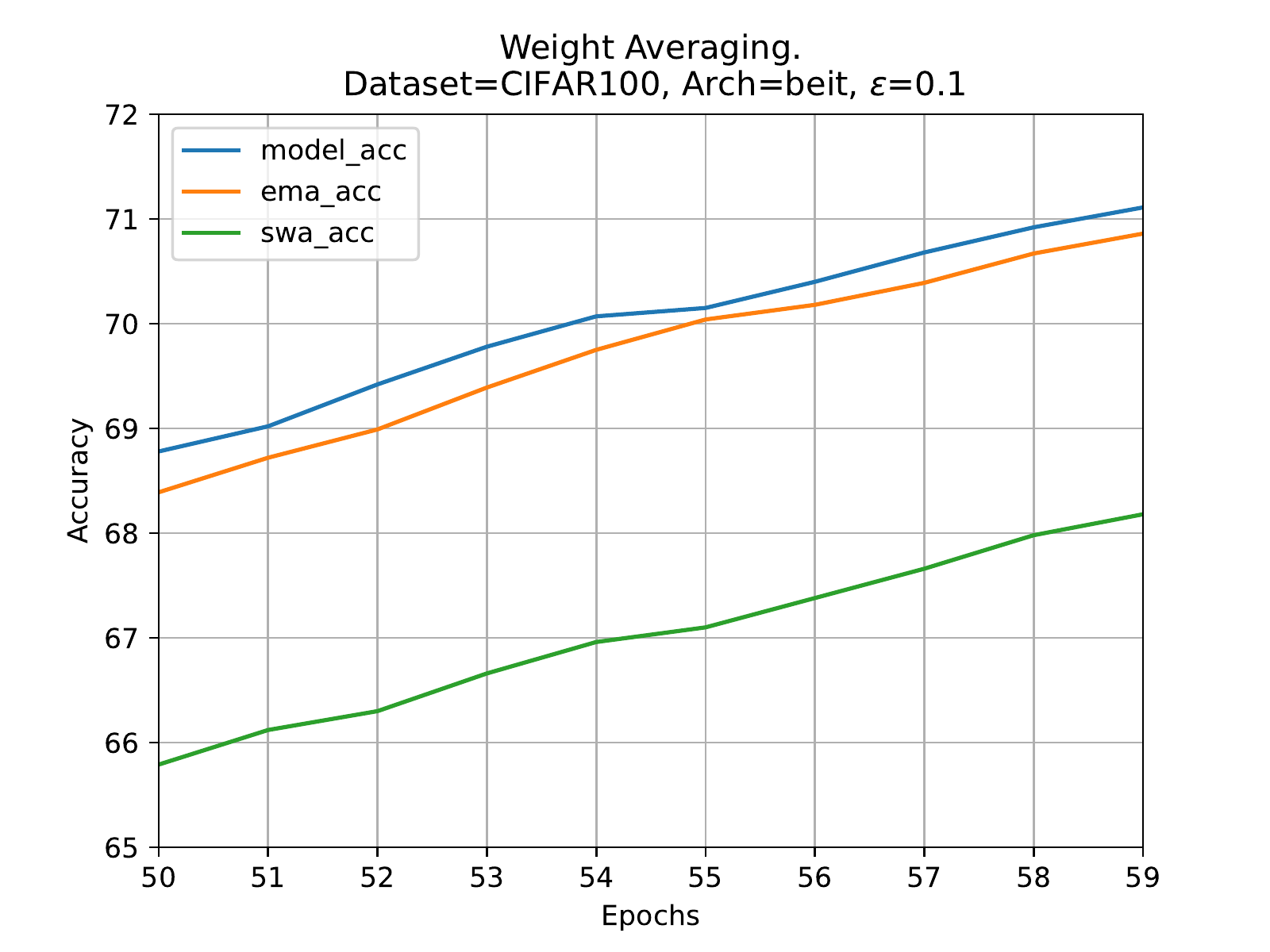}
    \end{minipage}
    }
    \subfigure[Weight Trajectory]{
    \begin{minipage}[t]{0.48\linewidth}
    \centering
    \includegraphics[width=2.8in]{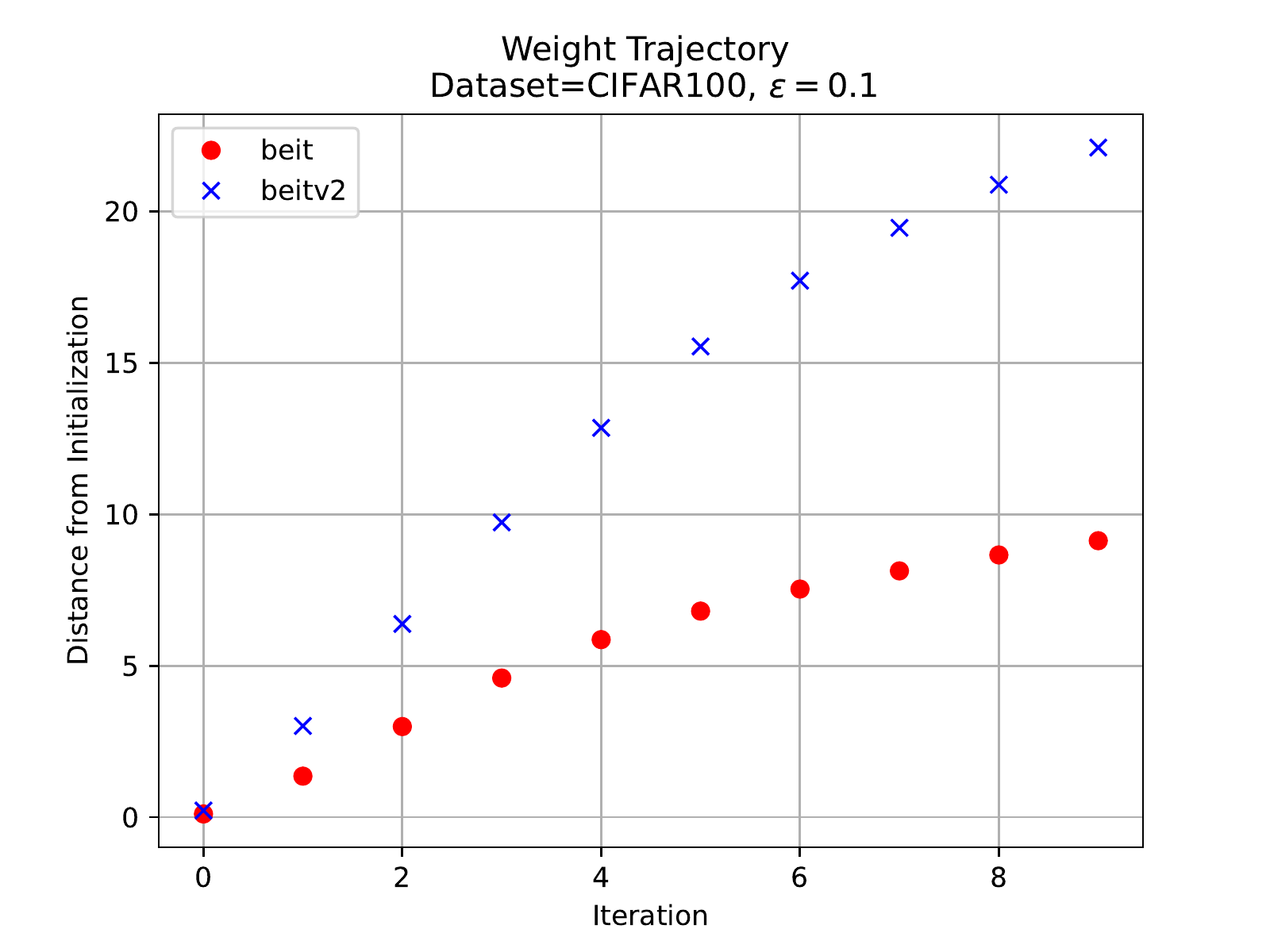}
    \end{minipage}
    }
    \centering
\caption{Left:Ablation of Weight Averaging. Right: Plot of distance from initialization. Weight Averaging does not improve performance because the model is monotonically moving away from the initialization and weight averaging cannot 'catch up'.}
\label{fig:ablations:averaging}
\end{figure*}

\paragraph{Weight Averaging Cannot Catch Up To Accelerated Fine-Tuning.}
~\citep{shejwalkardpcheckpoint} perform an in-depth empirical analysis and find that averaging the intermediate model checkpoints reduces the variance of DP-SGD and improves model performance.
~\citep{deepmind} first proposed the use of an Exponential Moving Average (EMA) to mitigate the noise introduced by DP-SGD.
Previously, methods that use stochastic weight averaging (SWA) during SGD have been proposed and are even available by default in PyTorch~\citep{izmailovswa}.
The idea of averaging weights to increase acceleration was first proposed by~\citep{Polyak1992AccelerationOS}, and is theoretically well-founded.
In Fig. \ref{fig:ablations:averaging} we compare EMA and SWA with no averaging and find that no averaging performs the best. 
This is because weight averaging methods work well when optimization has converged and the model is plotting a trajectory that orbits around a local minima in the loss landscape~\citep{izmailovswa}.
That is to say, the model's distance from the initialization does not continually increase and at some point stabilizes so that the weight averaging method can 'catch up'.
However, as discussed in Fig. \ref{fig:pareto} the optimal number of iterations for our method is to train for longer epochs without decaying the learning rate for convergence, because when the model converges the SNR decays.
This is corroborated by Fig. \ref{fig:ablations:averaging}, where we see that the distance from initialization is monotonically increasing.
Our findings run counter to those of~\citep{shejwalkardpcheckpoint} for hyperparameters in line with our proposed linear scaling rule because we find that the best optimization regime for our method is precisely one where weight averaging can never catch up to the optimization trajectory. 
Therefore, the averaging methods only serve to lag one step behind no averaging. 

\begin{figure*}[htbp]
    \centering
    \subfigure[CIFAR100 (hard)]{
    \begin{minipage}[t]{0.48\linewidth}
    \centering
    \includegraphics[width=2.8in]{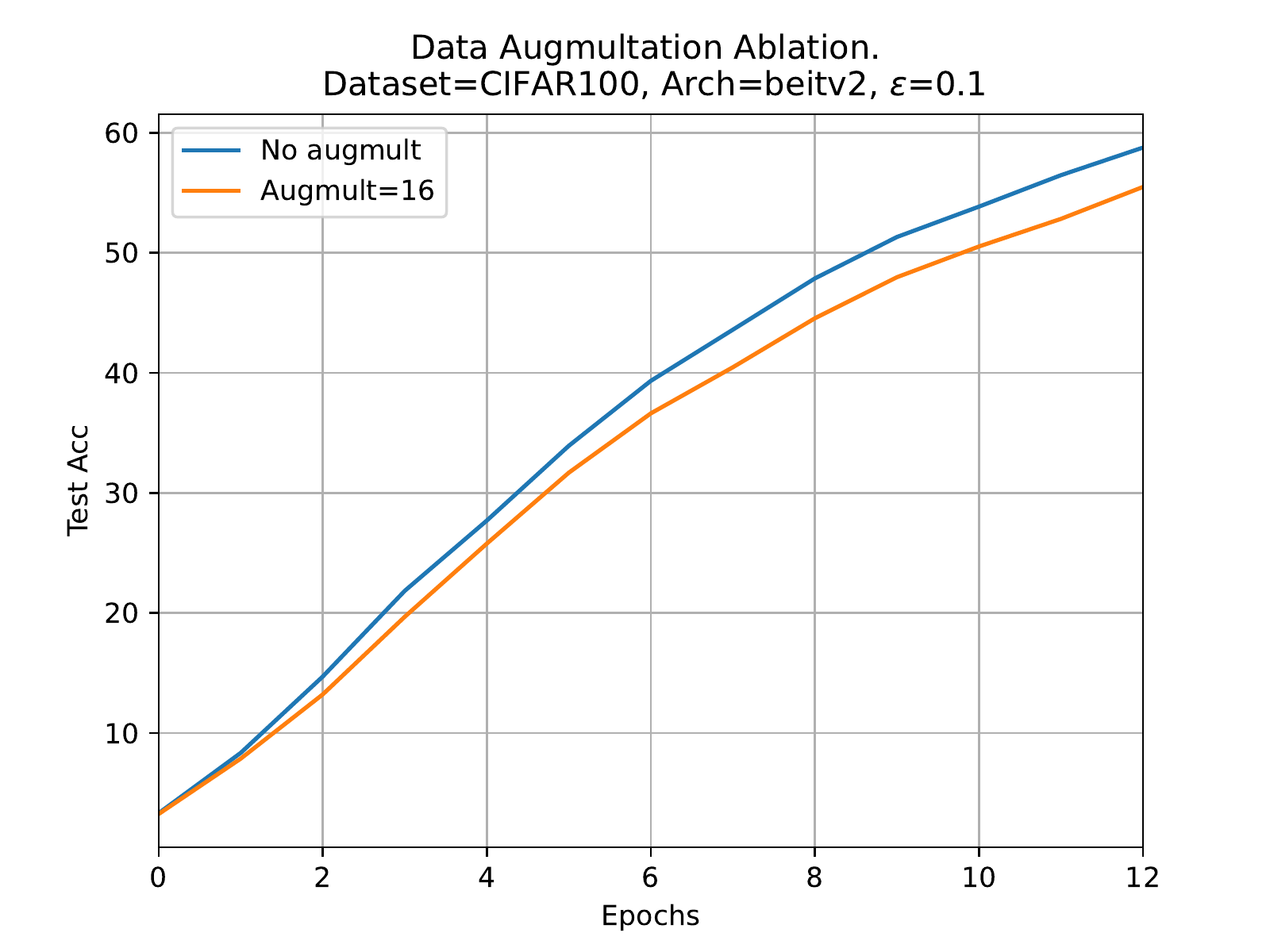}
    \end{minipage}
    }
    \subfigure[CIFAR10 (easy)]{
    \begin{minipage}[t]{0.48\linewidth}
    \centering
    \includegraphics[width=2.8in]{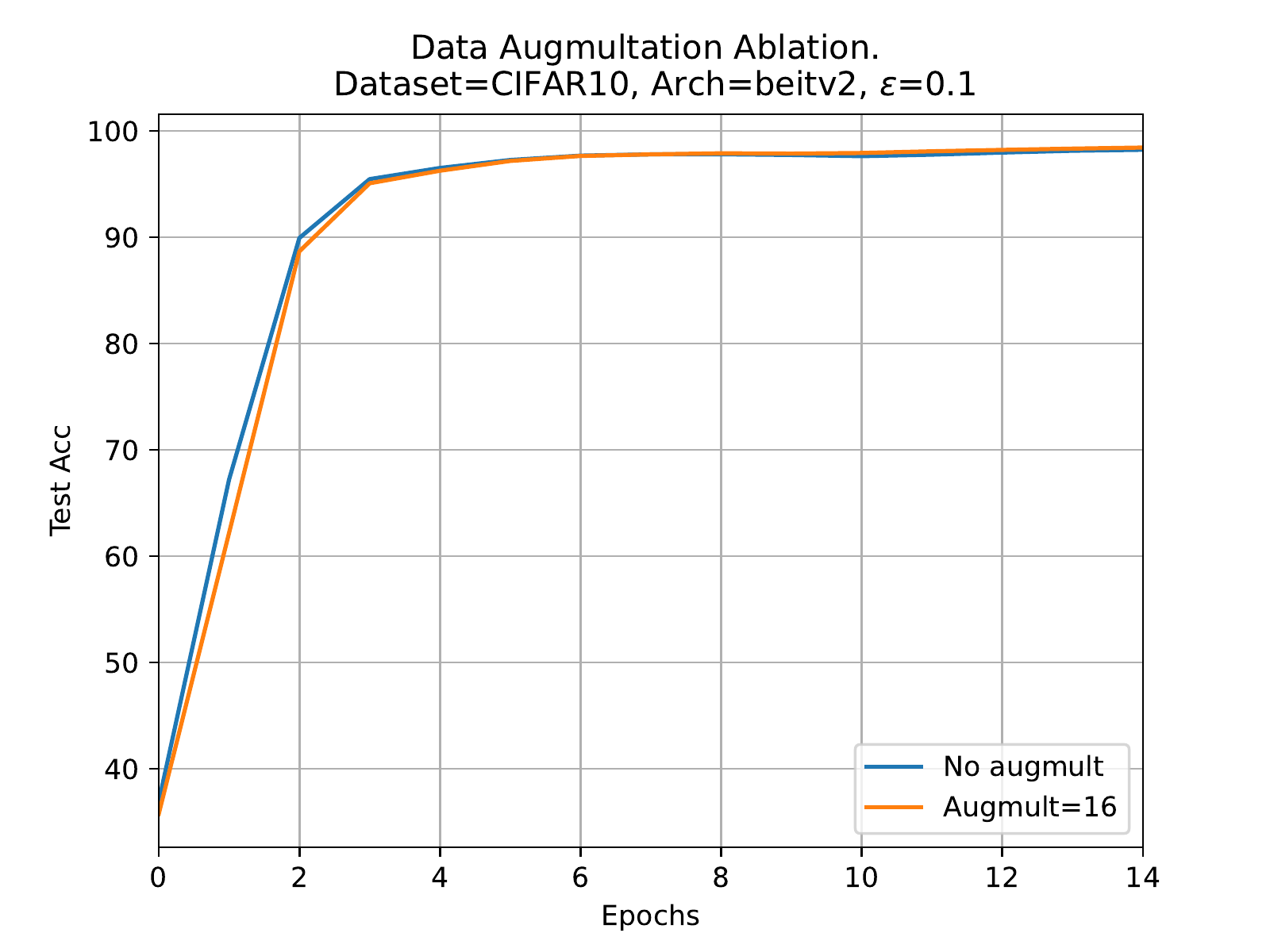}
    \end{minipage}
    }
    \centering
\caption{Ablation of Data Augmultation on two datasets. 
On both datasets, Data Augmultation lags behind the baseline because there is much more training data, and even at the end, Data Augmultation does not have a noticeable improvement.}
\label{fig:ablations:augmultation}
\end{figure*}

\paragraph{Data Augmentation Does Not Work When Freezing Embeddings.}
Data augmentation is used during training to bias the model towards selecting features that are invariant to the rotations we use in the augmentations.
~\citep{geirhosimagenetpriortexture} find that feature extractors pretrained on ImageNet are naturally biased towards texture features.
~\citep{deepmind} eschew traditional data augmentation and instead propose the use of multiple dataset augmentations or "batch augmentation", first introduced by~\citep{hofferaugmult}, to mitigate the variance of DP-SGD.
In Fig. \ref{fig:ablations:augmultation} we ablate the effectiveness of batch augmentation and find that it does not noticeably improve accuracy during transfer learning.
This is because dataset augmentation changes the prior of the model when training the entire network~\citep{Shorten2019dataaugmentation}, but when we freeze all layers but the classifier, the model does not have the capacity to change to optimize for the prior introduced by data augmentation, because the embedding layer is frozen.

\paragraph{Weight Decay Is Not Needed When Freezing Embeddings.}
Regularization methods such as weight decay are commonly used during pretraining to prevent overfitting, and the feature extractors we use are pretrained with AdamW~\citep{vit}.
One of the benefits of weight decay during fine-tuning is limiting the change of the embedding layer to not overfit and thus retain the features learned during pretraining~\citep{kumarfinetunedistort}.
In the ongoing debate on whether to use weight decay during fine-tuning~\citep{touvronweightdecayfinetune}, we submit that weight decay should not be used in private fine-tuning.
In Fig. \ref{fig:ablations:standardization-decay} we ablate a range of values of the weight decay parameter and observe that increasing the weight decay beyond a negligible amount (the gradient norm is $\approx 1e-2$) only decreases accuracy, and no value of the weight decay increases accuracy.
There are two reasons for this.
The first is that we initialize the weights of the model to zero, so we do not expect the gradients to be large.
The second is that we only train the last layer, and therefore there is no need to regularize the training of the embedding layer.
This supports the conclusion of~\citep{kumarfreezefinetune} that SGD with momentum is outperforms AdamW as long as the embedding layer is not updated.

\begin{table*}[htbp]
\centering
\caption{We compare the best private and best non-private test accuracy performances of our method to prior work using models pretrained on ImageNet-21k and fine-tuned on CIFAR10 and CIFAR100. Full results are in Section \ref{sec:evaluation}.}
\begin{tabular}{@{}lccccc@{}}
\toprule
Model & Dataset & $\varepsilon=0.1$ & $\varepsilon=1$ & $\varepsilon=\infty$ & Gap ($1 - \infty$) \\
\midrule
our method & CIFAR10 & $\textbf{98.65}$ & $\textbf{99.00}$ & $99.00$ & $\textbf{0.00}$ \\
 & CIFAR100 & $\textbf{81.9}$ & $\textbf{89.81}$ & $91.57$ & $1.76$ \\
\hline
~\citep{googlefinetune} & CIFAR10 & $95.8$ & $96.3$ & $96.6$ & $0.3$ \\
 & CIFAR100 & $78.5$ & $82.7$ & $85.29$ & $2.59$ \\
 \hline
 ~\citep{bu2022scalable} & CIFAR10 &- & $96.7$ & $97.4$ & $0.7$ \\
 & CIFAR100 &- & $83.0$ & $88.4$ & $5.4$ \\
 \hline
 ~\citep{papernotfinetuning} & CIFAR10 &- & $95.0$ & $96.4$ & $1.4$ \\
 & CIFAR100 &- & $73.7$ & $82.1$ & $8.4$ \\
 \hline
 ~\citep{deepmind} & CIFAR10 &- & $94.8$ & $96.6$ & $1.8$ \\
 & CIFAR100 &- & $67.4$ & $81.8$ & $14.4$ \\
\bottomrule
\end{tabular}
\end{table*}

\subsection{Hyperparameter Ablations}\label{appen:heatmaps}
We provide full heatmaps and pareto frontiers for all datasets and the 3 best performing models (we do not perform a full evaluation on the ViT in order to minimize any knowledge leak for the evaluation of the linear scaling rule with the strategy in ~\citep{googlefinetune}).
We note that while all of these datasets are arguably in-distribution, our focus is on comparing the regime of optimization preferred by our method to those of other works, and this is achieved by producing results on benchmark tasks.
We further note that STL10 is explicitly in-distribution for the pretraining dataset (ImageNet); we only use this dataset as a temporary stand-in for evaluation on ImageNet-1k, a common benchmark in prior work~\citep{googlefinetune} to minimize the computational burden.

\paragraph{Hyperparameter Tuning and Selecting Epsilon.}
Prior work often uses unrealistic values of $\varepsilon$ that provide no real privacy guarantee.
While some prior work makes the case that hyperparameters need to be tuned even for non-private learning and can be chosen beforehand, we show that this is not the case.
Not only are the optimal choices of key hyperparameters different between training from scratch and transfer learning~\citep{haohparamtuning}, they are also different for non-private and private transfer learning~\citep{xuechendpnlp,deepmind}.
We now provide guidelines for selecting $\varepsilon$ and broad intuition behind our choice to design a system that minimizes dependence on hyperparameters.

For a decade the standard values of $\varepsilon$ proposed for privacy preserving statistics queries have fallen in the range of $0.1$ in line with $e^{\varepsilon} \approx 1 + \varepsilon$ for $\varepsilon\ll 1$~\citep{dworkdp}, and recently surveyed DP deployments generally abide by the rule of selecting $\varepsilon \approx 0.1$~\citep{dworkepsilon}.
We know that while all small values of $\varepsilon$ generally behave the same, every large value of $\varepsilon$ is fundamentally different in a unique way~\citep{dworkepsilon}.
In line with these guidelines, we only evaluate $\varepsilon \in [0.01, 1.0]$ and perform most of our ablations on the most challenging task where we can see a range of performance: CIFAR100 for $\varepsilon=0.1$.

\subsection{Theory}\label{appen:theory}
\begin{proposition}
The model training subroutine in \ref{alg:dp-raft} is $(\sqrt{T}/\sigma)$-GDP.
\end{proposition}\label{thm:privacy}

\begin{corollary}\label{thm:corollary}
Algorithm \ref{alg:dp-raft} is $(\epsilon, \Phi(-\epsilon\cdot \sigma/\sqrt{T} + \sqrt{T}/2\sigma)) -e^\epsilon \cdot \Phi(-\epsilon\cdot \sigma/\sqrt{T} - \sqrt{T}/2\sigma))$-DP. Also, for $n$-fold repetition, the algorithm is  $(\epsilon, \Phi(-\epsilon\cdot \sigma/\sqrt{n\cdot T} + \sqrt{n\cdot T}/2\sigma)) -e^\epsilon \cdot \Phi(-\epsilon\cdot \sigma/\sqrt{n\cdot T} - \sqrt{n\cdot T}/2\sigma))$-DP
\end{corollary}
Proof of Proposition~\ref{thm:composition}:
\begin{proof}
Since we are using the full batch, each iteration of the algorithm is an instantiation of the Gaussian mechanism with sensitivity of 1 and Gaussian noise with standard deviation of $\sigma$. 
Hence, each iteration of the mechanism is $(1/\sigma)$-GDP by Theorem 3.7 in \cite{donggdp}. 
Then, since we have the adaptive composition of $T$ of these mechanisms, the algorithm is $(\sqrt{T}/\sigma)$-GDP overall, using the composition theorem for GDP, as stated in Corollary 3.3 in \cite{donggdp}.
\end{proof}
Proof of Corollary~\ref{thm:corollary}:
\begin{proof}
This directly follows from the GDP to DP conversion as stated in Corollary 2.13 in \cite{donggdp}.
\textbf{Why does our HPO have low privacy cost?}
Our HPO has low privacy cost because of the nature of composition under GDP.
Consider one sweep of our method with \(n=3\) that evaluates some \((T_1, \eta_1, \sigma_1), (T_2, \eta_2, \sigma_2))\) and we extrapolate \((T_f, \eta_f, \sigma_f)\), that works out to \(\eps_1=0.1, \eps_2=0.2, \eps_f=0.88\). The composition for this according to~\citep{dong2022gaussian} is \(\mu_f = \sqrt{n\mu_1^{2} + n\mu_2^{2} + \mu_f^{2}}\) for \(\mu_1 = \sqrt{T_1 / \sigma_1^{2}}\). If we convert \(\mu_f\) to \(\eps_f, \delta=1e-5\)-DP, we arrive at a final guarantee of \((1, 1e-5)\)-DP. The cost of HPO here in terms of the privacy utility tradeoff is actually just the marginal utility between \(\eps_f=0.88\) and \(\eps_t=1.0\). As we will show in~\cref{sec:evaluation}, in many cases this marginal utility is negligible, and the value of cheap one-time measures that improve the performance of the rest of training such as HPO is very much worth it due to the nature of composition under GDP.
\end{proof}
Proof of Thm.~\ref{thm:general}
The main idea of the proof is similar to the main result in~\citet{fang2023improved} but is simpler because we only prove the result for linear models.
\begin{proof}
    We first apply~\citep{boydnote} to see that gradient descent with step size $\frac{2}{\beta} > \eta > \frac{2}{\alpha + \beta}$ on a $\alpha$-strongly convex, $\beta$-smooth function is a $\max(1-\eta \beta, 1 - \eta \alpha)$-contraction. 
    Call this latter quantity $c$.\\
    Now consider a sequence of benign updates from gradient descent $w^{t}_{b}$ and a sequence of noisy updates for the same dataset $w^{t}$.
    Given the contractive property of GD
    , we have the following:\\
    \begin{align}\label{eq:contraction}
    \Big|(w^{t}_{b} - \eta \nabla f(w^{t}_{b})) - (w^{t} - \eta \nabla f(w^{(t)}))\Big| \leq c \Big|w^{t}_{b} - w^{t-1}_{b} \Big| 
    \end{align}
    We apply the update rule in \ref{eq:dpsgd} and use Eq.\ref{eq:contraction}
    \begin{align}
        w^{(t+1)} &= w^{(t)} - \eta (\nabla f(w^{(t)}) + \sigma \xi) \\
        \Big |w^{t+1}_{b} - w^{t+1} \Big| &= \\
        &= \Big |w^{t}_{b} - \eta \nabla f(w^{t}_{b}) - w^{(t)} + \eta \nabla f(w^{(t)}) - \sigma \xi \Big| \\
        &\leq c \Big |w^{t}_{b} - w^{(t)} \Big| + \eta \rho
    \end{align}
    Now we have the following
    \begin{align}\label{eq3}
    \Big|w^t - w^t_{b}\Big|\leq c \Big|w^{t-1} - w^{t-1}_{b}\Big| +\rho\eta
    \end{align}
    We now proceed via induction. Assume for $T-1$ the statement of Thm.~\ref{thm:general} holds. By Eq.\ref{eq3} and the induction hypothesis we have
\begin{align}\label{eq4}
\Big|w^{T-1} - w^{T-1}_{b}\Big| \leq \rho \eta \times (\sum_{i}^{T-2} c^i) \\
\Big|w^{T} - w^{T}_{b}\Big|\leq c(\rho \eta \times (\sum_{i}^{T-2} c^i)) + \rho \eta \\
\Big|w^{T} - w^{T}_{b}\Big|\leq \rho \eta \times(\sum_i^{T-1} c^i).
\end{align}

$$ \rho \eta \times(\sum_i^{T-1} c^i) = \frac{\rho \eta(1-c^T)}{1-c} $$

$$\rho \eta \frac{1-c^{T}}{1-c} = \frac{\rho \eta (1-c^T)}{\eta\cdot \min(\alpha,\beta)} = \frac{\rho(1-c^T)}{\min(\alpha,\beta)}$$

The intuition is clear: at iteration 0 there is no divergence. 
At iteration 1 there is $\eta \rho$ divergence.
At iteration 2 the previous divergence contracts by $c$ and increases by $\eta \rho$, so the divergence is $c^1 \eta \rho + \eta \rho$.
At iteration 3 the divergence is $c^2 \eta \rho + c^1 \eta \rho + \eta \rho = \eta \rho (c^2 + c + 1)$.
\end{proof}

It remains to show that the conditions for convexity and smoothness are satisfied for the problem at hand. For the case of, ex, training a single linear layer on top of extracted features with GD, this is easy to prove. We defer to the analysis from~\citet{sparsefed}, which we reproduce here for the reader's convenience.

\begin{example}[Computing the Lipschitz constant for single-layer SGD training (~\citet{sparsefed})]
We compute the coordinatewise Lipschitz constant of the SGD protocol for a single layer neural network defined as $\sigma(\theta x)$, where $\sigma$ is the softmax function and $\theta \in \cR^d$ are the network parameters. For cross-entropy loss-based training using dataset $D$, we show that the constant $c=\frac{1}{4}$. Formally,
\begin{equation}
\sup_{D \in Z, \theta_1, \theta_2 \in \cM} | \cG(\theta_1, D)[ i ]  - \cG(\theta_2, D)[ i ] |_1 \leq \frac{1}{4} | \theta_1 - \theta_2 |_1 \mbox{ for any coordinate index } i \in [d] \nonumber
\end{equation}

Without loss of generality, we assume that dataset $D$ is comprised of samples of the form $(x, y)$, where $x \in [0, 1]^m$, and $y \in \{0, 1\}^C$ is the one-hot encoded representation of any of the $C$ classes. 
For the single layer neural network, the model parameters are denoted by $\theta \in \cR^{C \times m}$, and the softmax layer by the function $\sigma(\cdot)$. The neural network can thus be represented as $\Phi(x, \theta) = \sigma(\theta x)$.

We define $g(\theta, x) = \frac{\partial \mathcal{L}(\Phi(x, \theta), y)}{\partial \theta}$ where $\mathcal{L}$ is the softmax cross entropy loss function. For the SGD protocol, $\cA(u) = u$, and $\cG(\theta, D) = g(\theta, x)$. Our goal is to find a Lipschitz constant $L$ such that, for all indices $i \in [C]$ and $j \in [m]$,
\begin{equation}
\sup_{x \in D, \theta_1, \theta_2} \frac{| g(\theta_1, x)_{ij} - g(\theta_2, x)_{ij} |_1}{| \theta_1 - \theta_2 |_1} \leq L
\label{eqn:lc_0}
\end{equation}
We define intermediate variable $z = \theta x$ and the neural network output distribution $p = \sigma(z)$, such that both $p, z \in \mathbb{R}^C$. Note, for a given target class $t$, the cross entropy loss function $\mathcal{L}(p, y) = -\log {p_t}$ where $p_t = \frac{e^{z_t}}{\sum_{j}e^{z_j}}$. Thus, 
\begin{equation}
    g(\theta, x)_{ij} = \frac{\partial \mathcal{L}}{\partial \theta_{ij}} = \sum_{c=1}^{C} \frac{\partial \mathcal{L}}{\partial z_c} \frac{\partial z_c}{\partial \theta_{ij}}. \label{eqn:lc_1}
\end{equation}
Computing the terms of \eqref{eqn:lc_1}, we have
$\frac{\partial \mathcal{L}}{\partial z_c} = p_t - 1 \mbox{ for } c=t \mbox {; and } \frac{\partial \mathcal{L}}{\partial z_c} = p_c \mbox{ otherwise; }$ 
and $\frac{\partial z_c}{\partial \theta_{{ij}}} = x_j$. Thus, 
\begin{eqnarray}
g(x, \theta)_{{ij}} & = & x_j (p_t - 1) \;\; \mbox{ for i = t} \nonumber \\ 
& = & x_j p_i  \;\; \mbox{ for } i \neq t \label{eqn:lc_2}
\end{eqnarray}
We compute the Hessian of $g(x, \theta)_{ij}$ as:
\begin{eqnarray}
\frac{\partial g(x, \theta)_{ij}}{\partial \theta_{kl}} & = &  x_j p_t (1-p_t) x_l \;\; \mbox{ for } k = t \nonumber \\
& = & x_j p_k (1 - p_k) x_l \;\; \mbox{ for } k \neq t \label{eqn:lc_3}
\end{eqnarray}
where $k \in [C], l \in [m]$. The maximum value of the Hessian in \eqref{eqn:lc_3}, occurs at $x_j = x_l = 1$, and $p_t = p_k = \frac{1}{2}$. Thus, 
\begin{eqnarray}
\max_{i, j, k, l}\frac{\partial g(x, \theta)_{{ij}}}{\partial \theta_{{kl}}} & \leq &  \frac{1}{4} \;\; \mbox{ for } k = t \nonumber \\
& \leq & \frac{1}{4} \;\; \mbox{ for } k \neq t \label{eqn:lc_4}
\end{eqnarray}
To obtain the Lipschitz constant, we first define the function
\[ h(t) = g((1-t)\theta_1 + t \theta_2, x)_{ij} \mbox{ where } t \in [0, 1] \]
Thus, $h(0) = g(\theta_1, x)_{ij}$ and $h(1) = g(\theta_2, x)_{ij}$. Since, the function $h(t)$ is differentiable everywhere in $(0, 1)$, using Mean Value Theorem, we know that there exists a point $t^{*} \in (0, 1)$ such that:
\begin{equation}
 h(1) - h(0) \leq h'(t^{*}) 
\mbox{ where } h'(t) = (\theta_2 - \theta_1) g'((1-t) \theta_1 + t \theta_2, x)_{ijkl}. \label{eqn:lc_5}
\end{equation}
Rewriting \eqref{eqn:lc_0}, we get
\begin{eqnarray}
& & \sup_{x \in D, \theta_1, \theta_2} |g(\theta_1, x) - g( \theta_2, x) |_1 \nonumber \\
& \leq & \sup_{x \in D, \theta_1, \theta_2} |\max_{i, j} \{ g( \theta_1, x)_{ij} - g(\theta_2, x)_{ij} \} |_1 \nonumber 
\end{eqnarray}
Let $i^{*}, j^{*}$ correspond to the indices where the maximum in the above equation occurs. Combining \eqref{eqn:lc_4} and \eqref{eqn:lc_5}, we get:
\begin{equation}
    \sup_{x \in D, \theta_1, \theta_2} | g(\theta_1, x)_{i^{*} j^{*}} - g(\theta_2, x)_{i^{*} j^{*}} |_1 \leq \frac{1}{4} |\theta_1 - \theta_2|_1 \label{eqn:lc_6}
\end{equation}
Comparing \eqref{eqn:lc_6} with \eqref{eqn:lc_0} we get $c = \frac{1}{4}$.
\end{example}

\section{Furthur Results for Language Modeling Tasks}\label{appen:lm}
In general it is not feasible to do full-batch experiments for the NLP tasks because the memory requirements of LLMs are very large. We therefore do the composition with the PoissonSubsampledGaussianMechanism class in the PLD accountant~\citep{glwmsaccountant}, ensuring that our method still accounts for the privacy cost of HPO. 
\subsection{Related Works}
\citet{xuechendpnlp} provide methods for fine-tuning large language models under DP-SGD by proposing new clipping methods to mitigate the memory burden of per-sample gradient clipping.
However, they do not achieve performance comparable to non-private models when fine-tuning a pretrained model on the PersonaChat dataset.
We adapt their techniques to the hyperparameter settings that we show are optimal for DP fine-tuning, and produce similar performance to non-private fine-tuning on the PersonaChat dataset.
~\citet{yu2021large} report compelling results by only updating a sparse subset of the LLMs with LoRA~\citep{hu2021lora}.
We fine-tune GPT2 and RoBerta;~\citet{basu2022benchmarking} also fine-tune BERT models.

\subsection{Experimental Set-up for Finetuning Language Models}
\label{appen:hyperparam}
\paragraph{Persona-Chat: } We write code based on winners of ConvAI2 competition\footnote{\href{https://github.com/huggingface/transfer-learning-conv-ai}{https://github.com/huggingface/transfer-learning-conv-ai}.} and  private-transformers library.\footnote{\href{https://github.com/lxuechen/private-transformers}{https://github.com/lxuechen/private-transformers}.} We first do clipping norm $[0.1,0.2,0.5,1.0]$, learning rate in $[2,5,10, 20, 50]\times 10^{-5}$, batch size $64$ and epochs $[3,10,20]$ at $\varepsilon=3$ and $\varepsilon=8$ and find that the clipping norm in this range achieves almost same perplexity with other hyperparams fixed. We then do hyperparameter tuning as reported in Table~\ref{tab:hyperparamchat} to finetune GPT-2.
\begin{table*}[htbp]
    \centering
    \caption{Set of hyper-parameters used in the finetuning GPT-2.}
    \begin{tabular}{cc}
    \toprule
         Parameter &Values\\
         \midrule
         Clipping Norm &$0.1$  \\
         Learning Rate & $[2,5,10, 20, 50, 100]\times 10^{-5}$\\
         Batch Size & $[64, 128, 256, 512, 1024, 2048]$ \\
         Epochs & $[3, 10, 20]$\\
    \bottomrule 
    \end{tabular}

    \label{tab:hyperparamchat}
\end{table*}

\paragraph{WikiText-2: } We write code based on the HuggingFace transformers library GPT-2 example,\footnote{\href{https://github.com/huggingface/transformers/blob/main/examples/pytorch/language-modeling/run_clm_no_trainer.py}{HuggingFace transformers GPT-2 example code.}} source code by~\citep{shi2022just}\footnote{\href{https://github.com/wyshi/sdp_transformers.
}{https://github.com/wyshi/sdp\_transformers}} and private-transformers library. The hyperparameter range for grid search is reported in Table~\ref{tab:hyperparamwiki}.

\begin{table*}[htbp]
    \centering
    \caption{Set of hyper-parameters for grid search to finetune GPT-2 on WikiText-2. $\delta=10^{-6}$.}
    \begin{tabular}{cc}
    \toprule
         Parameter &Values\\
         \midrule
         Clipping Norm &$1$  \\
         Batch Size & 2048 (Full Batch) \\
         Epochs & 20\\         
         Learning Rate for $\varepsilon=0.2$ & $[2,4,6,8,10]\times 10^{-4}$\\         
         Learning Rate for $\varepsilon=0.5$& $[0.8,1,2]\times 10^{-3}$\\

    \bottomrule 
    \end{tabular}

    \label{tab:hyperparamwiki}
\end{table*}
\paragraph{Enron Email: } For Enron email dataset, we use the preprocessed dataset in~\citep{gupta2022recovering}, where the non-private baseline of finetuned GPT-2 on this dataset is $7.09$. The hyperparameter range for grid search is reported in Table~\ref{tab:hyperparamenron}.

\begin{table*}[htbp]
    \centering
    \caption{Set of hyper-parameters for grid search to finetune GPT-2 on Enron Email dataset. $\delta=\frac{1}{2|D_{\text{train}}|}$. }
    \begin{tabular}{cc}
    \toprule
         Parameter &Values\\
         \midrule
         Clipping Norm &$1$  \\
         Batch Size & 1024\\
         Epochs & 5\\         
         Learning Rate for $\varepsilon=0.1$ & $[2,3,4,5,6,7,8,9,10]\times 10^{-4}$\\
         Learning Rate for $\varepsilon=0.2$& $[0.6,0.8,1,2,3,4,6,7]\times 10^{-3}$\\
         Learning Rate for $\varepsilon=0.5$& $[0.4, 0.6,0.8,0.9,1,1.1,1.2,1.3, 1.4, 1.5, 1.6, 1.8,2]\times 10^{-2}$\\
         Learning Rate for $\varepsilon=1.0$& $[1,2,3,4,5,6,7,8]\times 10^{-2}$\\
         Learning Rate for $\varepsilon=2.0$& $[2,3,4,5,6,7,8,9,10]\times 10^{-2}$\\
         Learning Rate for $\varepsilon=3.0$ & $[0.6, 0.7, 0.8, 0.9, 1.0, 1.1, 1.2, 1.3, 1.4, 1.6, 1.8, 2.0]\times 10^{-1}$\\

    \bottomrule 
    \end{tabular}

    \label{tab:hyperparamenron}
\end{table*}

\subsection{Additional Results on Persona-Chat}
\label{appen:chat}
\begin{figure*}[htbp]
    \centering
    \subfigure[$\varepsilon=3$]{
    \begin{minipage}[t]{0.45\linewidth}
    \centering
    \includegraphics[width=2.3in]{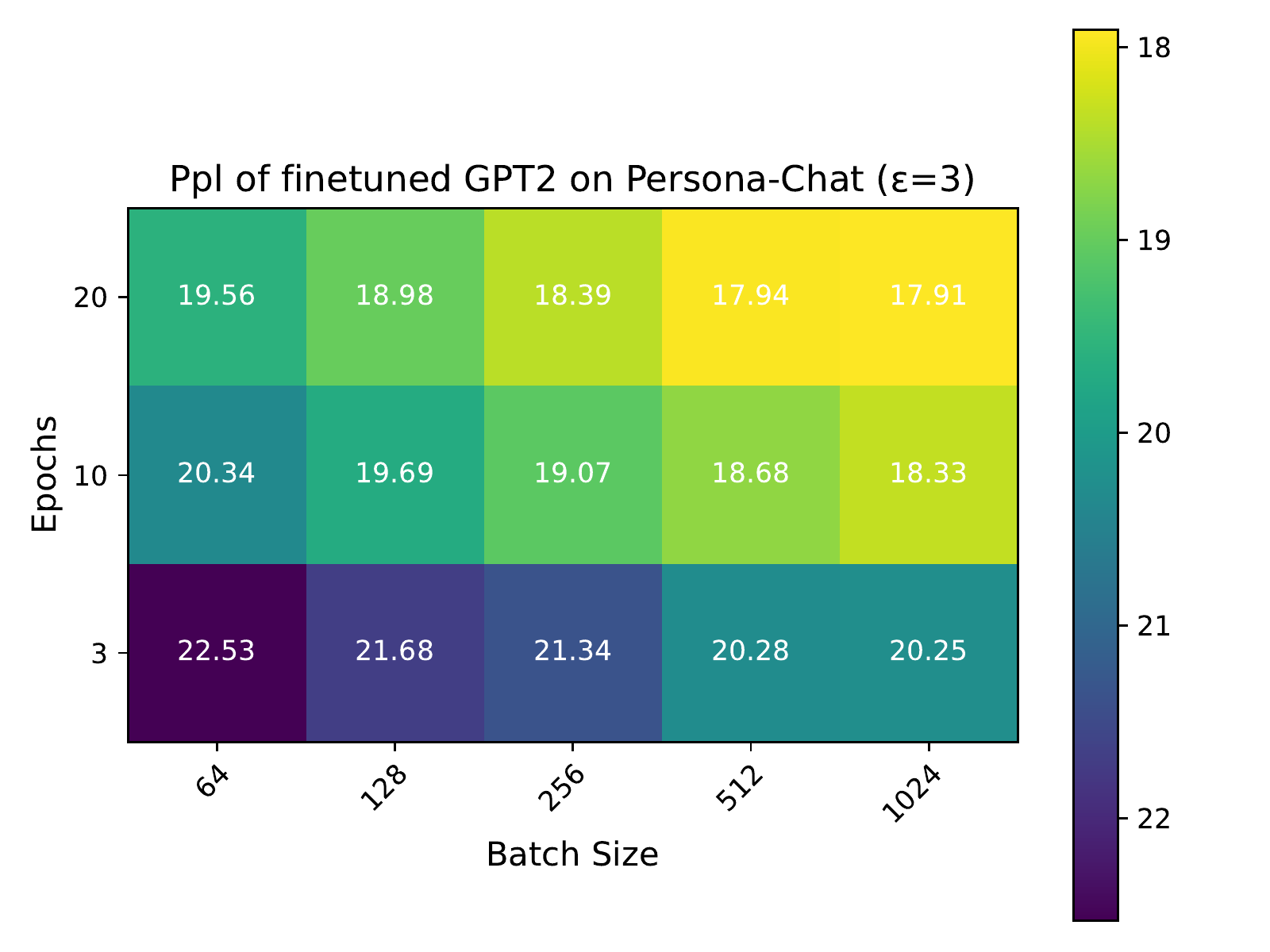}
    \end{minipage}
    }
    \subfigure[$\varepsilon=8$]{
    \begin{minipage}[t]{0.45\linewidth}
    \centering
    \includegraphics[width=2.3in]{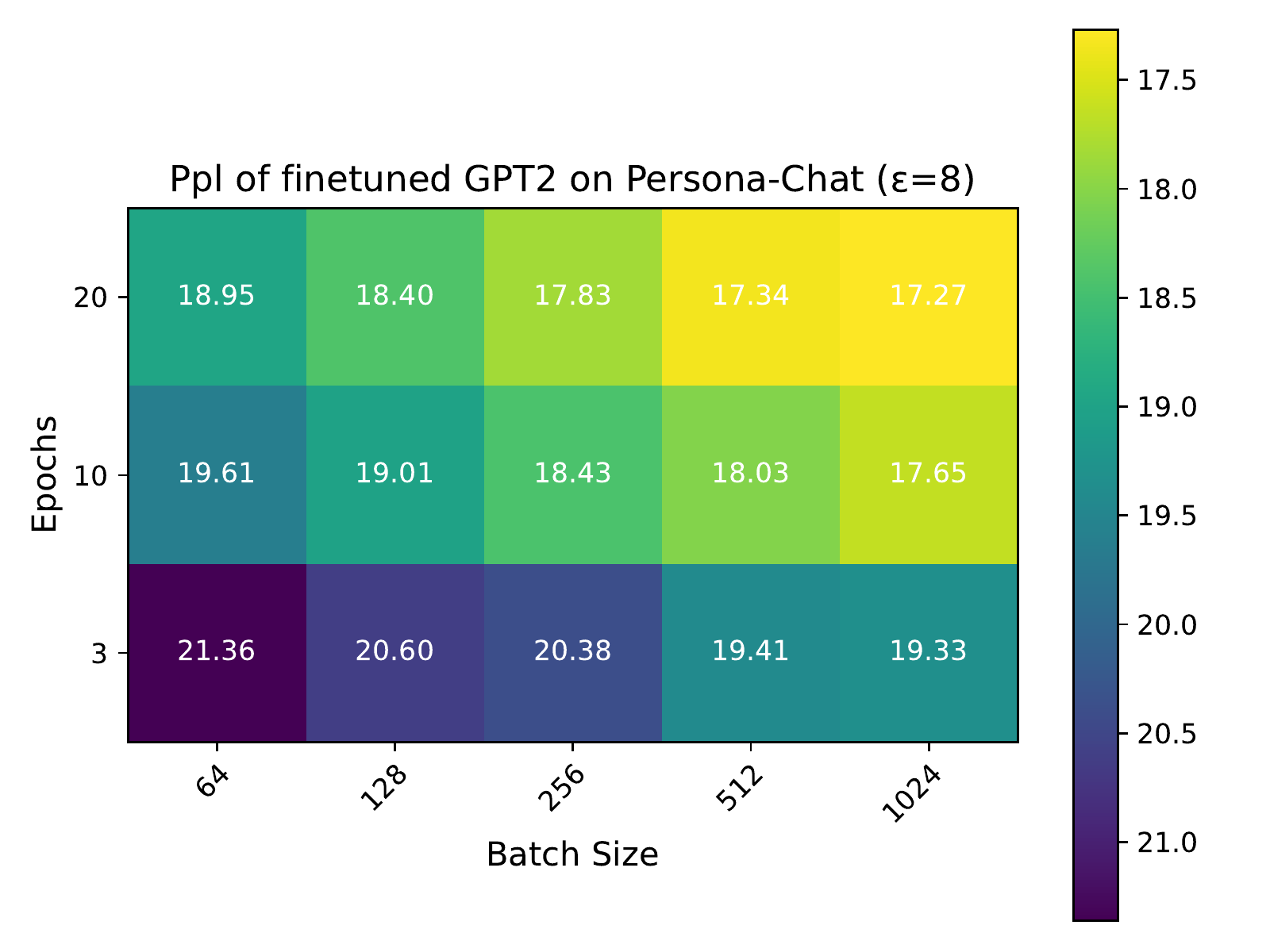}
    \end{minipage}
    }
    \centering
\caption{Comparison of perplexity at different batch size and epochs of GPT-2 on Persona-Chat dataset.}
\label{fig:evsbs}
\end{figure*}

We report the  perplexity of GPT-2 on the Persona-Chat dataset at different epochs and batch size in Figure~\ref{fig:evsbs} (with tuned learning rate in Table~\ref{tab:hyperparamchat}) and we can see that larger batch size and longer epochs can achieve better perplexity, which is consistent with our linear scale rule. Besides, we also investigate fine-tuning multiple layers. With letting the embedding layer and last LayerNorm layer in transformer trainable, we consider fine-tuning only last block in transformer, first and last block in transformer and report the result in Table~\ref{tab:layer} and we can see that the best perplexity is achieved by fine-tuning the whole model.

\begin{table}[htbp]
    \caption{Finetuning GPT-2 on Persona-Chat dataset including full model and different layers of model. We also include non-private baseline.}
    \begin{center}
    \begin{tabular}{cccc}
    \toprule
    $\varepsilon$  & $3$ & $8$  \\
    Full &$17.91$ &$17.27$\\
    Last Block& $19.80$ &$19.20$\\
    First-Last-Block  &$18.93$ &$18.26$  \\
    
    \bottomrule
    \end{tabular}
    \end{center}
    \label{tab:layer}
\end{table}

\subsection{Addtional Results on WikiText-2}
\label{appen:wiki}
\begin{figure*}[htbp]
    \centering
    \subfigure[Pareto Frontier for $\varepsilon$ vs Test Perplexity]{
    \begin{minipage}[t]{0.48\linewidth}
    \centering
    \includegraphics[width=2.5in]{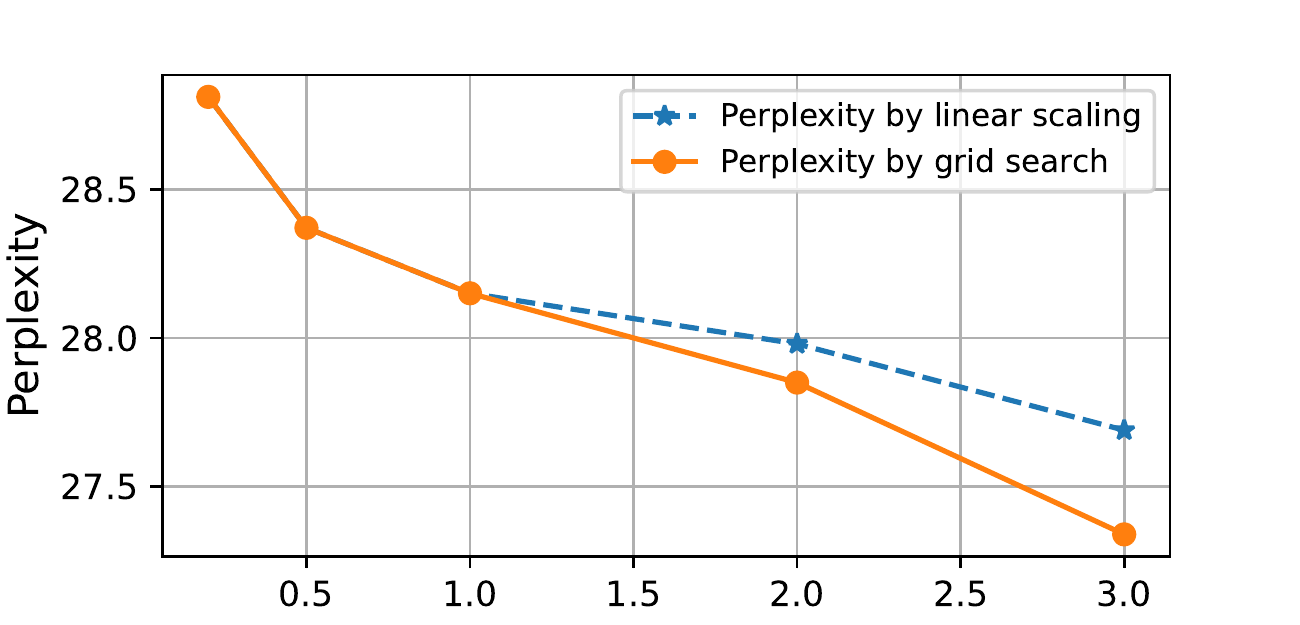}
    \end{minipage}
    }
    \subfigure[Pareto Frontier for $\varepsilon$ vs Total Step Size]{
    \begin{minipage}[t]{0.48\linewidth}
    \centering
    \includegraphics[width=2.5in]{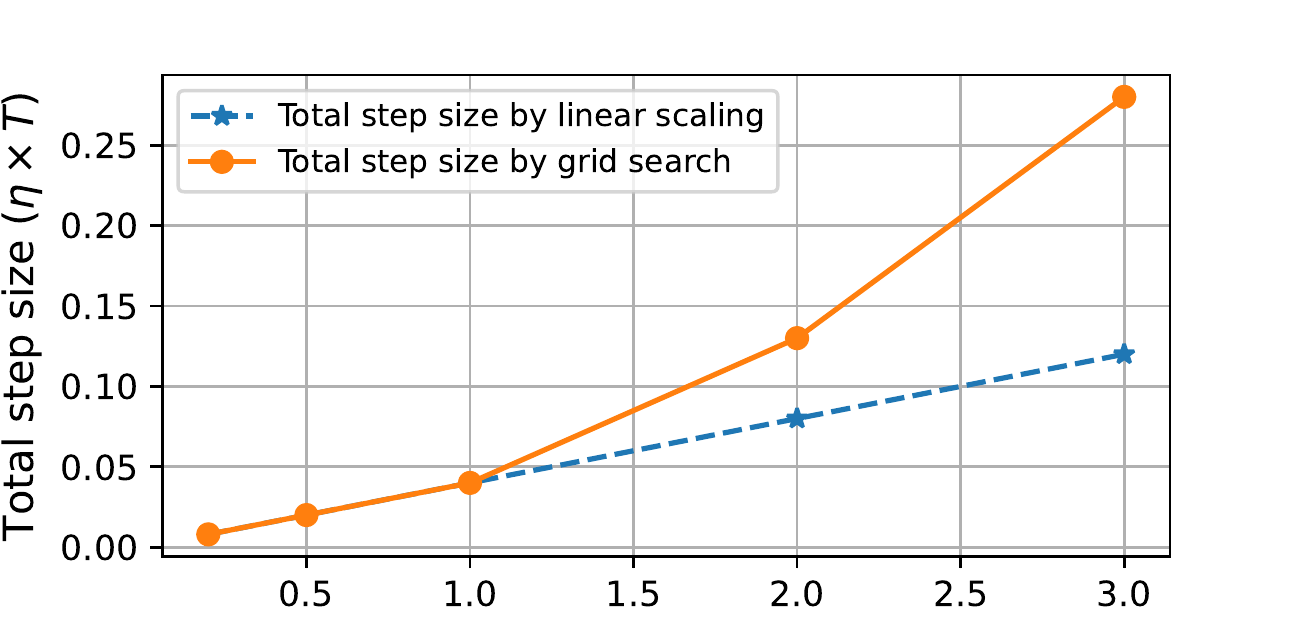}
    \end{minipage}
    }
    \centering
\caption{The linear scaling rule (accounting for the privacy cost of hyperparameter tuning) is competitive with grid search (non-private, doing N trials each with the given $\varepsilon$) in range $[0.2, 1.0]$ on the WikiText-2 dataset. Left: y-axis is Perplexity (lower is better).}
\label{fig:wiki}
\end{figure*}

\begin{figure*}[htbp]
    \centering
    \subfigure[CIFAR10 Beitv2]{
   \begin{minipage}[t]{1\linewidth}
    \centering
    \includegraphics[width=6.6in]{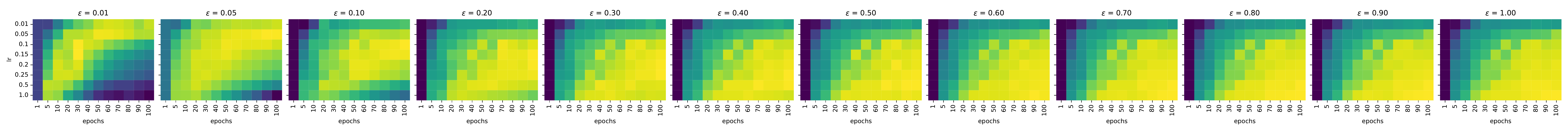}
    \end{minipage}
    }
    \subfigure[CIFAR10 Beit]{
   \begin{minipage}[t]{1\linewidth}
    \centering
    \includegraphics[width=6.6in]{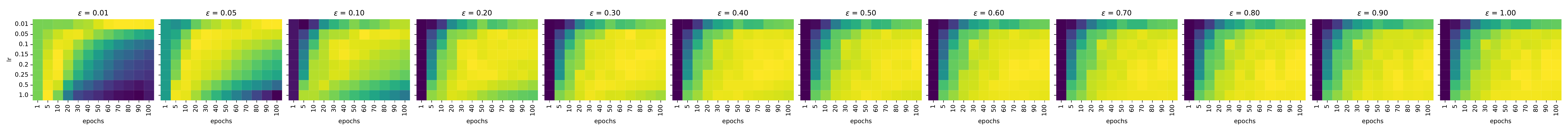}
    \end{minipage}
    }
    \subfigure[CIFAR10 Convnext]{
   \begin{minipage}[t]{1\linewidth}
    \centering
    \includegraphics[width=6.6in]{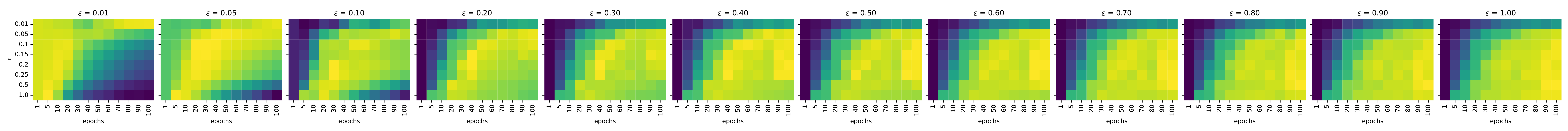}
    \end{minipage}
    }
    \subfigure[CIFAR100 Beitv2]{
   \begin{minipage}[t]{1\linewidth}
    \centering
    \includegraphics[width=6.6in]{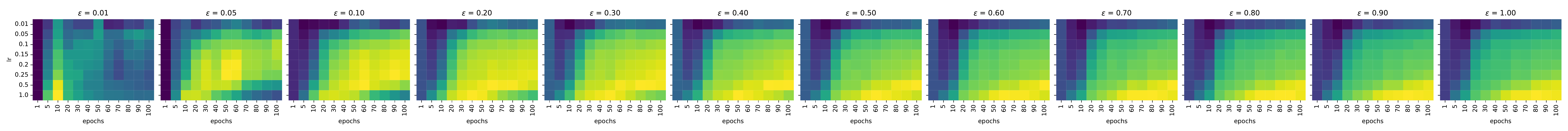}
    \end{minipage}
    }
    \subfigure[CIFAR100 Beit]{
   \begin{minipage}[t]{1\linewidth}
    \centering
    \includegraphics[width=6.6in]{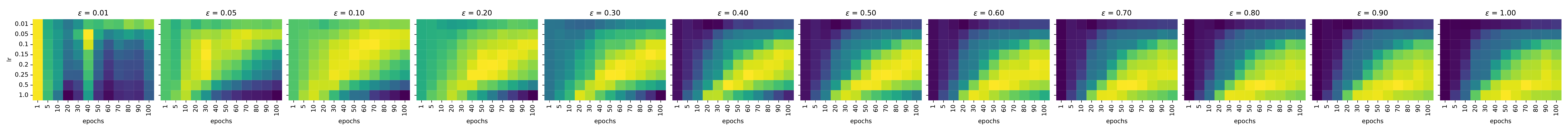}
    \end{minipage}
    }
        \subfigure[CIFAR100 Convnext]{
   \begin{minipage}[t]{1\linewidth}
    \centering
    \includegraphics[width=6.6in]{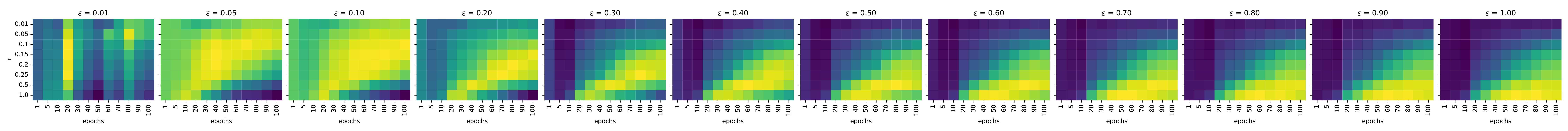}
    \end{minipage}
    }
    \centering
\caption{Heatmaps for the reported datasets and architectures; lighter is better. Note that the scale of the axes differs from the heatmaps in the main body; this will be fixed in a future update.
$\varepsilon$ increases left to right with a different value for each heatmap according to: \([0.01, 0.05, 0.1, 0.2, 0.3, 0.4, 0.5, 0.6, 0.7, 0.8, 0.9, 1.0]\), epochs increase from left to right on the x-axis of each heatmap according to: \([1, 5,10,20,30,40,50,60,70,80,90,100]\), and the learning increases from top to bottom on the y-axis of each heatmap according to: \([0.01, 0.05, 0.1, 0.15, 0.2, 0.25, 0.5, 1.0]\). As $\varepsilon$ increases, left to right, the optimal hyperparameters trend towards longer training with lower learning rates (bottom right).}
\label{fig:heatmaps-full-a}
\end{figure*}

We run the grid-search experiment for $\varepsilon \in \{0.2, 0.5, 1, 2,3\}$ to evaluate the performance gap between the optimal total step size and the estimated total step size.\footnote{Due to the limit of computation resources, all experiments are done by training for 20 iterations. Further increasing the number of iterations will help improve the utility as shown by previous study~\citep{xuechendpnlp, shi2022just}, we leave longer iterations for further study.}) and present the result in Figure~\ref{fig:wiki}. The linear rule scales well from $\varepsilon \in \{ 0.2, 0.5\}$ to $\varepsilon=1$. Though for $\varepsilon \in\{2,3\}$ the perplexity of total step size by linear scale rule is slightly higher than the optimal perplexity of total step size by grid search, the result by linear scale is better than previous SOTA~\citep{shi2022just}, which is $28.84$ at $(\varepsilon=3,\delta=10^{-6})$ by training 20 iterations.

\begin{figure*}[htbp]
    \centering
    \subfigure[STL10 Beitv2]{
   \begin{minipage}[t]{1\linewidth}
    \centering
    \includegraphics[width=6.6in]{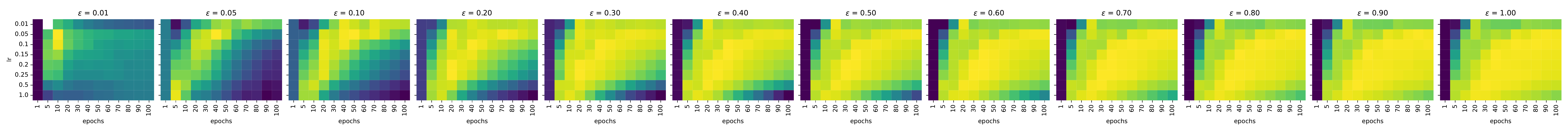}
    \end{minipage}
    }
    \subfigure[STL10 Beit]{
   \begin{minipage}[t]{1\linewidth}
    \centering
    \includegraphics[width=6.6in]{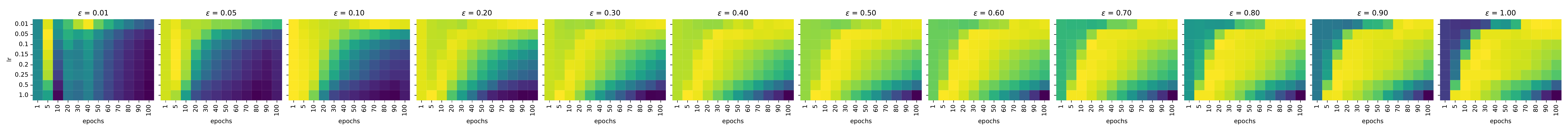}
    \end{minipage}
    }
    \subfigure[STL10 Convnext]{
   \begin{minipage}[t]{1\linewidth}
    \centering
    \includegraphics[width=6.6in]{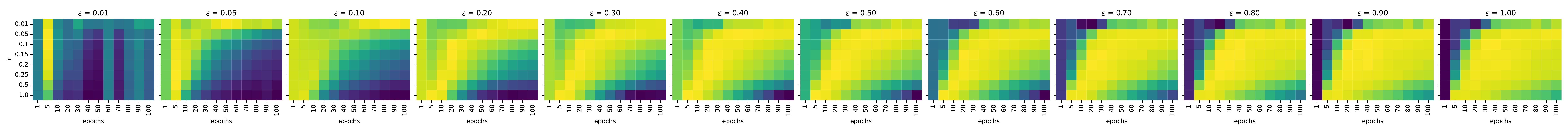}
    \end{minipage}
    }
    \subfigure[FashionMNIST Beitv2]{
   \begin{minipage}[t]{1\linewidth}
    \centering
    \includegraphics[width=6.6in]{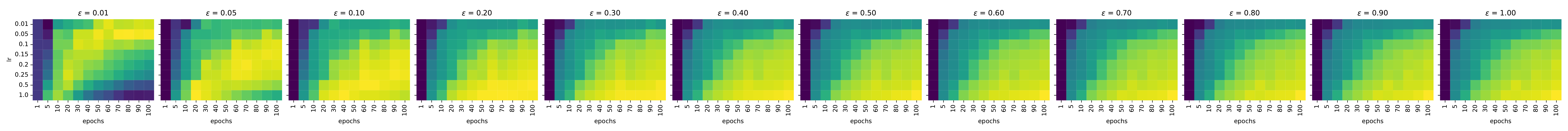}
    \end{minipage}
    }
    \subfigure[FashionMNIST Beit]{
   \begin{minipage}[t]{1\linewidth}
    \centering
    \includegraphics[width=6.6in]{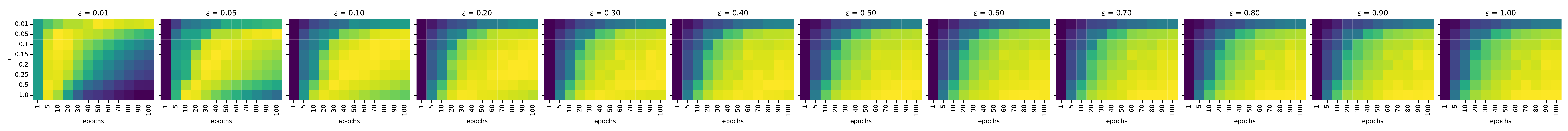}
    \end{minipage}
    }
    \subfigure[FashionMNIST Convnext]{
   \begin{minipage}[t]{1\linewidth}
    \centering
    \includegraphics[width=6.6in]{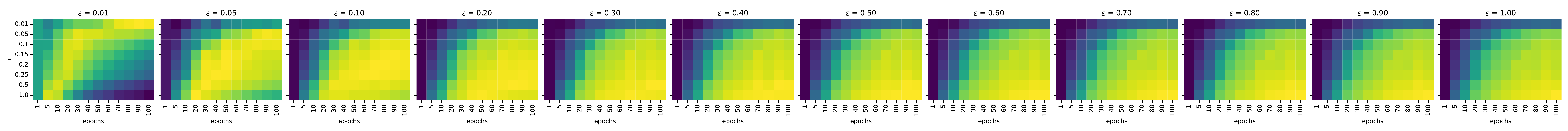}
    \end{minipage}
    }
    \centering
\caption{Heatmaps for the reported datasets and architectures; lighter is better. Note that the scale of the axes differs from the heatmaps in the main body; this will be fixed in a future update.
$\varepsilon$ increases left to right with a different value for each heatmap according to: \([0.01, 0.05, 0.1, 0.2, 0.3, 0.4, 0.5, 0.6, 0.7, 0.8, 0.9, 1.0]\), epochs increase from left to right on the x-axis of each heatmap according to: \([1, 5,10,20,30,40,50,60,70,80,90,100]\), and the learning increases from top to bottom on the y-axis of each heatmap according to: \([0.01, 0.05, 0.1, 0.15, 0.2, 0.25, 0.5, 1.0]\). As $\varepsilon$ increases, left to right, the optimal hyperparameters trend towards longer training with lower learning rates (bottom right).}
\label{fig:heatmaps-full-b}
\end{figure*}

\begin{figure*}[htbp]
    \centering
        \subfigure[CIFAR100 Test Accuracy]{
    \begin{minipage}[t]{0.48\linewidth}
    \centering
    \includegraphics[width=2.8in]{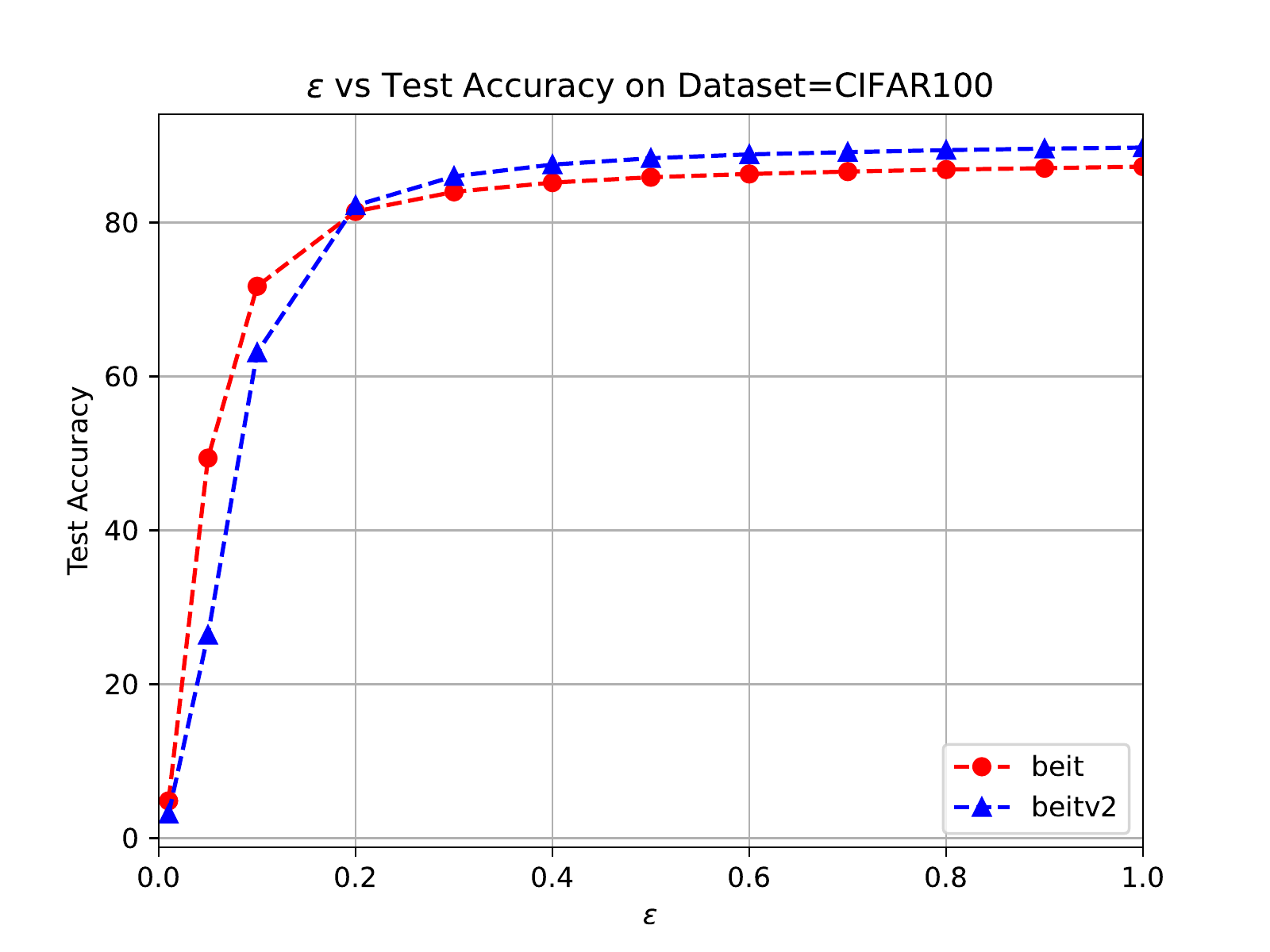}
    \end{minipage}
    }
    \subfigure[CIFAR100 Total Step Size]{
    \begin{minipage}[t]{0.48\linewidth}
    \centering
    \includegraphics[width=2.8in]{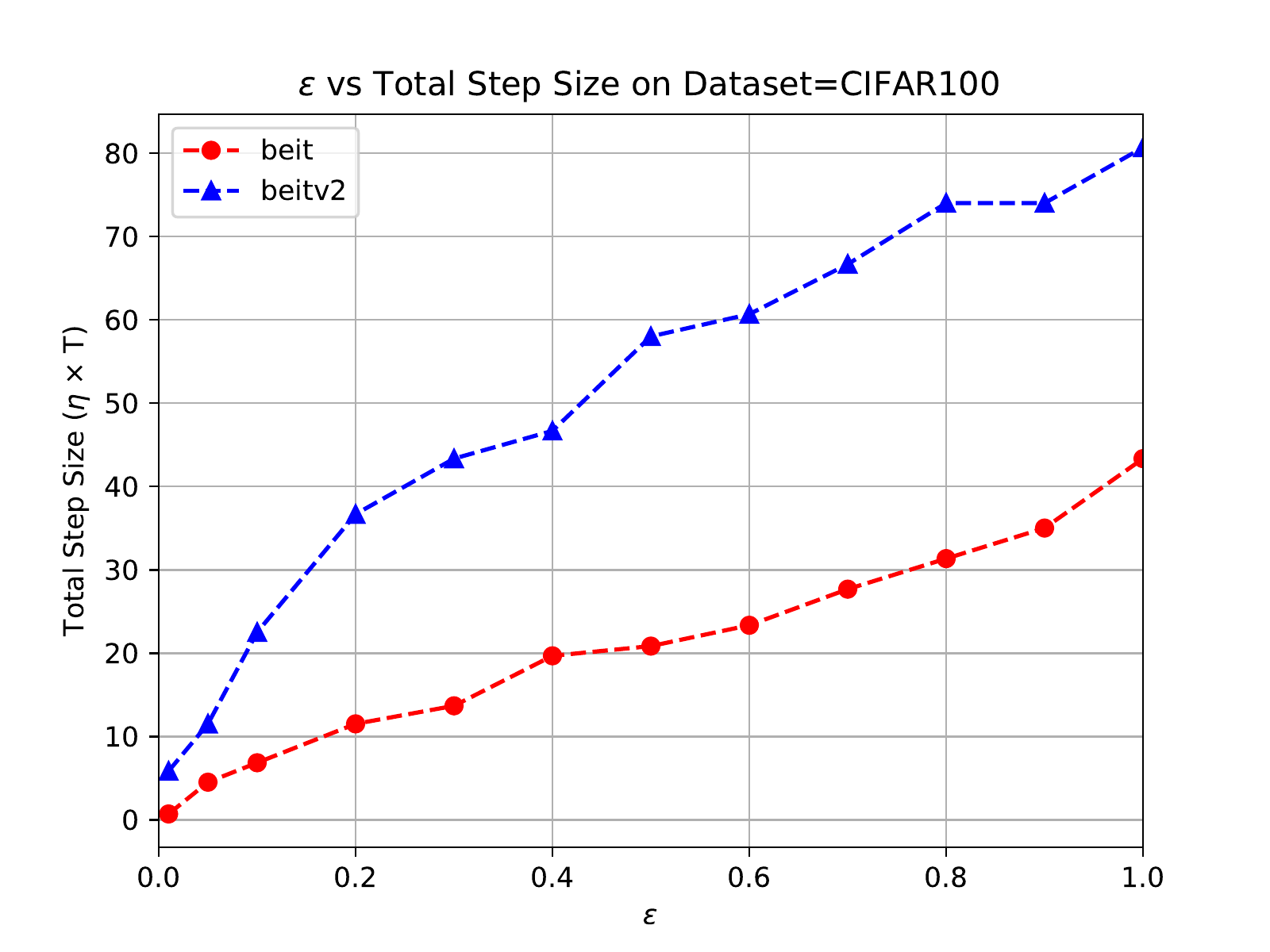}
    \end{minipage}
    }
    \subfigure[CIFAR10 Test Accuracy]{
    \begin{minipage}[t]{0.48\linewidth}
    \centering
    \includegraphics[width=2.8in]{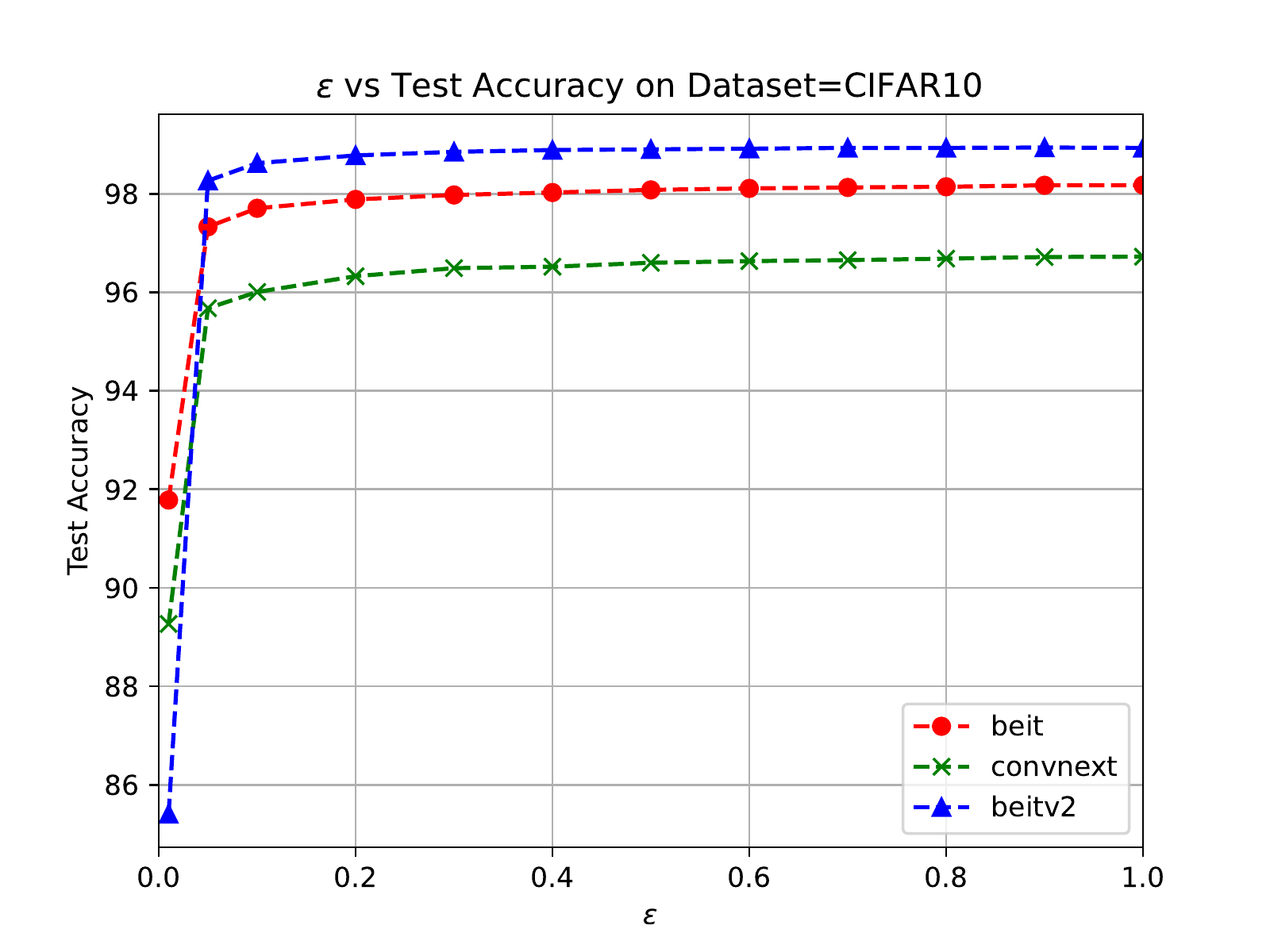}
    \end{minipage}
    }
    \subfigure[CIFAR10 Total Step Size]{
    \begin{minipage}[t]{0.48\linewidth}
    \centering
    \includegraphics[width=2.8in]{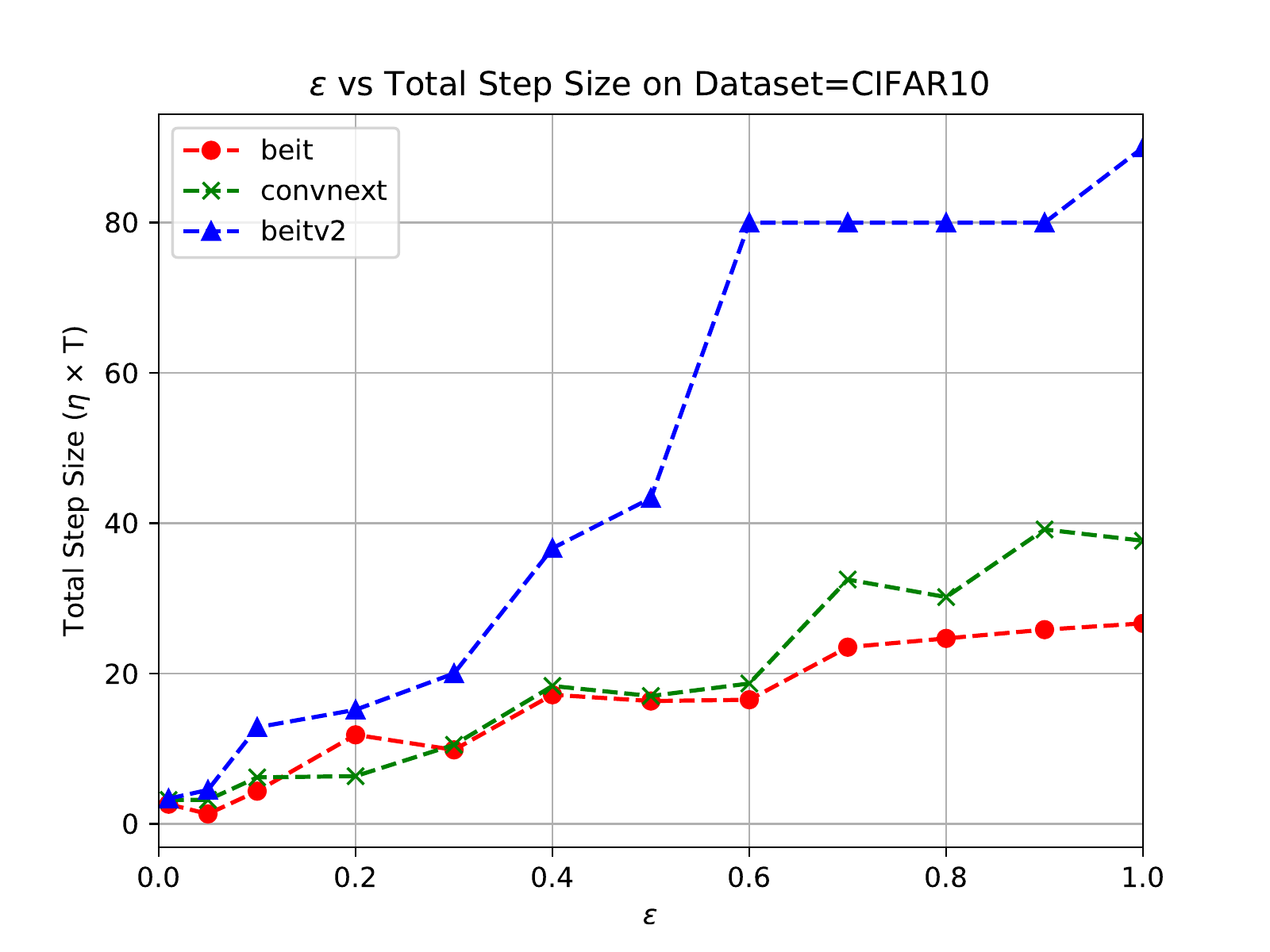}
    \end{minipage}
    }
    \centering
\caption{Pareto frontier for $\varepsilon$ vs test accuracy and total step size for CIFAR10, and CIFAR100. Beitv2 excels for larger values of $\varepsilon$ but beit and convnext are better for smaller values of $\varepsilon$. The inflection point varies across datasets.}
\label{fig:pareto-full-a}
\end{figure*}

\begin{figure*}[htbp]
    \centering
    \subfigure[FashionMNIST Test Accuracy]{
    \begin{minipage}[t]{0.48\linewidth}
    \centering
    \includegraphics[width=2.8in]{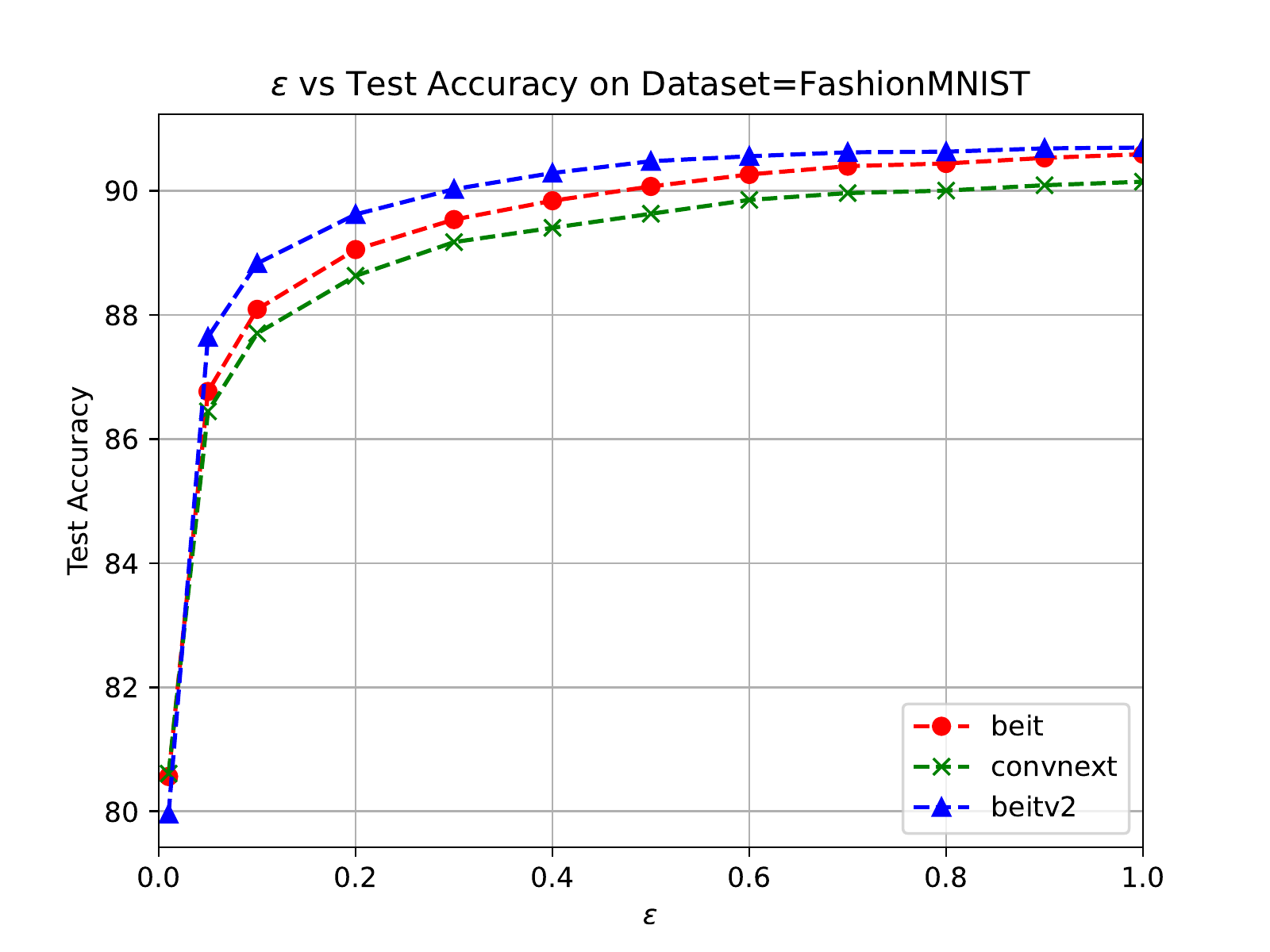}
    \end{minipage}
    }
    \subfigure[FashionMNIST Total Step Size]{
    \begin{minipage}[t]{0.48\linewidth}
    \centering
    \includegraphics[width=2.8in]{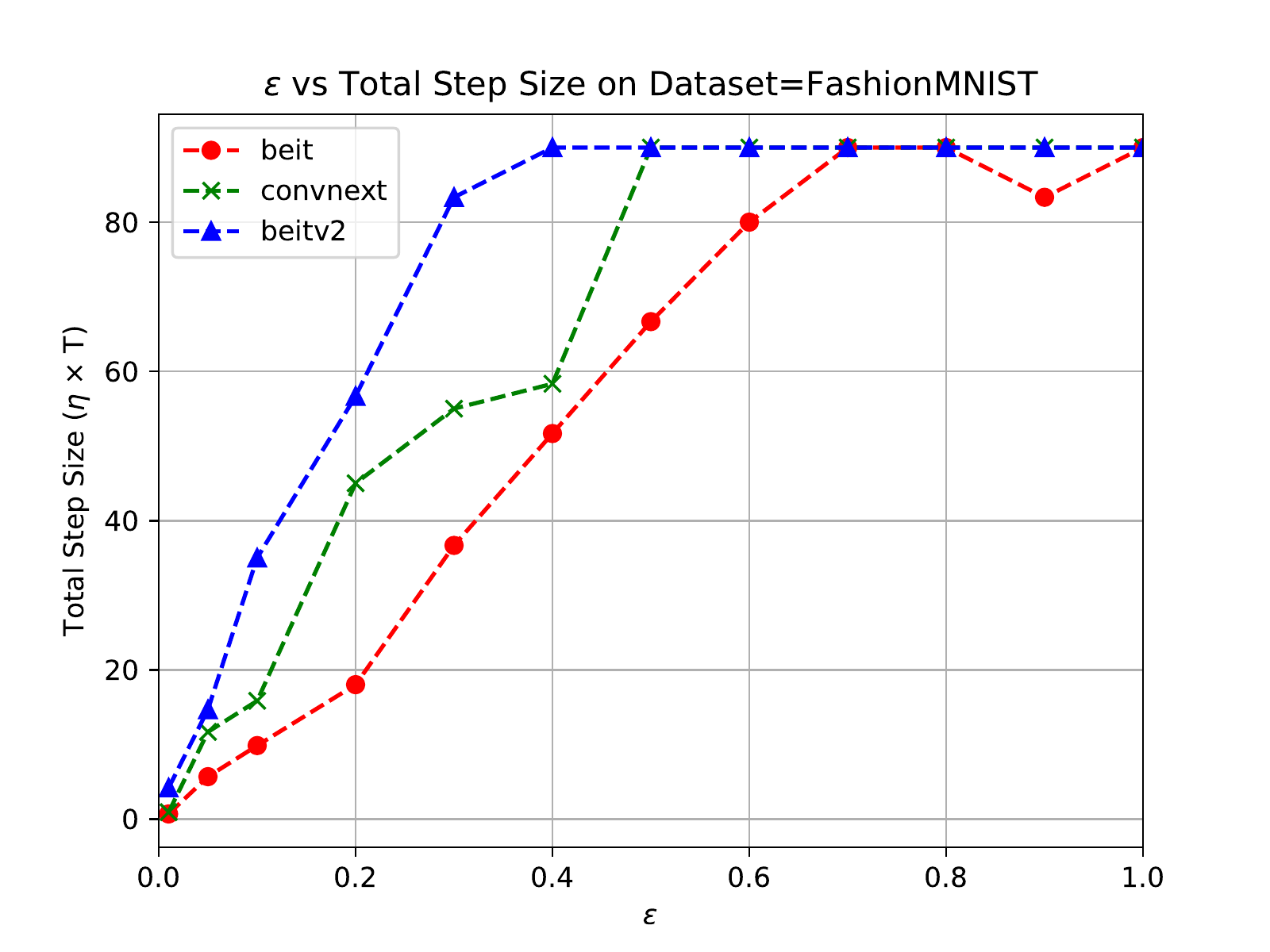}
    \end{minipage}
    }
    \subfigure[STL10 Test Accuracy]{
    \begin{minipage}[t]{0.48\linewidth}
    \centering
    \includegraphics[width=2.8in]{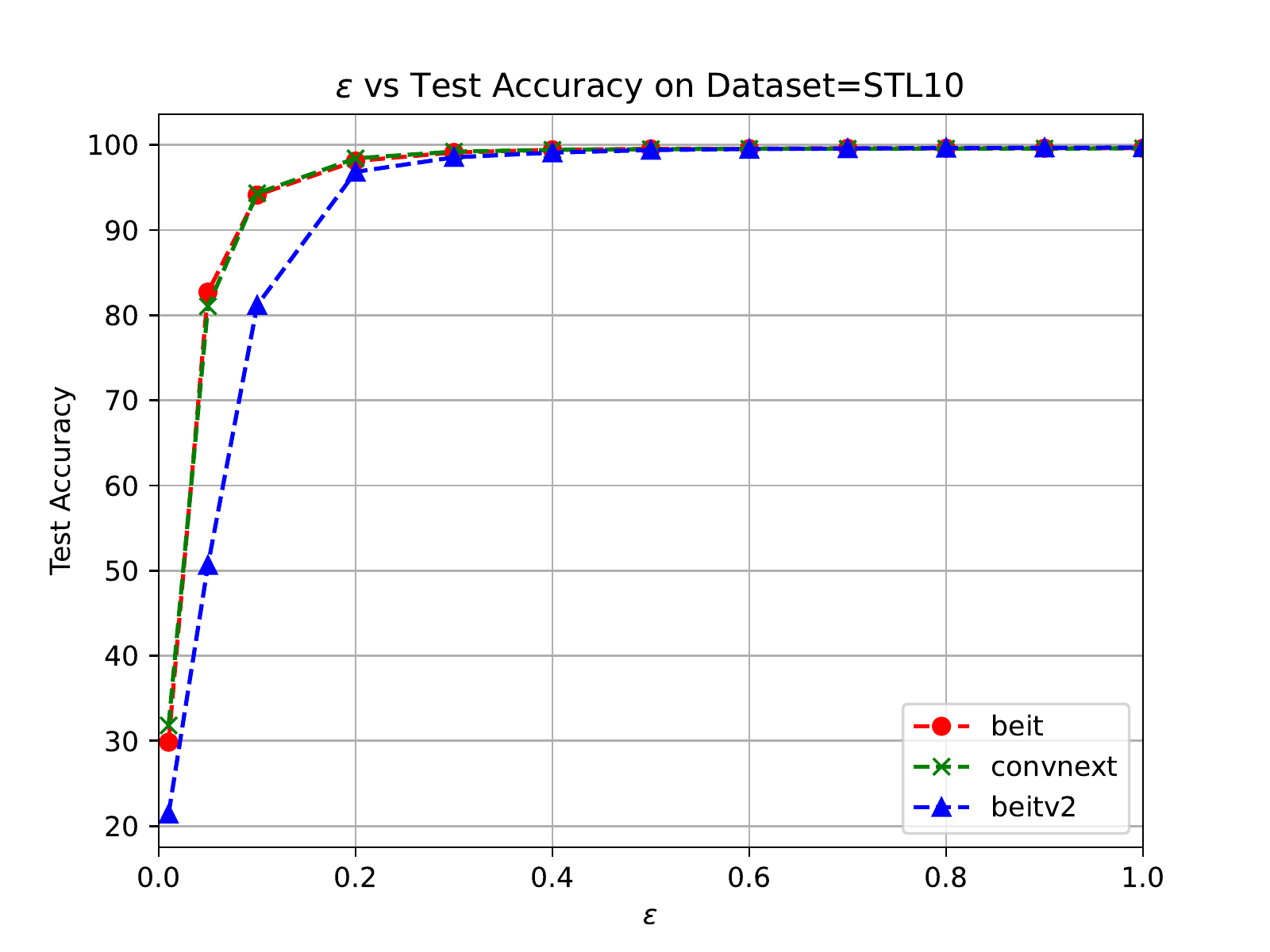}
    \end{minipage}
    }
    \subfigure[STL10 Total Step Size]{
    \begin{minipage}[t]{0.48\linewidth}
    \centering
    \includegraphics[width=2.8in]{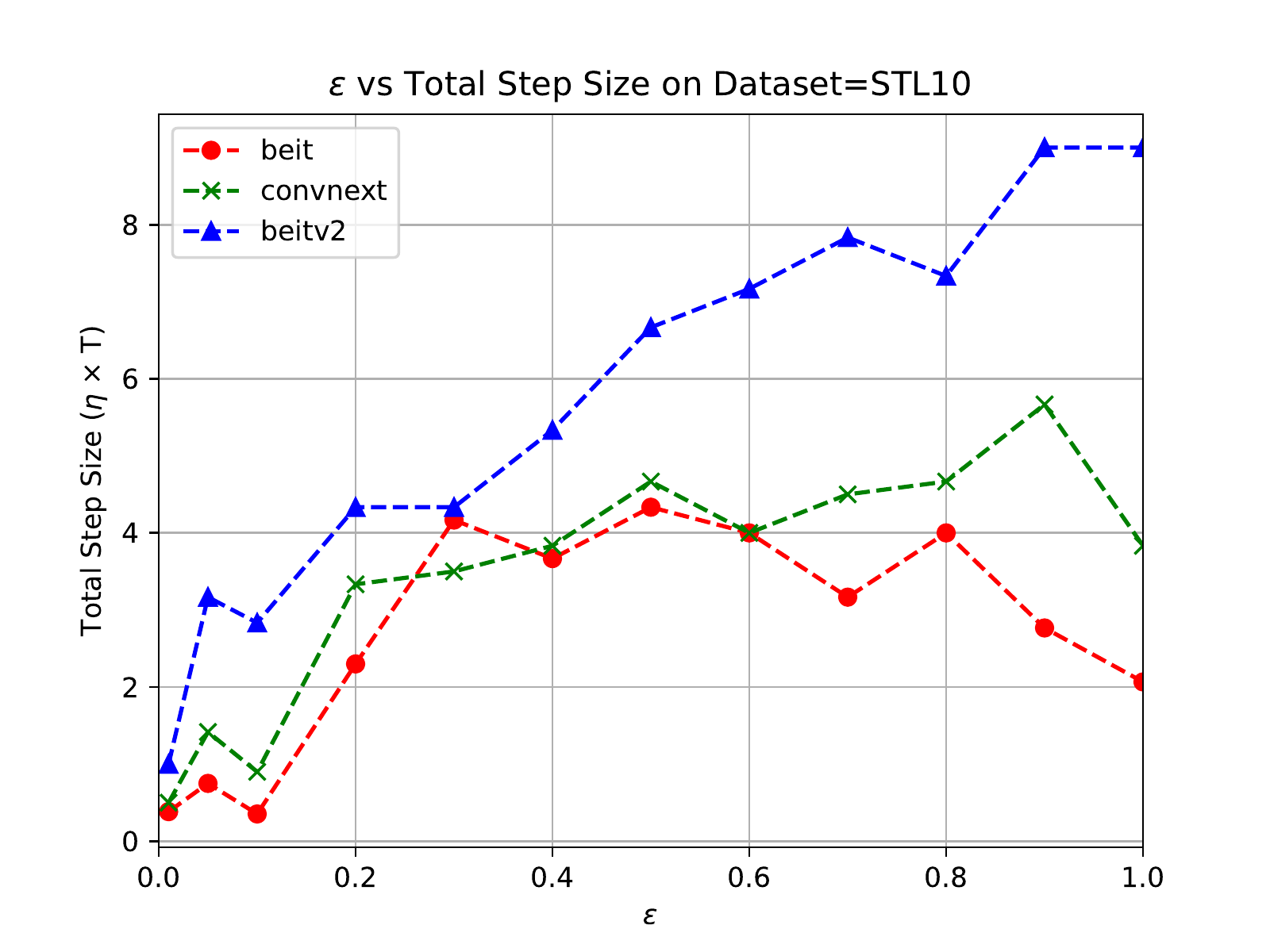}
    \end{minipage}
    }
    \centering
\caption{Pareto frontier for $\varepsilon$ vs test accuracy and total step size for STL10 and FashionMNIST. Beitv2 excels for larger values of $\varepsilon$ but beit and convnext are better for smaller values of $\varepsilon$. The inflection point varies across datasets.}
\label{fig:pareto-full-b}
\end{figure*}

\end{document}